\documentclass{article}

    \PassOptionsToPackage{numbers, compress}{natbib}

    \usepackage[preprint]{neurips_2025}

\usepackage[utf8]{inputenc} 
\usepackage[T1]{fontenc}    
\usepackage{hyperref}       
\usepackage{url}            
\usepackage{booktabs}       
\usepackage{amsfonts}       
\usepackage{nicefrac}       
\usepackage{microtype}      
\usepackage{xcolor}         

\usepackage{amsthm}
\usepackage{amsmath}
\usepackage{amssymb}
\usepackage{mathtools}
\theoremstyle{definition}
\newtheorem{definition}{Definition}

\usepackage{enumerate}
\usepackage{enumitem}
\usepackage{subcaption} 
\definecolor{softgreen}{RGB}{34, 139, 34}
\usepackage{algorithm}
\usepackage{algorithmic}
\usepackage{booktabs}
\usepackage{multirow}
\usepackage{float}
\usepackage{caption}
\usepackage{graphicx}
\usepackage{wrapfig}

\title{On the Fairness of Privacy Protection: Measuring and Mitigating the Disparity of Group Privacy Risks for Differentially Private Machine Learning}

\author{
Zhi Yang\textsuperscript{1} \thanks{Email: \texttt{12332454@mail.sustech.edu.cn}} \quad Chuangwu Huang\textsuperscript{1} \thanks{Email: \texttt{huangcw3@sustech.edu.cn}} \quad Ke Tang\textsuperscript{1} \quad Xin Yao\textsuperscript{2} \\
\textsuperscript{1}Southern University of Science and Technology\\
\textsuperscript{2}Lingnan University 
}

\begin{document}

\maketitle

\begin{abstract}
While significant progress has been made in conventional fairness-aware machine learning (ML) and differentially private ML (DPML), the fairness of privacy protection across groups remains underexplored.
Existing studies have proposed methods to assess group privacy risks, but these are based on the average-case privacy risks of data records.
Such approaches may underestimate the group privacy risks, thereby potentially underestimating the disparity across group privacy risks.
Moreover, the current method for assessing the worst-case privacy risks of data records is time-consuming, limiting their practical applicability.
To address these limitations, we introduce a novel membership inference game that can efficiently audit the approximate worst-case privacy risks of data records.
Experimental results demonstrate that our method provides a more stringent measurement of group privacy risks, yielding a reliable assessment of the disparity in group privacy risks.
Furthermore, to promote privacy protection fairness in DPML, we enhance the standard DP-SGD algorithm with an adaptive group-specific gradient clipping strategy, inspired by the design of canaries in differential privacy auditing studies.
Extensive experiments confirm that our algorithm effectively reduces the disparity in group privacy risks, thereby enhancing the fairness of privacy protection in DPML. The code is avaliable at: https://github.com/ruayz/fair-privacy.
\end{abstract}

\section{Introduction}

Artificial intelligence (AI), particularly machine learning (ML), has been widely adopted across various sectors, augmenting and even replacing human decision-making. However, its growing integration in critical domains like healthcare, finance, and judiciary has raised critical concerns, including data privacy breaches, algorithmic biases, lack of explainability, security vulnerabilities etc.~\cite{huang2022overview}. Among these ethical issues and risks, privacy and fairness have emerged as two pivotal and widely discussed challenges~\cite{ekstrand2018privacy}, attracting substantial attention from the research community.

Research in privacy protection and fairness has made significant strides independently, and the intersection of these two critical issues has also gained considerable attention. 
Some studies aim to achieve both privacy protection and outcome fairness simultaneously in ML models~\cite{xu2019achieving,jagielski2019differentially,ding2020differentially,tran2021differentially,lowy2023stochastic}.
Other works explore how privacy mechanisms affect the outcome fairness~\cite{bagdasaryan2019differential} and propose methods to mitigate the unfairness introduced by such mechanisms~\cite{xu2021removing,tran2021fairness,esipova2022disparate}.
However, whether AI systems can provide equal privacy protections to different groups is also a noteworthy yet understudied problem at the intersection of fairness and privacy.
As highlighted in~\cite{ekstrand2018privacy}, this raises an essential yet still underexplored and insufficiently addressed question: \textit{Do AI systems offer fair or equitable privacy protections across groups?} This question raises both ethical and practical concerns, as certain groups may face disproportionately higher privacy leakage risks, violating principles of fairness and equality. 

To address this issue, it is essential first to provide a rigorous answer to the question. Prior studies have empirically examined whether privacy leakage risks are evenly distributed across groups, and their auditing methods for group privacy risk rely on averaging the performance across data points within each group under membership inference attacks (MIAs)~\cite{chang2021privacy,yaghini2022disparate}. 
The MIAs employed in these studies are formulated based on the membership inference game (MIG) introduced in~\cite{yeom2018privacy}, which captures the average behavior across all data points.
However, such an average-case MIG may obscure the heterogeneity of individual privacy risks within groups and potentially underestimate the privacy leakage risks faced by certain groups. This, in turn, may lead to inaccurate and unreliable measurements of inter-group disparities in privacy risk.
Consequently, the issue of privacy inequality may not be adequately uncovered.

\textbf{Measuring.} Therefore, we aim to provide a tighter measurement of the privacy risk of data points and then analyze the disparities in privacy risk across different groups.
To this end, we can leverage the leave-one-out attack (LOOA), which is formulated based on the worst-case MIG proposed by~\cite{ye2022enhanced} and is capable of estimating the worst-case privacy leakage risk for each data point.
However, the computational cost of LOOA is prohibitively high, rendering its practical implementation nearly infeasible. 
To address this challenge, we propose an approximate version of the worst-case MIG to efficiently audit the approximate worst-case privacy risk of individual data points. 
Our experiments demonstrate that: 1) The attack simulating our proposed MIG can achieve comparable performance to LOOA as the number of attack rounds increases; 2) The individual privacy risks evaluated by our method can be reliable for assessing group privacy risk.

Then we define a fairness metric to quantify the degree of privacy unfairness (i.e., the disparity of privacy risk across groups).
Through experiments, we clearly demonstrate that our auditing method significantly outperforms previous auditing methods under the same conditions. Specifically, our method reveals greater group privacy risk and more effectively captures privacy inequality.
Consistent with prior research~\cite{chang2021privacy,yaghini2022disparate}, we find that existing ML algorithms exhibit significant unfairness in privacy risks across groups. While differentially private ML (DPML) algorithms can bind the magnitude of privacy risk disparities between groups, a certain degree of disparity still persists.

\textbf{Mitigating.} Upon providing a more thorough answer to the above question, we seek to alleviate this issue.
Inspired by the design of canaries in DP auditing studies~\cite{nasr2023tight,annamalai2024nearly,steinke2023privacy}, we confirm that groups with larger gradient norms under training process—indicating greater contributions to model updates—are more prone to higher privacy leakage risks.
Building on this insight, we enhance the existing DPML algorithm by adaptively setting group-specific gradient clipping norms. Extensive experimental results demonstrate that our algorithm effectively mitigates the disparity of group privacy risk, promoting the ethical and effective deployment of AI systems.

In summary, our main contributions are as follows:
\begin{itemize}
    \item We propose a novel MIG to efficiently and approximately audit the worst-case privacy risks of individual data points.
    \item Our auditing mechanism offers a more stringent measurement of group privacy risks, enabling a tighter and more accurate assessment of disparities between groups.
    \item We design an enhanced DPML algorithm to reduce the group privacy risk disparities, thereby improving the fairness of privacy protection.
\end{itemize}

\section{Background}
\label{sec:2}

\subsection{Differential privacy}

Differential Privacy (DP), proposed by~\cite{dwork2006calibrating}, is a privacy framework designed to address privacy leakage. It has become the predominant method for ensuring algorithmic privacy~\cite{ponomareva2023dp}. In the following, we introduce the approximate (\(\epsilon, \delta\))-DP definition.
\begin{definition}
    [$(\epsilon, \delta)$-Differential Privacy~\cite{dwork2006our}] 
    {An algorithm $\mathcal{M}$ is said to satisfy approximate differential privacy if for all pairs of adjacent databases \(D\) and \(D'\) that differ on a single data record and all possible outputs \(O \subseteq \text{Range}(\mathcal{M})\), the following condition holds:
    \begin{equation}
        P [\mathcal{M}(D) \in O] \leq e^\epsilon \times P [\mathcal{M}(D') \in O] + \delta,
    \end{equation} 
    where \(e^{\epsilon}\) provides an upper bound such that the adversary cannot distinguish whether the algorithm $\mathcal{M}$ was trained on \(D\) or \(D'\). 
    }
\label{def:dp}
\end{definition}

\paragraph{Differentially private stochastic gradient descent (DP-SGD).} 
DP-SGD~\cite{abadi2016deep} is a widely adopted algorithm in DPML~\cite{ponomareva2023dp}.
It integrates DP concepts with stochastic gradient descent (SGD). This integration ensures model privacy by employing gradient clipping and noise addition within the SGD framework, adhering to the $(\epsilon, \delta)$-DP definition. 
The pseudocode of DP-SGD is shown in Algo.~\ref{algo:DP-SGD}.

\subsection{Black-box member inference attacks}
The goal of membership inference attacks (MIAs) is to determine whether a specific data record is part of the training dataset. We focus on a black-box setting, where the adversary only has access to model outputs, reflecting a more realistic scenario where the training process is inaccessible~\cite{annamalai2024nearly}.
In black-box MIAs, the adversary infers membership by analyzing the model's output behavior, typically using the sample's output loss as an inference score or decision basis, relying on the observation that models tend to show smaller losses for training samples~\cite{yeom2018privacy}.

\paragraph{Different definitions of membership inference games (MIGs).}
MIGs conceptualize MIAs as inference games between a privacy auditor (i.e., the adversary) and a challenger. MIAs are typically carried out by simulating the MIGs through multiple rounds of random experiment. Various definitions of MIGs have been proposed, each designed to capture different aspects of privacy risk~\cite{ye2022enhanced}.

Most MIAs follow the average-case MIG framework~\cite{yeom2018privacy} (see Def.\ref{def:std_mig} in App.\ref{A})), which evaluates the vulnerability of a target model to the adversary, emphasizing the average behavior across data points~\cite{ye2022enhanced}. A common strategy formalized under this framework is the global attack (GA), where a single inference threshold is determined based on the aggregate behavior of all data points in a given round~\cite{yeom2018privacy,sablayrolles2019white}. 
The group-based attack (GBA) extends GA by assigning a distinct threshold to each group, enabling a more fine-grained analysis of group-level privacy risks and offering deeper insight into disparities across demographic partitions~\cite{chang2021privacy,yaghini2022disparate}.

Nonetheless, such average-case MIAs fail to capture worst-case privacy risks for individual data records. 
The worst-case MIG ~\cite{ye2022enhanced} (see Def.\ref{def:worst_mig} in App.\ref{A}) addresses this limitation by evaluating the maximum risk a single record may encounter. A concrete instance is the Leave-One-Out Attack (LOOA), which independently evaluates each data point’s worst-case exposure, aligning closely with the principles of DP. However, LOOA is computationally intensive, as it requires evaluating each record separately.

\paragraph{Privacy auditing.}
Privacy auditing is designed to proactively assess privacy risks and quantify potential leakage, typically during the model development phase. In contrast, MIAs are conducted post-deployment by adversaries aiming to exploit trained models.
Privacy auditing uses MIAs for evaluation but with more background knowledge for the adversary, including access to the original dataset and knowledge of the optimal threshold. This setup simulates worst-case scenarios, enabling rigorous assessment of privacy leakage risks~\cite{chang2021privacy}. 
In this work, we focus on privacy auditing to systematically evaluate privacy vulnerabilities.
Privacy auditing using the GA (PA-GA)\cite{yeom2018privacy}, the GBA (PA-GBA)~\cite{chang2021privacy,yaghini2022disparate}, and the LOOA (PA-LOOA)~\cite{nasr2023tight,annamalai2024nearly} for evaluating privacy leakage risks is detailed in Algos.~\ref{algo:pa-ga},~\ref{algo:pa-gba}, and~\ref{algo:pa-looa} of App.~\ref{A}, respectively.


\subsection{The fairness of privacy}
Few studies have explored fairness in the context of privacy, and those that do vary in their research focus.
For instance, one study finds that fairness-aware algorithms can exacerbate disparities in privacy leakage across groups~\cite{chang2021privacy}.
Meanwhile, other research highlights that ML algorithms exhibit significant group-level disparities in privacy leakage, and that DPML algorithms can help reduce these disparities~\cite{yaghini2022disparate}.
These auditing mechanisms typically rely on average-case attacks, which assess privacy risks based on the average behavior of data points.
While effective at capturing general trends, such approaches may overlook the nuanced privacy risks faced by individual samples, potentially concealing disparities between groups.

In our work, we aim to more precisely evaluate group-level privacy risks compared to prior studies.
To ensure fair comparison, we adopt the same attack metrics as in~\cite{chang2021privacy, yaghini2022disparate}, including inference scores based on per-example loss and attack success rates.

\section{Measuring the disparity of group privacy risks with approximate worst-case privacy auditing}
\label{sec:3}
In this section, we first propose an alternative to PA-LOOA and demonstrate that, while it improves efficiency, it achieves comparable performance as the number of repeated experiments increases. We further introduce a fairness metric to assess the disparity of inter-group privacy risks. Experimental results show that our method uncovers more pronounced group privacy risks and offers a more reliable assessment of privacy inequalities between groups compared to existing approaches, thereby exhibiting stronger auditing capability.

\subsection{Approximate worst-case privacy auditing}
As previously discussed, the LOOA is computationally expensive. Specifically, obtaining statistically reliable results for a single sample typically requires $2R$ repeated experiments. Consequently, auditing $m$ samples would involve training $m \times 2R$ models, making this approach impractical for real-world applications due to the extensive time and computational resources required. To address this limitation, we propose a new MIG that allows for the simultaneous auditing of multiple samples within a single auditing process, as presented in Def.~\ref{def:a_mig}. 

\begin{definition}[{Approximate Worst-case MIG}] Let $\Omega$ denotes the underlying population data pool, $\mathcal{M}$ the training algorithm, and $\mathcal{A}$ the inference algorithm. 
We assume that the challenger samples $n$ i.i.d. records from $\Omega$ to construct the training dataset $D$, and $Z=\{z_i\} _{i=1}^{m} \subseteq D$ represents the auditing samples.
    \begin{itemize}[itemsep=0pt]
        \item [1)] The challenger flips fair choices $ \{ h_i \}_{i=1}^{m} $ randomly, where $ h_i \in \{0, 1\} $, indicating whether each record $ z_i $ is included in the training or not.
        \item [2)] The challenger samples a fixed record $ z \sim Z $ along with its status $h$.
        \item [3)] The challenger trains a model $ f_{h} \gets \mathcal{M}(D \setminus \{z_i \mid h_i = h\}) $, and a model $ f_{\sim h} \gets \mathcal{M}(D \setminus \{z_i \mid h_i = \sim h\}) $.
        \item [4)] The challenger flips a fair coin $b \in \{0, 1\}$, and sends the target model and record $ (f_b, z) $ to the adversary.
        \item [5)] The adversary, with access to the target model, outputs a guess $ \hat{b} \leftarrow \mathcal{A}(f_b, z) $.
        \item [6)] The game outputs 1 (success) if $ \hat{b} = b $, and 0 otherwise.
    \end{itemize}
\label{def:a_mig}
\end{definition}

\begin{wrapfigure}{r}{0.45\linewidth}
    \centering
    \includegraphics[width=\linewidth]{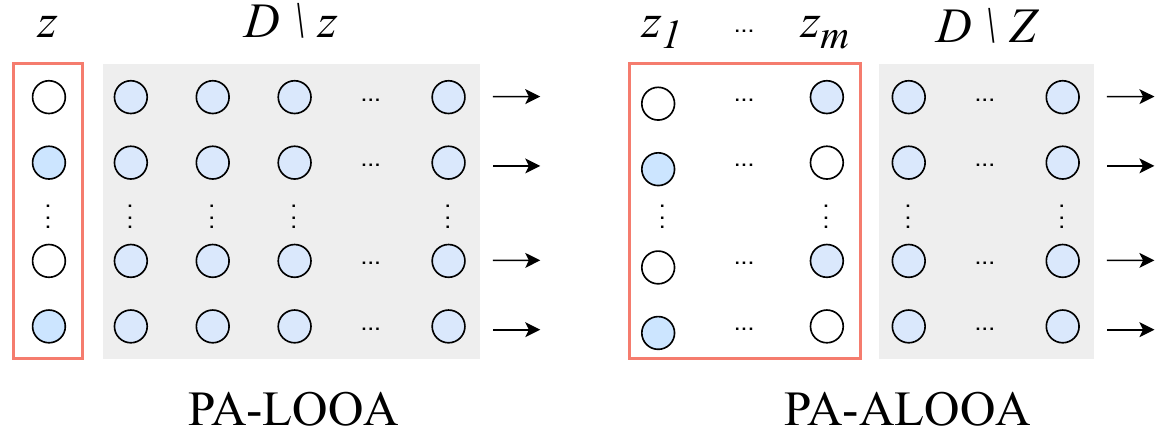}
    \caption{Left: PA-LOOA audits a single sample. Right: PA-ALOOA audits $m$ samples. Solid circles indicate training points; hollow circles are excluded. Arrows denote model training using solid-circle data.}
    \label{fig:1}
\end{wrapfigure}

We refer to the attack that simulates this game as the approximate leave-one-out attack (ALOOA). 
The process of privacy auditing using ALOOA (PA-ALOOA) to evaluate privacy risks is detailed in Algo.~\ref{algo:pa-alooa} of App.~\ref{A}. 
The auditing mechanism achieves computational efficiency by auditing multiple samples simultaneously while preserving analytical granularity by analyzing the behavior of each sample, rather than relying on aggregate statistics across multiple data points.


\paragraph{Comparison with PA-LOOA.}

The difference between the two approaches stems from their sample selection strategies.
As shown in Fig.~\ref{fig:1}, PA-LOOA introduces minimal randomness into the training set, as only a single sample is randomly included or excluded in each round. In contrast, PA-ALOOA audits $m$ samples simultaneously, with each experiment randomly choosing which samples are included in the training set, thus introducing greater variability. However, we argue that with sufficient rounds, the random fluctuations in PA-ALOOA will average out, resulting in performance comparable to that of PA-LOOA.

We validate our hypothesis through practical experiments using the widely used MNIST dataset, training a Convolutional Neural Network (CNN) with SGD.
Due to computational limitations, we randomly select 60 samples per class, totaling 600 samples to audit for PA-LOOA and PA-ALOOA.
We evaluate the attacker's performance using the accuracy metric (i.e., attack success rate) from~\cite{shokri2017membership}, which measures the agreement between the adversary's guesses and the actual status. Instead of evaluating overall accuracy, we compute individual accuracy for each data point. 
Detailed experimental settings and additional results are provided in the App. \ref{app:cop_pa-looa}.

\begin{figure}[htbp]
    \centering
    \begin{subfigure}[b]{0.48\textwidth} 
        \centering
        \includegraphics[width=0.9\linewidth]{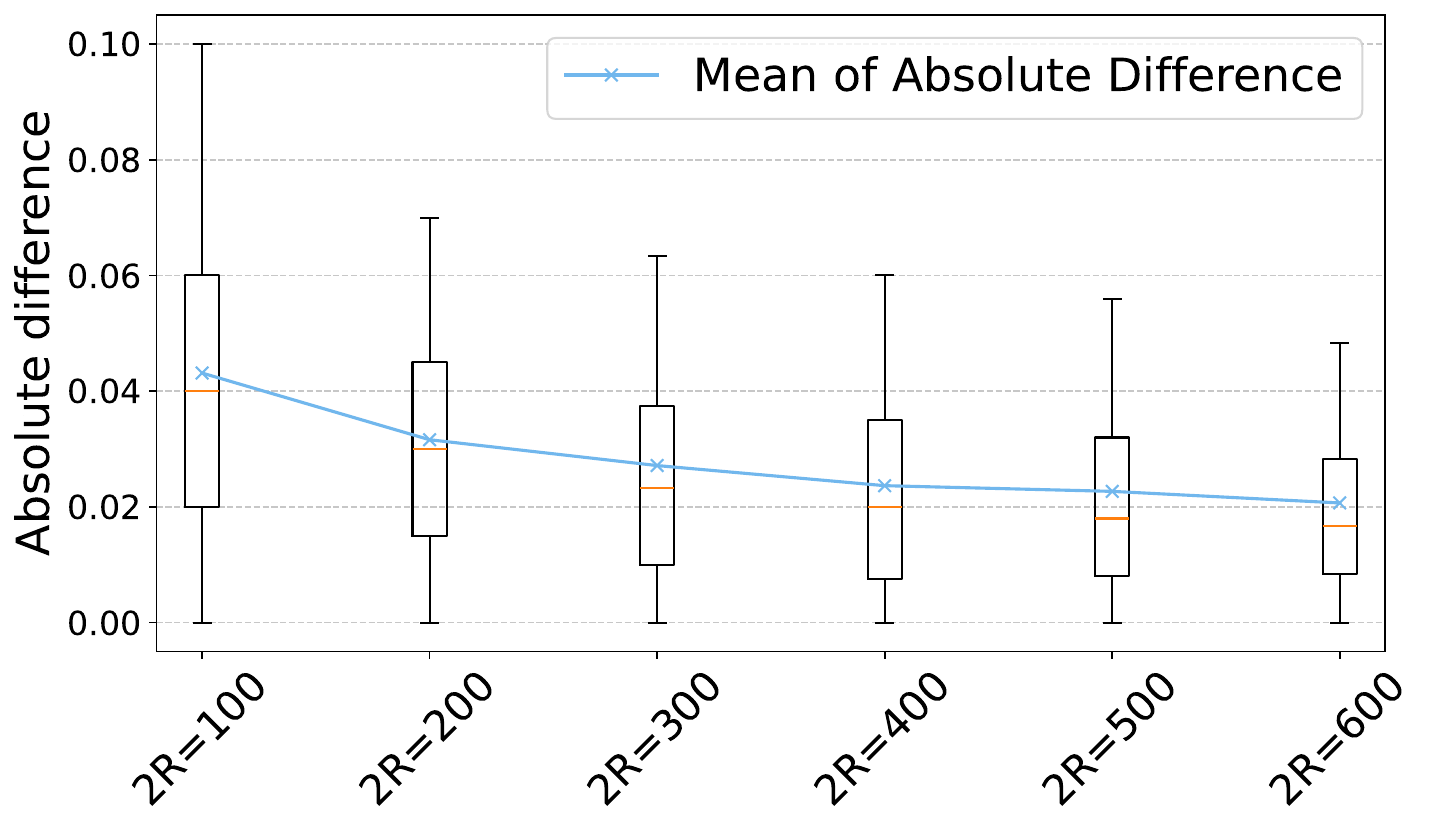} 
        \caption{Comparison at the individual level.}
        \label{model_compare}
    \end{subfigure}%
    \hfill
    \begin{subfigure}[b]{0.48\textwidth} 
        \centering
        \includegraphics[width=0.9\linewidth]{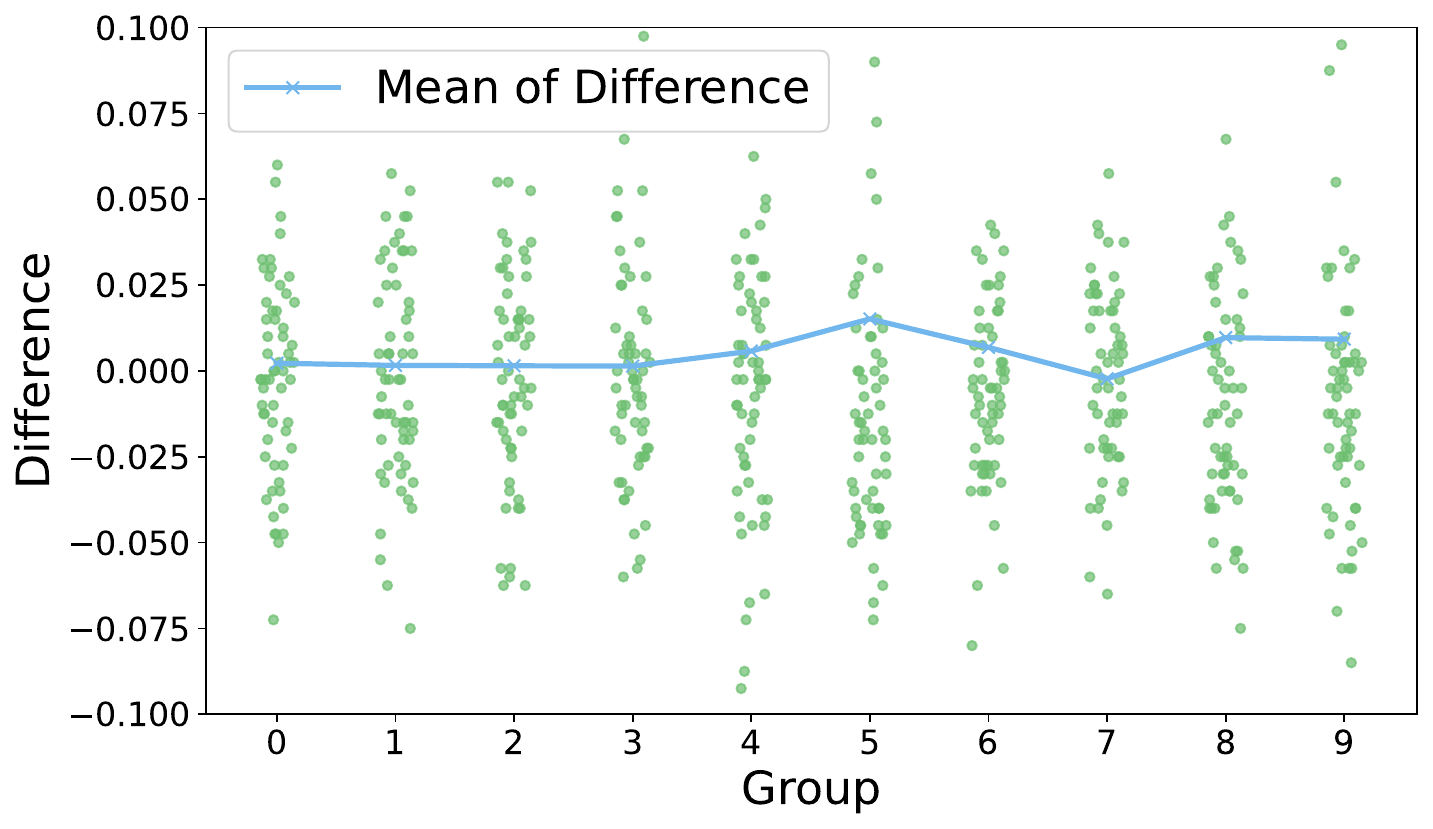} 
        \caption{Comparison at the group level.}
        \label{group_compare}
    \end{subfigure}%
    \caption{Left: The horizontal axis represents the number of random experiments for a single audit, while the vertical axis represents the absolute difference in auditing performance between the two attacks for each audited sample. Right: The horizontal axis represents different groups of the MNIST dataset, while the vertical axis indicates the performance difference between PA-LOOA and PA-ALOOA for individual data points in each group at $2R=400$.}
\end{figure}

As shown in Fig.~\ref{model_compare}, the experimental results indicate that as the number of random trials increases, the average absolute difference in individual accuracy between the two methods gradually decreases and eventually stabilizes. The signed differences are presented in Fig. \ref{model_compare_signed} of App. \ref{app:cop_pa-looa}, from which we observe that the two approaches exhibit similar behavior on average, with the mean difference consistently remaining below 0.01 across all $2R$ values. Moreover, the variance of the differences further decreases as $2R$ increases.
However, it is evident that the discrepancy between the two approaches still exhibits notable variance across individual data points. 

Although the estimation errors for individual samples may vary significantly, we find that statistical outcomes across groups are reliable, forming a solid foundation for analyzing group privacy risk. 
As shown in Fig.~\ref{group_compare}, the distribution of performance differences between PA-LOOA and PA-ALOOA within each group is highly similar. Moreover, the Kruskal-Wallis test yields $p$-values greater than 0.4 for all $2R$ values considered in our experiment, indicating no statistically significant differences in performance between the two methods across groups.
Furthermore, the average performance difference of each group between PA-LOOA and PA-ALOOA is minimal, suggesting that extending individual-level estimates to group-level statistical analysis introduces only negligible error.

\subsection{Definition of group privacy risk parity}
We assess the privacy leakage risk of data points using the attacker's membership advantage, following prior work~\cite{yaghini2022disparate, yeom2018privacy}. Below, we formally define the notion of individual privacy risk (IPR) as applied to a single data record in this study.
\begin{definition}[Individual Privacy Risk] Let $Acc_i(\mathcal{A}, Z)$ represent the attack accuracy of a data $i$ under privacy auditing algorithm $\mathcal{A}$ and the auditing dataset $Z$. The individual privacy risk is defined as:
    \begin{equation}
        Adv_i(\mathcal{A}, Z) = 2Acc_i(\mathcal{A}, Z) - 1
    \end{equation} 
\end{definition}
This formulation quantifies the adversary's normalized advantage over random guessing.
Building on the concept of IPR, we can extend it to define Group Privacy Risk (GPR) as in ~\cite{chang2021privacy}.  
\begin{definition}[Group Privacy Risk] Let $D^k$ denote the subset of the dataset $D$ belonging to group $k$. The group privacy risk is defined as:  
    \begin{equation}
        Adv^k(\mathcal{A}, Z) = \mathbb{E}_{i \in D^k}[Adv_i(\mathcal{A}, Z)]
    \end{equation} 
\end{definition}
Based on GPR, we evaluate whether privacy leakage risk is fair or equitable across different groups by introducing the notion of Group Privacy Risk Parity (GPRP).
\begin{definition}[Group Privacy Risk Parity] Let $K$ represent the set of all groups. We define group privacy risk parity as:
    \begin{equation}
        \Delta = \max_{k \in K}(Adv^k(\mathcal{A}, Z)) - \min_{k \in K}(Adv^k(\mathcal{A}, Z))
    \end{equation} 
\end{definition}
This metric provides a systematic means of quantifying the disparity in privacy risk across groups, capturing the gap between the most and least vulnerable groups.

\subsection{Comparison with privacy auditing by average-case attacks}
We compare the GPR and the GPRP metrics measured by our auditing method, PA-ALOOA, with those obtained from privacy auditing by average-case attacks (PA-ACAs) in previous studies: PA-GA~\cite{yaghini2022disparate} and PA-GBA~\cite{chang2021privacy}. 
To ensure a fair comparison, all three methods are configured identically, keeping the training dataset and model consistent across each repeated experiment. 
Specifically, for PA-ALOOA, we set the number of audit samples to $m=n$, following the same setting used in PA-ACAs. This setup ensures consistency and reflects a real-world scenario in which the privacy risk of every training sample is audited.
The key distinction lies in threshold determination: PA-ACAs computes thresholds based on the aggregate behavior of multiple data points, whereas PA-ALOOA assigns a unique threshold to each sample, based on its behavior across all repeated experiments.

\begin{figure}[htbp]
    \centering
    \begin{subfigure}[b]{0.3\textwidth} 
        \centering
        \includegraphics[width=\linewidth]{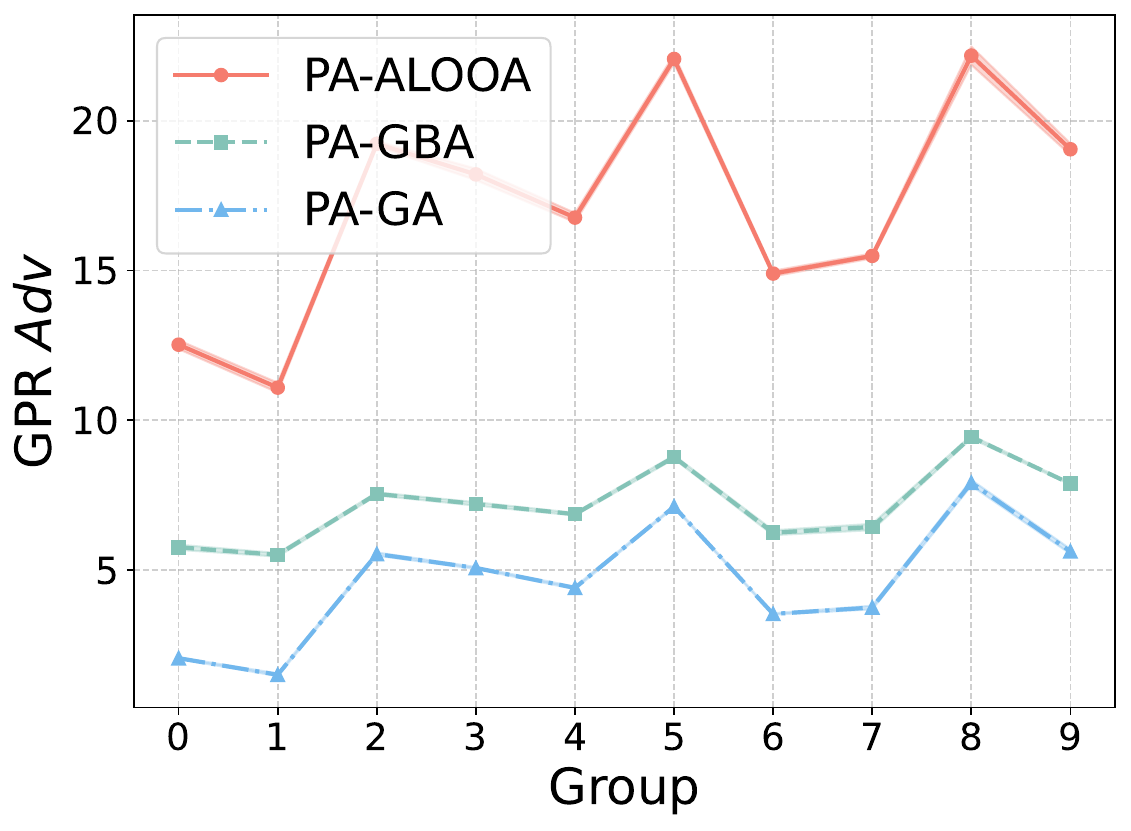} 
        \caption{LR.}
        \label{fig3.3:LR}
    \end{subfigure}%
    \hfill
    \begin{subfigure}[b]{0.3\textwidth} 
        \centering
        \includegraphics[width=\linewidth]{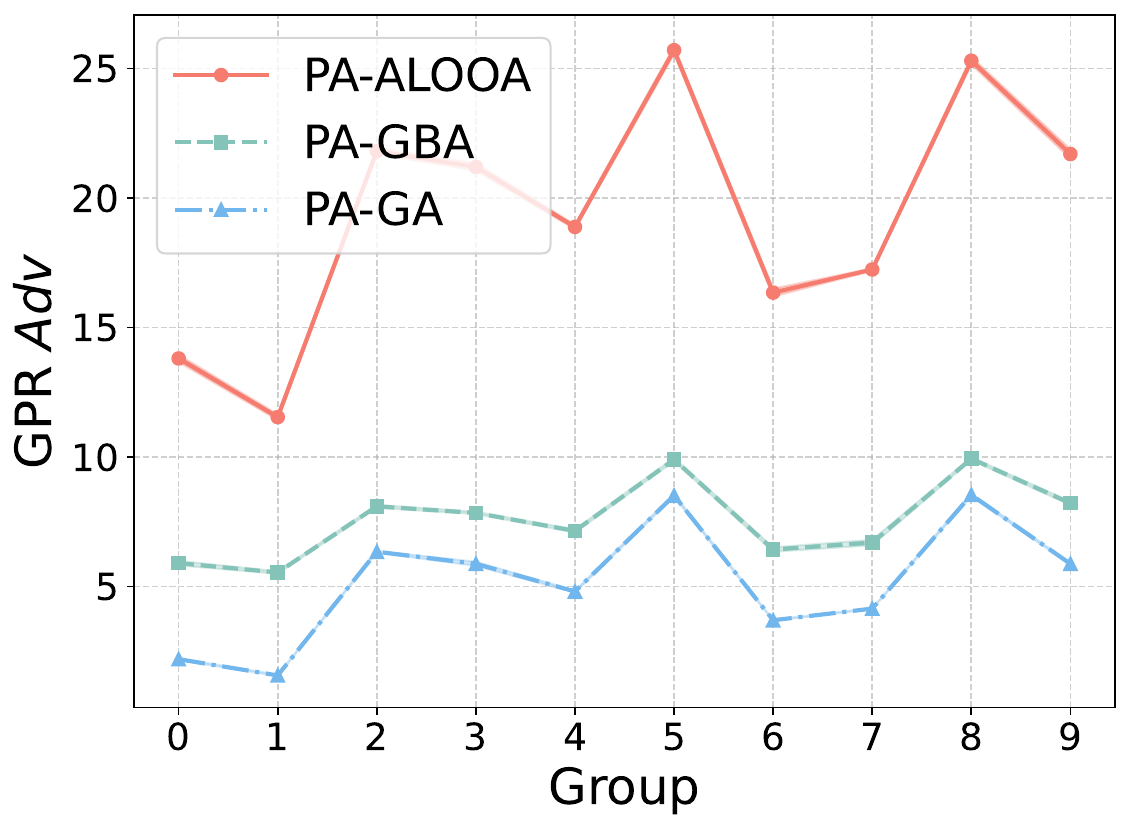} 
        \caption{MLP.}
        \label{fig3.3:MLP}
    \end{subfigure}%
    \hfill
    \begin{subfigure}[b]{0.3\textwidth} 
        \centering
        \includegraphics[width=\linewidth]{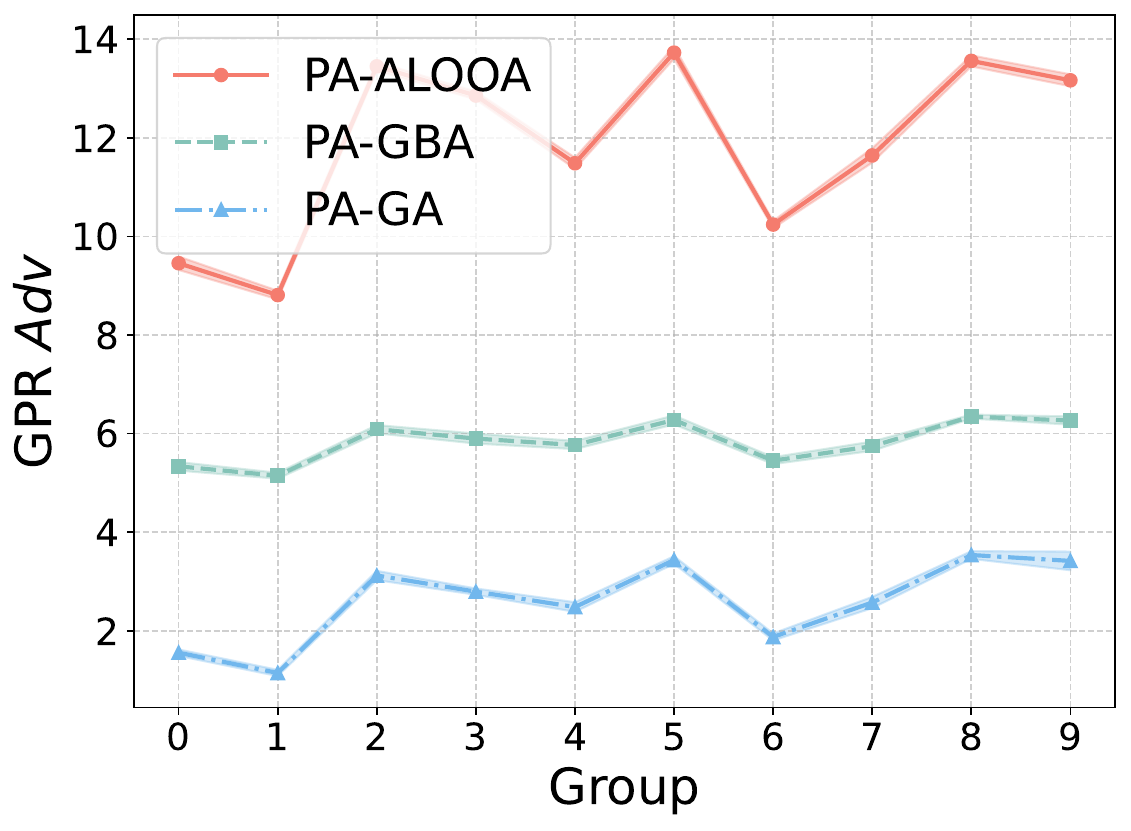} 
        \caption{CNN.}
        \label{fig3.3:CNN}
    \end{subfigure}
    \caption{The comparison of GPR value across three model types—Logistic Regression (LR), Multilayer Perceptron (MLP), and CNN—trained on the MNIST dataset using SGD algorithm. The x-axis represents the groups, and the y-axis shows the corresponding GPR value at $2R=400$.}
    \label{gpr_compare}
\end{figure}

Consistent with prior studies on privacy and fairness~\cite{bagdasaryan2019differential,xu2021removing,esipova2022disparate}, our foundational analysis focuses on the MNIST dataset. Detailed experimental configurations and supplementary results for other datasets are included in the App. \ref{app:cop_pa-acas}.
The results of the GPR and GPRP metrics on the MNIST dataset are presented in Fig.~\ref{gpr_compare} and Tab.~\ref{tab:delta_compare}, with all values reported in percentage points for clarity.  
As shown in Fig.~\ref{gpr_compare}, the three auditing methods exhibit consistent patterns in the distribution of privacy risks across groups. Among them, PA-ALOOA consistently yields significantly higher GPR values across all model architectures compared to the other two auditing methods.
Specifically, Tab.~\ref{tab:delta_compare} shows that for the CNN model, the GPRP value obtained by PA-GA and PA-GBA suggests that DP-SGD results in higher disparity than standard SGD. This observation contradicts the conclusion in \cite{yaghini2022disparate}, which asserts that DPML algorithms should be able to bind the disparity of GPR relative to non-private counterparts. Such inconsistency implies that PA-ACAs underestimate the GPR, leading to inaccurate measurements and incorrect conclusions.
In summary, PA-ALOOA offers a more rigorous means of capturing privacy risk, and thus providing more reliable evaluations of privacy unfairness across groups.

\begin{table}[h]
    \centering
    \caption{The comparison of GPRP value computed by different privacy auditing methods for a CNN model trained on the MNIST dataset at $2R=400$.}
    \label{tab:delta_compare}
    \begin{tabular}{lccc}
        \toprule
        \textbf{Method} &  \textbf{$\Delta_{PA-GA}$} & \textbf{$\Delta_{PA-GBA}$} &  \textbf{$\Delta_{PA-ALOOA}$} \\
        \midrule
        SGD & 2.452 & 1.248 & 4.920 \\
        DP-SGD & 2.625 & 1.254 & 3.540\\
        \bottomrule
    \end{tabular}
\end{table}

\section{Mitigating the disparity of group privacy risks for DP-SGD}
While DPML algorithms can limit the extent of privacy risk disparities across groups, previous results demonstrate that such disparities still persist to a noticeable degree. In this section, we investigate and confirm a strong correlation between GPR and the group contribution of gradients during training. Motivated by this finding, we propose an enhanced DP-SGD algorithm designed to improve the fairness of privacy protection across different groups.

\subsection{Experimental observations}

In DP auditing literature, canaries are often created as mislabeled samples~\cite{nasr2023tight,annamalai2024nearly,steinke2023privacy}. These samples generate larger gradient values during model training, which contribute more to parameter updates, thereby increasing the likelihood of being memorized by the model.
Motivated by the design of canaries, we hypothesize that during training, the larger a group's contribution to the gradient, the more likely the model is to memorize that group. Thus, groups with larger contributions are expected to face a higher privacy leakage risk compared to those with smaller contributions. 

We conduct experimental analysis to validate our hypothesis.
We first compute the sum of gradient vectors for a group $k$ within a batch $B$ and average it by dividing by the number of samples in that group $|D^k|$, i.e., $\sum_{i \in B \cap D^k}{g}_{i} / |D^k|$. The norm of this vector is then divided by the norm of the gradient used for the model update, i.e., $\sum_{i \in B}{g_i}/|B|$, to represent the group's relative contribution in this iteration. We obtain group relative contribution (GRC) by averaging this ratio across all training iterations.
As shown in Fig.~\ref{app:adv_and_grc} of App.~\ref{app:add_set&res}, there is a significant correlation between GRC and GPR across different models, confirming our hypothesis. Specifically, groups contributing more during training exhibit higher privacy leakage risks.

\subsection{Design of mitigation algorithm for DP-SGD}
\label{sec:4.2}
Building on the previous observation, we propose an improvement to the DP-SGD algorithm to promote fair privacy protection across groups. In DP-SGD, the gradient clipping operation uses a unified clipping bound for all groups, whereas we adaptively set different clipping bounds for each group based on the GRC during training.
\begin{algorithm}[H]
    \caption{DP-SGD-S}
    \label{algo:DP-SGD-Scale}
    \begin{algorithmic}[1]
        \REQUIRE Training dataset $D=\{ (\mathbf{x}_{i},y_{i}) \} _{i=1}^{n}$, 
        the parameterized model $f_{w}(\cdot)$, loss function $\ell$,
        iterations $T$, batch size $b$, learning rate $\eta$, noise scale $\sigma_1$, $\sigma_2$, clipping bound $C$, scale bound $\tau$.
        \STATE Initialize $w^{(0)}$ randomly.
        \FOR {$ t = 0, ..., T-1 $} 
            \STATE Sample a batch $B$ from $D$ with probability $b/N$.
            \FOR { ${i} \in B$}
                \STATE $ g_{i} \leftarrow \nabla\ell(f_{{w^{(t)}}}(\mathbf x_{i}), y_{i}) $
                \STATE $ \bar{g}_{i} \leftarrow g_{i} \cdot \min\left(1, \frac{1}{\|g_i\|_2} \right) $ 
            \ENDFOR 
            \textcolor{softgreen}{
            \FOR { $k \in K$}
                \STATE $ {C}^{k} \leftarrow C \cdot \min\left(\tau, \frac{\| \frac{1}{b} (\sum_{i \in B}{\bar{g}_{i}} + \mathcal{N}(0, \sigma_1^2 \mathbf{I})) \|_2}{\| \frac{1}{|D^k|} (\sum_{i \in B \cap D^k}{\bar{g}_{i}} + \mathcal{N}(0, \sigma_1^2 \mathbf{I}))   \|_2}\right)$
            \ENDFOR
            \FOR { ${i} \in B$}
                \STATE $ \bar{g}_{i} \leftarrow g_{i} \cdot \min(1, \frac{C^k}{\|g_{i}\|_2}) $
            \ENDFOR     
            \STATE $C = \max_{k \in K}({C}^{k})$
            }
            \STATE $ \tilde{g} \leftarrow \frac{1}{b} \left ( \sum_{i \in B}\bar{g}_{i} + \mathcal{N}(0, \sigma_2^2 C^2 \mathbf{I}) \right ) $
            \STATE $ w^{(t+1)} \leftarrow w^{(t)} - \eta \tilde{g} $
        \ENDFOR
        \ENSURE Model $f_{w^{(T)}}(\cdot)$ and accumulated $(\epsilon, \delta)$.
    \end{algorithmic}
\end{algorithm}
As shown in Algo.~\ref{algo:DP-SGD-Scale}, our proposed algorithm, DP-SGD-Scale (abbreviated as DP-SGD-S), differs from the standard DP-SGD in Lines 6–14. In each iteration, it estimates the relative contribution of each group’s samples to the overall gradient and uses this information to adaptively adjust the clipping bound for each group.
To preserve privacy, we add noise to the clipped group-level gradient statistics used for computing the group-specific clipping bounds $C^k$. This additional privacy cost is incorporated into the overall privacy accounting via the composition theorem~\cite{abadi2016deep,xu2021removing}. 
Although different groups are assigned distinct clipping bounds, we conservatively bound the sensitivity of the final aggregated gradient by $\max_k C^k$, ensuring that the overall mechanism satisfies $(\epsilon, \delta)$-DP in the same sense as DP-SGD. Following prior work~\cite{xu2021removing,esipova2022disparate}, we set $\sigma_1 \approx 10\sigma_2$ so that the privacy cost of computing $C^k$ is negligible relative to the total privacy budget.


In DP-SGD-S, groups with higher contributions have their clipping bounds scaled down, leading to stricter clipping operations. This adjustment limits the influence of these groups on model updates, thereby reducing the model's memorization of these groups and mitigating their privacy leakage risks. 
Conversely, groups with relatively smaller contributions are assigned larger clipping bounds. The scaling factor of clipping bounds is constrained by the hyperparameter $\tau$, as excessively large clipping norms would introduce too much noise, making the model's performance unreliable.

\section{Experimental study}
In this section, we validate the effectiveness of our algorithm, DP-SGD-S, in mitigating the disparity of privacy risk across groups through extensive experiments. 

\subsection{Experimental setup}
Full experimental details are provided in App.~\ref{app:setup}.
We conduct experiments on datasets commonly used in privacy and fairness research~\cite{annamalai2024nearly,bagdasaryan2019differential,xu2021removing}, including MNIST~\cite{lecun1998mnist}, as well as three fairness-related datasets: two tabular datasets, Adult and Law\cite{le2022survey}, and one image dataset, UTKFace\cite{zhang2017age}.
Our study compares three training algorithms: standard SGD, DP-SGD, and our proposed DP-SGD-S. For both DP-SGD and DP-SGD-S, the default privacy parameters are set to $(\epsilon, \delta) = (10, 1\mathrm{e}{-5})$, and the default scale bound $\tau$ for DP-SGD-S is set to 2. Three model architectures are considered: LR, MLP, and CNN.
To measure fairness in privacy protection across groups, we use the GPRP metric, assessed via our auditing method PA-ALOOA. The model utility is evaluated through classification accuracy.
All results reported represent the average of five independent runs, with all values presented in percentage points for clarity.

\subsection{Experimental results}
\paragraph{Results across different datasets.}
We evaluate our proposed algorithm, DP-SGD-S, on multiple datasets to demonstrate its effectiveness in mitigating disparities in privacy leakage risks among groups.
As shown in Table~\ref{tab:diff_datasets}, DP-SGD-S consistently achieves the lowest $\Delta$ across all datasets. These results highlight that our enhancement to DP-SGD leads to a more equitable privacy protection mechanism.
For the tabular datasets Adult and Law, the classification accuracy remains nearly unchanged between the non-private and private training algorithms. In these cases, DP-SGD-S successfully reduces $\Delta$ without compromising model utility.
For the image datasets, MNIST and UTKFace, DP-SGD leads to an accuracy drop of approximately 2\% compared to standard SGD, and DP-SGD-S incurs a drop of about 2\% compared to DP-SGD. 
This indicates that DP-SGD-S incurs a slight accuracy trade-off in this scenario, but the degradation is modest and accompanied by enhanced fairness in privacy protection. The results of the other dataset are shown in App.~\ref{app:5.2}.

\begin{table*}[htbp]
    \centering
    \caption{The results of three training algorithms under different datasets.}
    \label{tab:diff_datasets}
    \begin{tabular}{llcccccc}
        \toprule
        \textbf{Metric} & \textbf{Method} & \textbf{MNIST} & \textbf{Adult} & \textbf{Law} & \textbf{UTKFace} \\
        \midrule
        \multirow{3}{*}{Accuracy~$(\uparrow)$} 
          & SGD & 95.89 $\pm$ 0.29 & 85.00 $\pm$ 0.07 
          & 89.75 $\pm$ 0.12 & 85.93 $\pm$ 3.79\\
          & DP-SGD & 94.46 $\pm$ 0.13 & 84.92 $\pm$ 0.04 
          & 89.74 $\pm$ 0.08 & 86.84 $\pm$ 0.48 \\
          & DP-SGD-S & 92.57 $\pm$ 0.42 & 84.86 $\pm$ 0.08 
          & 89.60 $\pm$ 0.11 & 84.56 $\pm$ 0.11\\
        \midrule
        \multirow{3}{*}{GPRP $\Delta~(\downarrow)$} 
          & SGD & 4.92 $\pm$ 0.18 & 0.42 $\pm$ 0.04 
          & 0.90 $\pm$ 0.16 & 1.75 $\pm$ 0.11 \\
          & DP-SGD & 3.54 $\pm$ 0.13 & 0.27 $\pm$ 0.04 
          & 0.59 $\pm$ 0.06 & 1.19 $\pm$ 0.07\\
          & DP-SGD-S & 2.92 $\pm$ 0.14 & 0.16 $\pm$ 0.02 
          & 0.43 $\pm$ 0.03 & 0.74 $\pm$ 0.07\\
        \bottomrule
    \end{tabular}
\end{table*}

\begin{figure}[htb]
    \centering
    \begin{minipage}[t]{0.48\linewidth}
    \centering
    \includegraphics[width=0.8\linewidth]{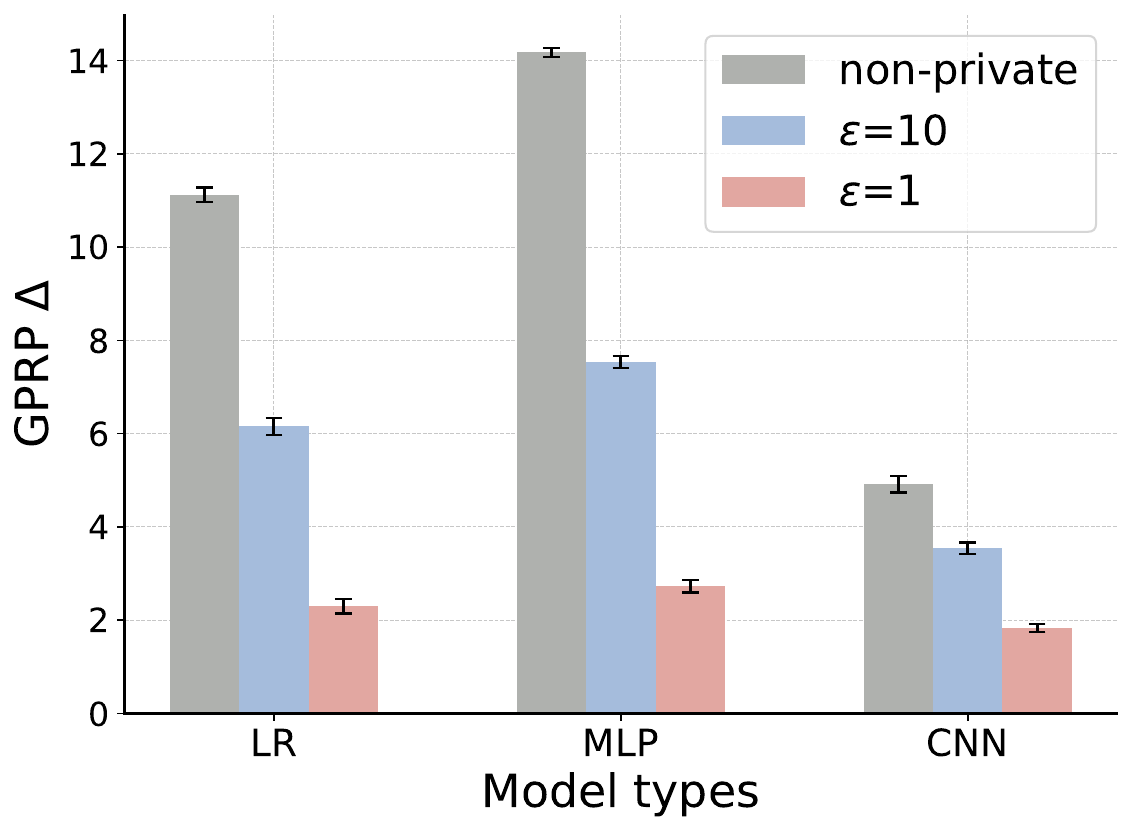}
    \caption{The results of the SGD and DP-SGD algorithms on the MNIST dataset under varying privacy guarantees and model architectures.}
    \label{fig:eps}
    \end{minipage}\hfill
    \begin{minipage}[t]{0.48\linewidth}
    \centering
    \includegraphics[width=0.8\linewidth]{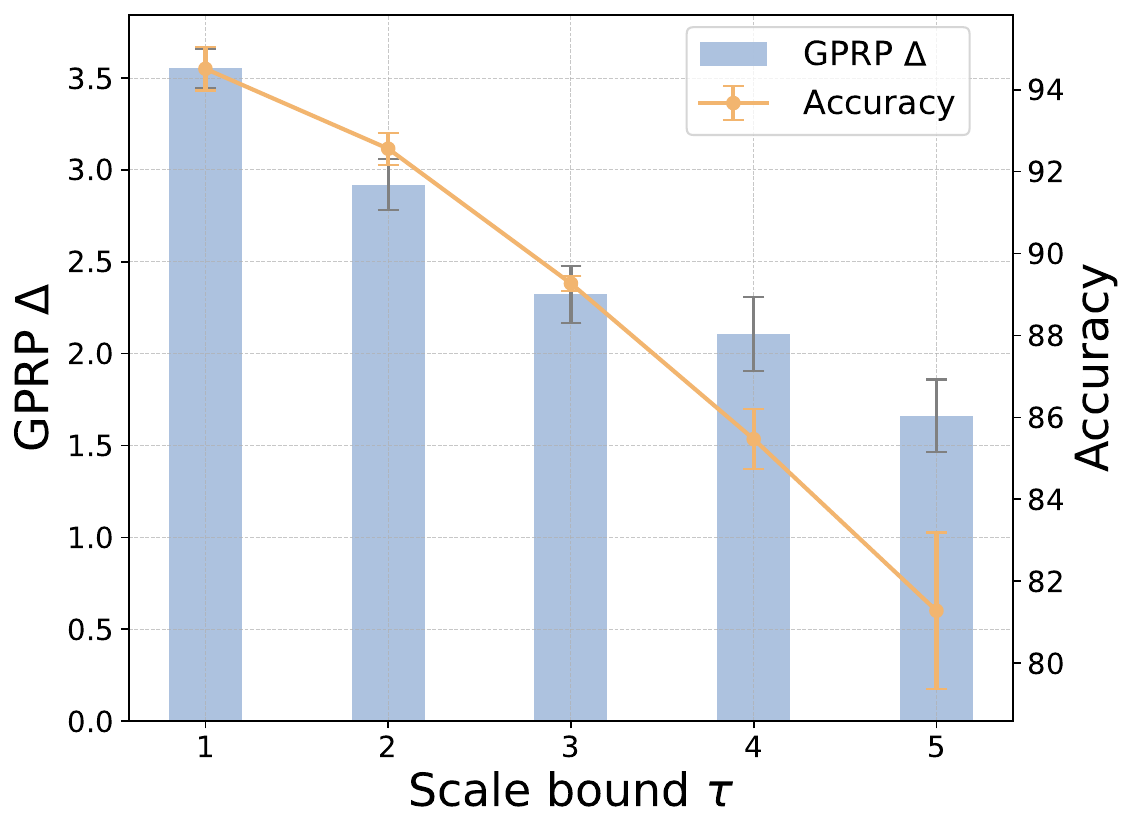}
    \caption{The results of the DP-SGD-S algorithm on the MNIST dataset using a CNN model under varying scale bounds.}
    \label{fig:scale}
    \end{minipage}
\end{figure}

\paragraph{Results across different privacy guarantees.} 
We conduct extensive experiments to compare the performance of three training algorithms under varying levels of privacy guarantees. In particular, we include $\epsilon = 1$ for each dataset when applying the differentially private training algorithms DP-SGD and DP-SGD-S.
Due to space constraints, we present only the results on the MNIST dataset with DP-SGD in the main text; comprehensive results for all datasets and methods are provided in App.~\ref{app:5.2}.
As illustrated in Fig.~\ref{fig:eps}, the stronger the model's privacy protection capability, the smaller the differences in privacy risk between groups. 
This is actually a rather intuitive conclusion. 
Imagine an extreme scenario where all data points in the model can ensure a privacy budget of 0; in this case, there would be no privacy risk differences between any points or groups. However, in practice, this is not feasible because the stricter the privacy budget, the less usable the model's prediction accuracy becomes. 
Therefore, our method manages to achieve more equitable privacy protection under the same privacy guarantees compared to DP-SGD, which is meaningful.


\paragraph{Results across different scale bounds.} 
We evaluate the impact of different scale bounds $\tau$ in DP-SGD-S on both accuracy and GPRP metrics on the MNIST dataset.
As shown in Fig.~\ref{fig:scale}, increasing the $\tau$ leads to a decrease in $\Delta$, as larger $\tau$ further limits the contribution of groups with larger norms to model updates.
However, this improvement in fairness comes with a trade-off in accuracy, likely due to the model's diminished ability to extract optimization information from these groups.

\subsection{Limitation and discussion}
\label{limitation}
While our proposed method, DP-SGD-S, demonstrates strong effectiveness in reducing disparities in group privacy risks across diverse datasets, it also presents certain limitations that warrant further discussion.
First, DP-SGD-S may introduce a slight drop in model accuracy compared to standard DP-SGD. This reflects a common trade-off where enhancing fairness in privacy protection may come at the expense of predictive performance. In practice, this trade-off is often acceptable, but it remains an important consideration in high-accuracy applications. Second, as mentioned in Sec.~\ref{sec:4.2}, DP-SGD-S requires a small portion of the overall privacy budget to protect the gradient statistics used during training.

Moreover, to provide a comprehensive understanding of group-level privacy behavior, we report detailed results in App.~\ref{app:5.3} across all datasets and settings. In over 90\% of the cases, DP-SGD-S does not increase the privacy risk for advantaged groups. Instead, privacy risks either decrease or remain stable for all groups, with more notable improvements in disadvantaged groups. This demonstrates that DP-SGD-S enhances privacy fairness without causing a “leveling down” effect, which is essential for real-world applications.
We also examine the impact of DP-SGD-S on conventional outcome fairness metrics, such as demographic parity and accuracy parity, with the results also provided in App.~\ref{app:outcome_fairness}. Our findings suggest that DP-SGD-S does not exacerbate outcome unfairness. However, any outcome unfairness already present in models trained with DP-SGD still exists under DP-SGD-S. Addressing these remaining limitations and developing algorithms that simultaneously promote both outcome fairness and privacy fairness is an important avenue for future research.

\section{Conclusion}
This work addresses a fundamental challenge at the intersection of fairness and privacy in AI systems: ensuring equitable privacy protection across different demographic groups. Our study makes two significant contributions to this emerging research direction. First, we develop a novel membership inference game-based privacy auditing mechanism that enables more rigorous measurement of group privacy risks. The empirical results prove that our method provides a more rigorous and reliable assessment of privacy risk disparities across groups while maintaining computational efficiency.
Second, to mitigate the identified privacy protection disparities, we propose an enhanced DP-SGD algorithm that incorporates an adaptive group-specific gradient clipping strategy. Through extensive experimental evaluation across diverse datasets, we demonstrate that our algorithm successfully reduces group privacy risk disparities while preserving model utility. 
This research advances both empirical understanding and practical implementation of fair privacy protection in ML systems, contributing to the broader goal of responsible AI deployment.



\bibliographystyle{plainnat}
\bibliography{paper}

@InProceedings{nasr2023tight,
  title={Tight auditing of differentially private machine learning},
  author={Nasr, Milad and Hayes, Jamie and Steinke, Thomas and Balle, Borja and Tram{\`e}r, Florian and Jagielski, Matthew and Carlini, Nicholas and Terzis, Andreas},
  booktitle={32nd USENIX Security Symposium (USENIX Security 23)},
  pages={1631--1648},
  year={2023}
}

@inproceedings{steinke2023privacy,
  title={Privacy auditing with one (1) training run},
  author={Steinke, Thomas and Nasr, Milad and Jagielski, Matthew},
  booktitle={Proceedings of the 37th International Conference on Neural Information Processing Systems},
  pages={49268--49280},
  year={2023}
}

@inproceedings{zanella2023bayesian,
  title={Bayesian estimation of differential privacy},
  author={Zanella-Beguelin, Santiago and Wutschitz, Lukas and Tople, Shruti and Salem, Ahmed and R{\"u}hle, Victor and Paverd, Andrew and Naseri, Mohammad and K{\"o}pf, Boris and Jones, Daniel},
  booktitle={International Conference on Machine Learning},
  pages={40624--40636},
  year={2023},
  organization={PMLR}
}

@inproceedings{jagielski2020auditing,
  title={Auditing differentially private machine learning: how private is private SGD?},
  author={Jagielski, Matthew and Ullman, Jonathan and Oprea, Alina},
  booktitle={Proceedings of the 34th International Conference on Neural Information Processing Systems},
  pages={22205--22216},
  year={2020}
}

@inproceedings{nasr2021adversary,
  title={Adversary instantiation: Lower bounds for differentially private machine learning},
  author={Nasr, Milad and Songi, Shuang and Thakurta, Abhradeep and Papernot, Nicolas and Carlin, Nicholas},
  booktitle={2021 IEEE Symposium on security and privacy (SP)},
  pages={866--882},
  year={2021},
  organization={IEEE}
}

@inproceedings{annamalai2024nearly,
  title={Nearly Tight Black-Box Auditing of Differentially Private Machine Learning},
  author={Meenatchi Sundaram Muthu Selva Annamalai and Emiliano De Cristofaro},
  booktitle={The Thirty-eighth Annual Conference on Neural Information Processing Systems},
  year={2024},
  note={\url{https://openreview.net/forum?id=cCDMXXiamP}}
}

@inproceedings{ye2022enhanced,
  title={Enhanced membership inference attacks against machine learning models},
  author={Ye, Jiayuan and Maddi, Aadyaa and Murakonda, Sasi Kumar and Bindschaedler, Vincent and Shokri, Reza},
  booktitle={Proceedings of the 2022 ACM SIGSAC Conference on Computer and Communications Security},
  pages={3093--3106},
  year={2022}
}

@inproceedings{shokri2017membership,
  title={Membership inference attacks against machine learning models},
  author={Shokri, Reza and Stronati, Marco and Song, Congzheng and Shmatikov, Vitaly},
  booktitle={2017 IEEE symposium on security and privacy (SP)},
  pages={3--18},
  year={2017},
  organization={IEEE}
}

@inproceedings{chang2021privacy,
  title={On the privacy risks of algorithmic fairness},
  author={Chang, Hongyan and Shokri, Reza},
  booktitle={2021 IEEE European Symposium on Security and Privacy (EuroS\&P)},
  pages={292--303},
  year={2021},
  organization={IEEE}
}

@inproceedings{yaghini2022disparate,
  title={Disparate Vulnerability to Membership Inference Attacks},
  author={Yaghini, M and Kulynych, B and Cherubin, G and Veale, M and Troncoso, C},
  booktitle={Proceedings on Privacy Enhancing Technologies},
  number={1},
  pages={460--480},
  year={2022}
}

@inproceedings{dwork2006our,
  title={Our data, ourselves: Privacy via distributed noise generation},
  author={Dwork, Cynthia and Kenthapadi, Krishnaram and McSherry, Frank and Mironov, Ilya and Naor, Moni},
  booktitle={24th Annual International Conference on the Theory and Applications of Cryptographic Techniques},
  pages={486--503},
  year={2006},
  organization={Springer}
}

@inproceedings{dwork2006calibrating,
  title={Calibrating noise to sensitivity in private data analysis},
  author={Dwork, Cynthia and McSherry, Frank and Nissim, Kobbi and Smith, Adam},
  booktitle={Theory of cryptography conference},
  volume={3},
  pages={265--284},
  year={2006},
  organization={Springer},
}

@article{ponomareva2023dp,
  title={How to {DP}-fy {ML}: A practical guide to machine learning with differential privacy},
  author={Ponomareva, Natalia and Hazimeh, Hussein and Kurakin, Alex and Xu, Zheng and Denison, Carson and McMahan, H Brendan and Vassilvitskii, Sergei and Chien, Steve and Thakurta, Abhradeep Guha},
  journal={Journal of Artificial Intelligence Research},
  volume={77},
  pages={1113--1201},
  year={2023}
}

@inproceedings{abadi2016deep,
  title={Deep learning with differential privacy},
  author={Abadi, Martin and Chu, Andy and Goodfellow, Ian and McMahan, H Brendan and Mironov, Ilya and Talwar, Kunal and Zhang, Li},
  booktitle={Proceedings of the 2016 ACM SIGSAC conference on computer and communications security},
  pages={308--318},
  year={2016}
}

@inproceedings{mironov2017renyi,
  title={R{\'e}nyi differential privacy},
  author={Mironov, Ilya},
  booktitle={2017 IEEE 30th computer security foundations symposium (CSF)},
  pages={263--275},
  year={2017},
  organization={IEEE}
}

@inproceedings{yeom2018privacy,
  title={Privacy risk in machine learning: Analyzing the connection to overfitting},
  author={Yeom, Samuel and Giacomelli, Irene and Fredrikson, Matt and Jha, Somesh},
  booktitle={2018 IEEE 31st computer security foundations symposium (CSF)},
  pages={268--282},
  year={2018},
  organization={IEEE}
}

@inproceedings{sablayrolles2019white,
  title={White-box vs black-box: Bayes optimal strategies for membership inference},
  author={Sablayrolles, Alexandre and Douze, Matthijs and Schmid, Cordelia and Ollivier, Yann and J{\'e}gou, Herv{\'e}},
  booktitle={International Conference on Machine Learning},
  pages={5558--5567},
  year={2019},
  organization={PMLR}
}

@inproceedings{xu2019achieving,
  title={Achieving differential privacy and fairness in logistic regression},
  author={Xu, Depeng and Yuan, Shuhan and Wu, Xintao},
  booktitle={Companion proceedings of The 2019 world wide web conference},
  pages={594--599},
  year={2019}
}

@inproceedings{jagielski2019differentially,
  title={Differentially private fair learning},
  author={Jagielski, Matthew and Kearns, Michael and Mao, Jieming and Oprea, Alina and Roth, Aaron and Sharifi-Malvajerdi, Saeed and Ullman, Jonathan},
  booktitle={International Conference on Machine Learning},
  pages={3000--3008},
  year={2019},
  organization={PMLR}
}

@inproceedings{ding2020differentially,
  title={Differentially private and fair classification via calibrated functional mechanism},
  author={Ding, Jiahao and Zhang, Xinyue and Li, Xiaohuan and Wang, Junyi and Yu, Rong and Pan, Miao},
  booktitle={Proceedings of the AAAI Conference on Artificial Intelligence},
  volume={34},
  number={01},
  pages={622--629},
  year={2020}
}

@inproceedings{tran2021differentially,
  title={Differentially private and fair deep learning: A lagrangian dual approach},
  author={Tran, Cuong and Fioretto, Ferdinando and Van Hentenryck, Pascal},
  booktitle={Proceedings of the AAAI Conference on Artificial Intelligence},
  volume={35},
  number={11},
  pages={9932--9939},
  year={2021}
}

@inproceedings{lowy2023stochastic,
  title={Stochastic differentially private and fair learning},
  author={Lowy, Andrew and Gupta, Devansh and Razaviyayn, Meisam},
  booktitle={Workshop on Algorithmic Fairness through the Lens of Causality and Privacy},
  pages={86--119},
  year={2023},
  organization={PMLR}
}

@inproceedings{bagdasaryan2019differential,
  title={Differential privacy has disparate impact on model accuracy},
  author={Bagdasaryan, Eugene and Poursaeed, Omid and Shmatikov, Vitaly},
  booktitle={Advances in neural information processing systems},
  publisher = {Curran Associates Inc.},
  pages={15479--15488},
  year={2019}
}

@inproceedings{xu2021removing,
  title={Removing disparate impact on model accuracy in differentially private stochastic gradient descent},
  author={Xu, Depeng and Du, Wei and Wu, Xintao},
  booktitle={Proceedings of the 27th ACM SIGKDD Conference on Knowledge Discovery \& Data Mining},
  pages={1924--1932},
  year={2021}
}

@inproceedings{tran2021fairness,
  title={Differentially private empirical risk minimization under the fairness lens},
  author={Tran, Cuong and Dinh, My and Fioretto, Ferdinando},
  booktitle={Advances in Neural Information Processing Systems},
  publisher = {Curran Associates Inc.},
  volume={34},
  pages={27555--27565},
  year={2021}
}

@inproceedings{esipova2022disparate,
  title={Disparate Impact in Differential Privacy from Gradient Misalignment},
  author={Maria S. Esipova and Atiyeh Ashari Ghomi and Yaqiao Luo and Jesse C Cresswell},
  booktitle={The Eleventh International Conference on Learning Representations },
  year={2023},
  note={\url{https://openreview.net/forum?id=qLOaeRvteqbx}}
}

@inproceedings{ekstrand2018privacy,
  title={Privacy for all: Ensuring fair and equitable privacy protections},
  author={Ekstrand, Michael D and Joshaghani, Rezvan and Mehrpouyan, Hoda},
  booktitle={Conference on fairness, accountability and transparency},
  pages={35--47},
  year={2018},
  organization={PMLR}
}

@article{huang2022overview,
  title={An overview of artificial intelligence ethics},
  author={Huang, Changwu and Zhang, Zeqi and Mao, Bifei and Yao, Xin},
  journal={IEEE Transactions on Artificial Intelligence},
  volume={4},
  number={4},
  pages={799--819},
  year={2023},
  publisher={IEEE}
}

@article{le2022survey,
  title={A survey on datasets for fairness-aware machine learning},
  author={Le Quy, Tai and Roy, Arjun and Iosifidis, Vasileios and Zhang, Wenbin and Ntoutsi, Eirini},
  journal={Wiley Interdisciplinary Reviews: Data Mining and Knowledge Discovery},
  volume={12},
  number={3},
  pages={e1452},
  year={2022},
  publisher={Wiley Online Library}
}

@inproceedings{zhang2017age,
  title={Age progression/regression by conditional adversarial autoencoder},
  author={Zhang, Zhifei and Song, Yang and Qi, Hairong},
  booktitle={Proceedings of the IEEE conference on computer vision and pattern recognition},
  pages={5810--5818},
  year={2017}
}

@article{opacus,
  title={Opacus: {U}ser-Friendly Differential Privacy Library in {PyTorch}},
  author={Ashkan Yousefpour and Igor Shilov and Alexandre Sablayrolles and Davide Testuggine and Karthik Prasad and Mani Malek and John Nguyen and Sayan Ghosh and Akash Bharadwaj and Jessica Zhao and Graham Cormode and Ilya Mironov},
  journal={arXiv preprint arXiv:2109.12298},
  year={2021}
}

@misc{lecun1998mnist,
  author       = {Yann LeCun and Corinna Cortes and Christopher J. C. Burges},
  title        = {The {MNIST} Database of Handwritten Digits},
  year         = {1998},
  howpublished = {\url{http://yann.lecun.com/exdb/mnist/}}
}

@article{pessach2022review,
  title={A review on fairness in machine learning},
  author={Pessach, Dana and Shmueli, Erez},
  journal={ACM Computing Surveys (CSUR)},
  volume={55},
  number={3},
  pages={1--44},
  year={2022},
  publisher={ACM New York, NY}
}







\clearpage
\appendix

\section{Supplementary Definitions and Algorithms}
\label{A}
\subsection{DP-SGD}
In our work, we concentrate on DP-SGD to uphold model privacy. Building upon SGD, the fundamental method for training a model $f$ with parameters $w$ by minimizing the empirical loss function $\ell(\hat{y}, y)$ for prediction $\hat{y}$ and label $y$, DP-SGD (as illustrated in Algo.~\ref{algo:DP-SGD}) integrates gradient clipping and noise addition for achieving the $(\epsilon, \delta)$-DP guarantees. In Algo.~\ref{algo:DP-SGD}, during each epoch, per-sample gradients $g_i$ are computed (Line 5). Since these gradients typically have unbounded sensitivity, they are clipped to ensure their norm does not exceed the hyperparameter $C$ (Line 6). The clipped gradients are then aggregated and Gaussian noise is added to yield $\tilde{g}$ (Line 8). $\tilde{g}$ is subsequently scaled by the learning rate $\eta$ and utilized for parameter update (Line 9). The final accumulated $(\epsilon, \delta)$, which is calculated by \textit{Rényi} differential privacy (RDP)~\cite{mironov2017renyi} and the moment accounting mechanism proposed by~\cite{abadi2016deep}, quantifies the privacy protection ability. 

\begin{algorithm}[htbp]
    \caption{DP-SGD~\cite{abadi2016deep}}
    \label{algo:DP-SGD}
    \begin{algorithmic}[1]
        \REQUIRE Training dataset $D=\{ (\mathbf{x}_{i},y_{i}) \} _{i=1}^{N}$, 
        the parameterized model $f_{w}(\cdot)$, loss function $\ell(\hat{y}, y)$,
        iterations $T$, batch size $b$, learning rate $\eta$, noise scale $\sigma$, clipping bound $C$.
        \STATE {Initialize $w^{(0)}$ randomly.}
        \FOR {$ t = 0, ..., T-1 $} 
            \STATE {Sample a batch $B$ from $D$ with probability $b/N$.}
                \FOR { ${i} \in B$}
                    \STATE{ $ g_{i} \leftarrow \nabla\ell(f_{{w^{(t)}}}(\mathbf x_{i}), y_{i}) $ }
                    \STATE{ $ \bar{g}_{i} \leftarrow g_{i} \cdot min(1, \frac{C}{\|g_{i}\|_2}) $ }
                \ENDFOR
                \STATE{ $\tilde{g} \leftarrow \frac{1}{b} \left ( \sum_{i \in B}\bar{g}_{i} + \mathcal{N}(0, \sigma^2 C^2 \mathbf{I}) \right )$ }
                \STATE{ $w^{(t+1)} \leftarrow w^{(t)} - \eta \tilde{g} $ }
        \ENDFOR
        \ENSURE Model $f_{w^{(T)}}(\cdot)$ and accumulated $(\epsilon, \delta)$.
    \end{algorithmic} 
\end{algorithm}

\subsection{Definitions of Membership Inference Games}
\begin{definition}[Average-case Membership Inference Game~\cite{yeom2018privacy}] Let $\Omega$ denotes the underlying population data pool, $\mathcal{M}$ the training algorithm, and $\mathcal{A}$ the inference algorithm. 
We assume that the challenger samples $n$ i.i.d. records from $\Omega$ to construct the training dataset $D$.
    \begin{itemize}
        \item [1)]  The challenger trains a target model $f \gets \mathcal{M}(D)$.
        \item [2)]  The challenger randomly selects a record $z_0 \gets \Omega$ and a record $z_1 \sim D$, ensuring that $z_0 \notin D$ .
        \item [3)]  The challenger flips a fair coin $b \in \{0, 1\}$, and sends the target model and target record $(f, z_b)$ to the adversary.
        \item [4)] The adversary, with access to the target model, outputs a guess $\hat{b} \leftarrow \mathcal{A}(f, z_b)$.
        \item [5)]  The game outputs 1 (success) if $\hat{b} = b$, and 0 otherwise.
    \end{itemize}
\label{def:std_mig}
\end{definition}

\begin{definition}[Worst-case Membership Inference Game~\cite{ye2022enhanced}] 
\textnormal{}
    \begin{itemize}[itemsep=0pt]
        \item [1)]  The challenger samples a fixed record $z \sim D$, and trains a model $f_0 \leftarrow \mathcal{M}(D \setminus z)$.
        \item [2)]  The challenger trains a model $f_1 \gets \mathcal{M}(D)$.
        \item [3)]  The challenger flips a fair coin $b \in \{0, 1\}$, and sends the target model and record $(f_b, z)$ to the adversary.
        \item [4)]  The adversary, with access to the target model, outputs a guess $\hat{b} \leftarrow \mathcal{A}(f_b, z)$.
        \item [5)]  The game outputs 1 (success) if $\hat{b} = b$, and 0 otherwise.
    \end{itemize}
\label{def:worst_mig}
\end{definition}

\subsection{Privacy Audting by Different Attacks}
\paragraph{Privacy auditing by average-case attacks.}
We introduce existing algorithms that use average-case attacks for privacy auditing (PA-ACAs). 
Specifically, one approach conducts privacy auditing using the global attack (PA-GA)~\cite{yaghini2022disparate}, which is detailed in Algo.~\ref{algo:pa-ga}. In Algo.~\ref{algo:pa-ga}, a single threshold $\beta$ is determined based on the overall behavior of all auditing samples.
Another approach performs privacy auditing through the group-based attack (PA-GBA)~\cite{chang2021privacy}, as described in Algo.~\ref{algo:pa-gba}. PA-GBA provides the adversary with background knowledge about which group $g_i$ each data point $(\mathbf{x}_{i},y_{i})$ belongs to. It then determines $K$ thresholds $\beta^k$ based on the behavior of all auditing samples within each group, where $K$ represents the number of groups.
As illustrated in Algo.~\ref{algo:pa-ga} and Algo.~\ref{algo:pa-gba}, a single execution of the attack generates a prediction for the membership status of each data point in the auditing dataset. To evaluate the privacy risk associated with individual data points more comprehensively, researchers typically conduct multiple iterations of privacy auditing using average-case attacks~\cite{chang2021privacy}. Each iteration yields new results, which are aggregated to estimate the likelihood of a data point being accurately identified as either a member or a non-member.
\begin{algorithm}[htbp]
    \caption{PA-GA~\cite{yaghini2022disparate}}
    \label{algo:pa-ga}
    \begin{algorithmic}[1]
        \REQUIRE Training dataset $D=\{ (\mathbf{x}_{i},y_{i}) \} _{i=1}^{n}$, auditing dataset $Z=\{ z_i=(\mathbf{x}_{i},y_{i}) \} _{i=1}^{m}$, loss function $\ell(\hat{y}, y)$, optimal threshold $\beta$.
        \STATE {Initialize outputs $O \leftarrow [\ ]$, membership status $H \leftarrow [\ ]$, and membership guesses $G \leftarrow [\ ]$.}
        \STATE{ $ f \gets \mathcal{M}(D) $ }
        \FOR {$ i = 1, \dots, m $} 
            \STATE{ $ O[i] \gets \ell(f(\mathbf{x_i}), y_i) $ }
            \STATE{ $ H[i] \gets \begin{cases} 
                    1 & \text{if } z_i \in D \\
                    0 & \text{otherwise} 
                    \end{cases} $ }
        \ENDFOR
        \STATE {$G \gets [1 \{ O[i] \geq \beta \} \text{ for } i = 1, \dots, m ]$}.
        \ENSURE Membership status $H$ and guesses $G$.
    \end{algorithmic} 
\end{algorithm}

\begin{algorithm}[htbp]
    \caption{PA-GBA~\cite{chang2021privacy}}
    \label{algo:pa-gba}
    \begin{algorithmic}[1]
        \REQUIRE Training dataset $D=\{ (\mathbf{x}_{i},y_{i},g_{i}) \} _{i=1}^{n}$, auditing dataset $Z=\{ z_i=(\mathbf{x}_{i},y_{i},g_{i}) \} _{i=1}^{m}$, loss function $\ell(\hat{y}, y)$, optimal threshold $\{\beta^k\}_{i=1}^{K}$.
        \STATE {Initialize outputs $O \leftarrow [\ ]$, membership status $H \leftarrow [\ ]$, and membership guesses $G \leftarrow [\ ]$.}
        \STATE{ $ f \gets \mathcal{M}(D) $ }
        \FOR {$ i = 1, \dots, m $} 
            \STATE{ $ O[i] \gets \ell(f(\mathbf{x_i}), y_i) $ }
            \STATE{ $ H[i] \gets \begin{cases} 
                    1 & \text{if } z_i \in D \\
                    0 & \text{otherwise} 
                    \end{cases} $ }
        \ENDFOR
        \STATE {$G \gets [1 \{ O[i] \geq \beta^{g_i} \} \text{ for } i = 1, \dots, m ]$}.
        \ENSURE Membership status $H$ and guesses $G$.
    \end{algorithmic} 
\end{algorithm}

\paragraph{Privacy auditing by LOOA.}
In recent years, numerous studies have focused on using privacy auditing to evaluate the differential privacy (DP) guarantees of the DP-SGD algorithm~\cite{nasr2023tight,steinke2023privacy,zanella2023bayesian,jagielski2020auditing,annamalai2024nearly,nasr2021adversary}. These studies aim to bridge the gap between theoretical guarantees and practical performance, offering empirical insights into the actual privacy leakage in real-world deployments.
A common approach in these studies is Privacy Auditing via the Leave-One-Out Attack (PA-LOOA), as outlined in Algo.~\ref{algo:pa-looa}. 
The algorithm iteratively assesses the impact of including or excluding a specific data record $z$—often crafted as a worst-case scenario for auditing DP-SGD—within the training dataset $D$ (Lines 2--8).
For each repetition, the framework trains two models: $f_0$, using the modified dataset $D \setminus z$, and $f_1$, using the original dataset $D$ (Lines 3--4). The outputs of these models are recorded, and the membership status of the data record is tracked (Lines 5--7). 
Based on these outputs and the membership status, attack scores are computed to estimate the likelihood of the record's inclusion (Line 9). Finally, assuming an optimal adversary conducting the attack, an optimal threshold is applied to infer whether the record $z$ was part of the training dataset (Line 10).

In our work, we focus on evaluating the empirical privacy leakage risk of each individual data record within the training dataset, rather than the worst guarantees of a mechanism in DP auditing studies.
\begin{algorithm}[htbp]
    \caption{PA-LOOA~\cite{nasr2023tight,steinke2023privacy,zanella2023bayesian,jagielski2020auditing,annamalai2024nearly,nasr2021adversary}}
    \label{algo:pa-looa}
    \begin{algorithmic}[1]
        \REQUIRE Training dataset $D=\{ (\mathbf{x}_{i},y_{i}) \} _{i=1}^{n}$, auditing data record $z=(\mathbf{x}, y)$, loss function $\ell(\hat{y}, y)$, number of repetitions $R$, optimal threshold $\beta$.
        \STATE {Initialize outputs $O \leftarrow [\ ]$, membership status $H \leftarrow [\ ]$, and membership guesses $G \leftarrow [\ ]$.}
        \FOR {$ r = 1, ..., R $} 
            \STATE{ $ f_0 \gets \mathcal{M}(D\setminus z) $ }
            \STATE{ $ f_1 \gets \mathcal{M}(D) $ }
            \STATE{ $ O[2r-1] \gets \ell(f_0(\mathbf{x}), y) $ }
            \STATE{ $ O[2r] \gets \ell(f_1(\mathbf{x}), y) $ }
            \STATE{ $ H \gets H + [0, 1] $ }
        \ENDFOR
        \STATE {$G \gets [1 \{ O[r] \geq \beta \} \text{ for } r = 1, \dots, 2R]$}.

        \ENSURE Membership status $H$ and guesses $G$.
    \end{algorithmic} 
\end{algorithm}

\paragraph{Privacy auditing by ALOOA.}
In each iteration, $m$ audit samples are randomly and independently assigned inclusion or exclusion statuses for training (Line~3). Based on this membership status set, the training dataset $D \setminus \{z_i \mid h_i = 1\}$ is constructed, and a model $f_1$ is trained. Similarly, a model $f_0$ is trained using the inverse of this state set (Lines~4--5). The membership states for each audit record, indicating whether it was used in training, are then recorded (Lines~6--7). Subsequently, the output for each audit sample is logged (Lines~8--11). Based on these outputs and the true membership statuses, membership states are inferred through the outputs and an optimal threshold (Line~13--15).

\begin{algorithm}[htbp]
    \caption{PA-ALOOA}
    \label{algo:pa-alooa}
    \begin{algorithmic}[1]
        \REQUIRE Training dataset $D=\{ (\mathbf{x}_{i},y_{i}) \} _{i=1}^{n}$, Auditing dataset $Z=\{ z_i=(\mathbf{x}_{i},y_{i}) \} _{i=1}^{m}$, loss function $\ell(\hat{y}, y)$, number of repetitions $R$, optimal thresholds $\{\beta_i\}_{i=1}^{m}$.
        \STATE {Initialize outputs $O \leftarrow [\ ]$, membership status $H \leftarrow [\ ]$, and membership guesses $G \leftarrow [\ ]$.}
        \FOR {$ r = 1, ..., R $} 
            \STATE{Randomly generate membership statuses $\{ h_i \}_{i=1}^{m}$, where $ h_i \in \{0, 1\} $ for each $ z_i $.}
            \STATE{ $ f_{0} \gets \mathcal{M}(D \setminus \{z_i \mid h_i = 0\}) $ }
            \STATE{ $ f_{1} \gets \mathcal{M}(D \setminus \{z_i \mid h_i = 1\}) $ }
            \STATE{ $ H[2r-1] \gets \{ h_i \}_{i=1}^{m}$}
            \STATE{ $ H[2r] \gets \{ \sim h_i \}_{i=1}^{m}$}
            \FOR {$ i = 1, ..., m $} 
                \STATE{ $ O[2r-1][i] \gets \ell(f_0(\mathbf{x_i}), y_i) $ }
                \STATE{ $ O[2r][i] \gets \ell(f_1(\mathbf{x_i}), y_i) $ }
            \ENDFOR
        \ENDFOR
        \FOR {$ i = 1, ..., m $} 
        \STATE {$G[i] \gets [1 \{ O[r][i] \geq \beta_i \} \text{ for } r = 1, \dots, 2R]$}.
        \ENDFOR
        \ENSURE Membership status $H$ and guesses $G$.
    \end{algorithmic} 
\end{algorithm}

\section{Comparison with PA-LOOA}
\label{app:cop_pa-looa}
We conduct experiments using the MNIST dataset and CNN models trained with the SGD optimizer. Detailed hyperparameter settings are provided in App.~\ref{app:setup}. In the main paper, we set $m = 600$ for both PA-LOOA and PA-ALOOA. Here, we further evaluate a different configuration: $m = 600$ for PA-LOOA and $m = n$ for PA-ALOOA, where $m = n$ better reflects realistic auditing scenarios.

\begin{figure}[h]
    \centering
    \includegraphics[width=0.75\linewidth]{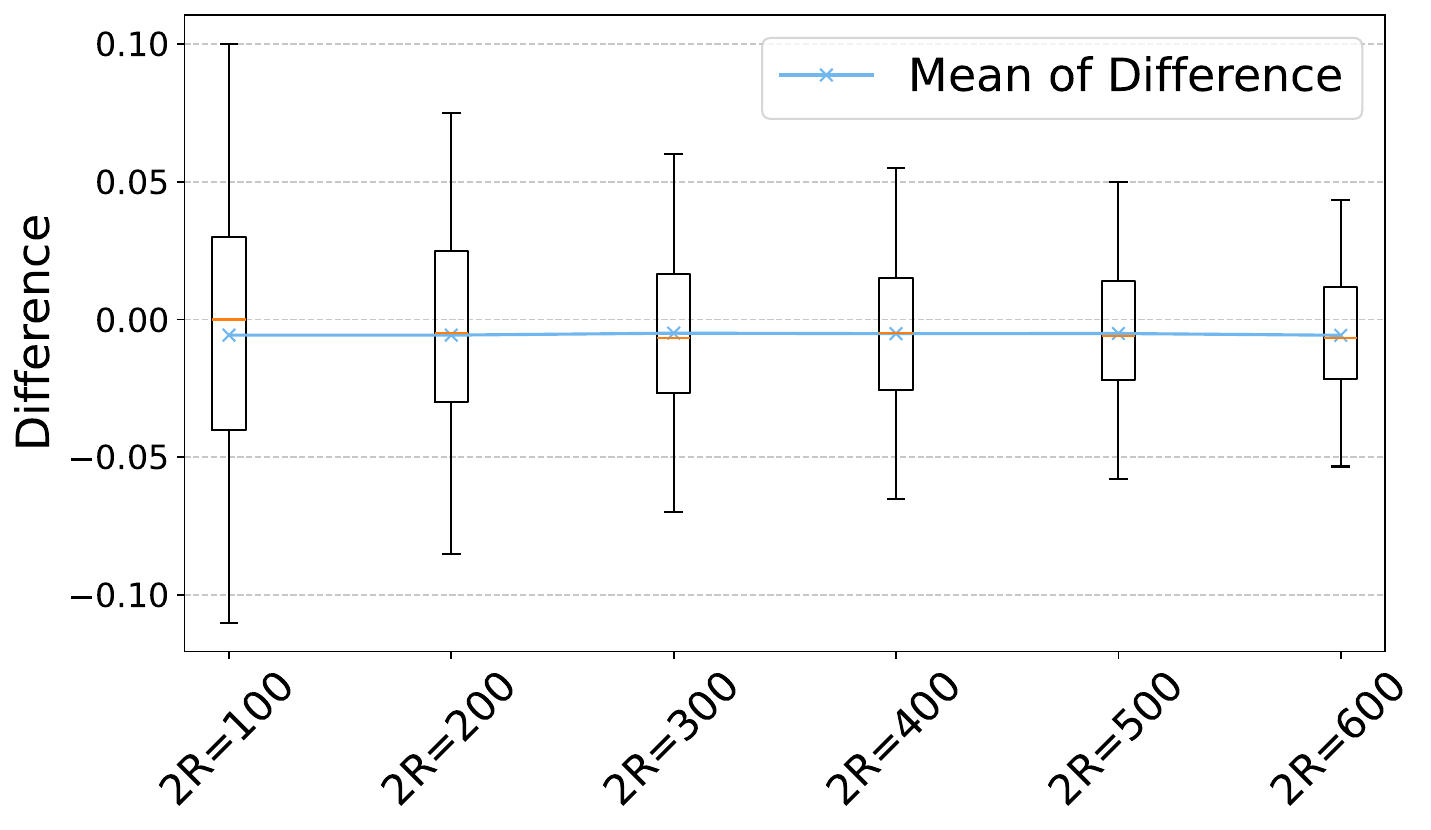}
    \caption{The horizontal axis represents the number of random experiments for a single audit, while the vertical axis represents the signed difference in auditing performance between the two attacks for each audited sample.}
\label{model_compare_signed}
\end{figure}

\begin{figure}[H]
    \centering
    \begin{subfigure}[b]{0.48\textwidth} 
        \centering
        \includegraphics[width=0.9\linewidth]{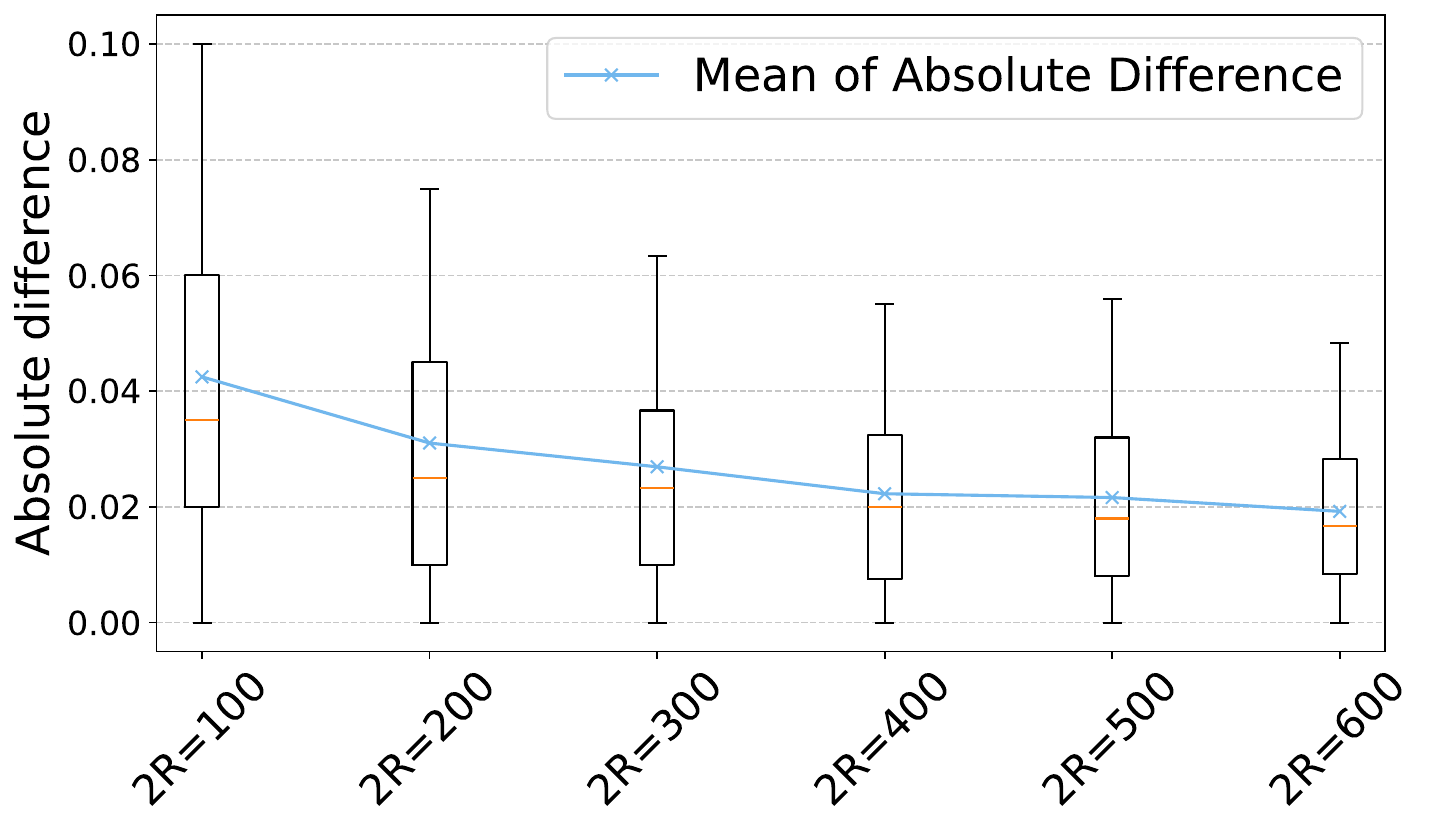} 
        \caption{Comparison at the individual level.}
        \label{app:model_compare}
    \end{subfigure}%
    \hfill
    \begin{subfigure}[b]{0.48\textwidth} 
        \centering
        \includegraphics[width=0.9\linewidth]{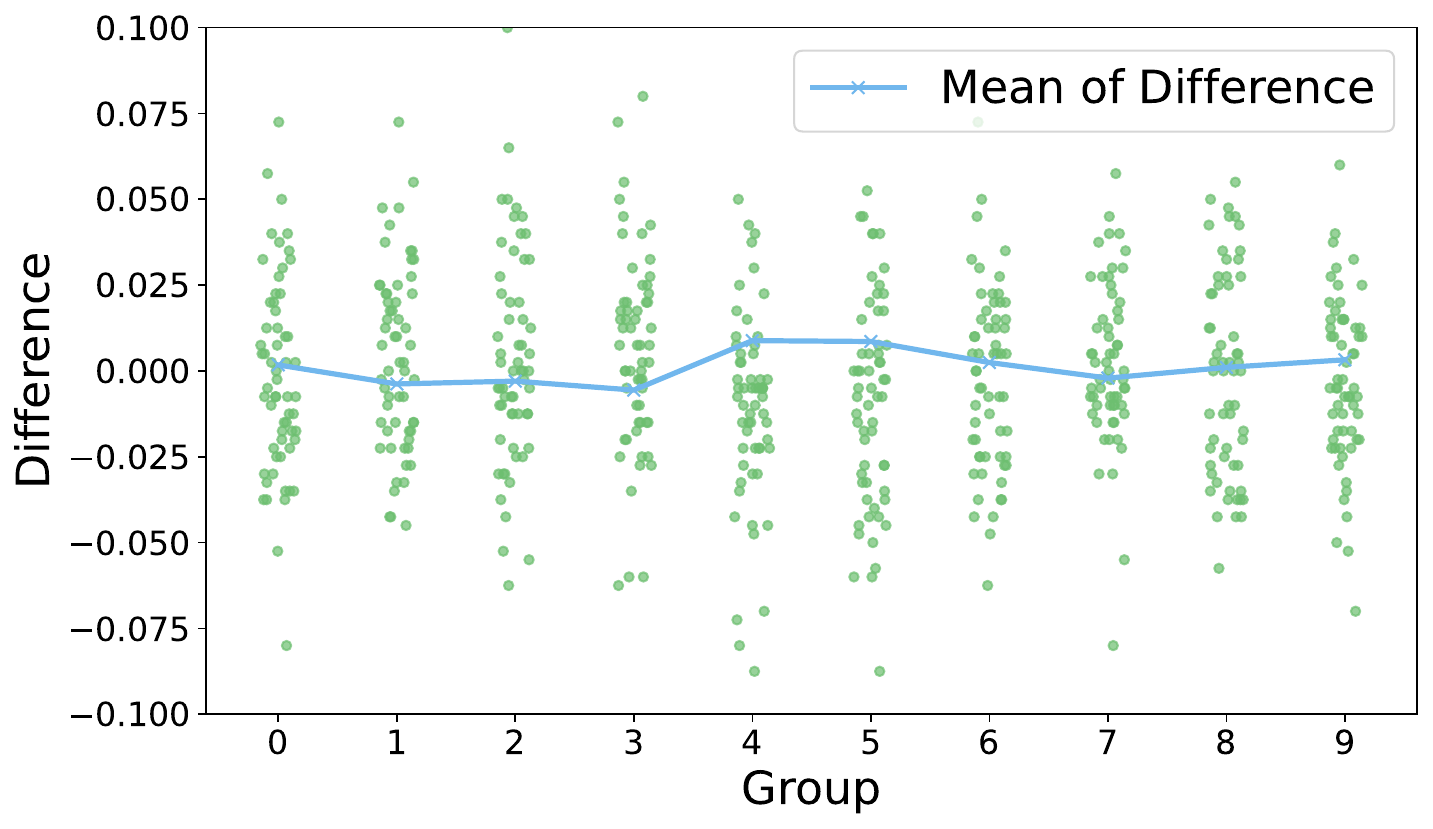} 
        \caption{Comparison at the group level.}
        \label{app:group_compare}
    \end{subfigure}%
    \caption{Left: The horizontal axis represents the number of random experiments for a single audit, while the vertical axis represents the absolute difference in auditing performance between the two attacks for each audited sample. Right: The horizontal axis represents different groups of MNIST dataset, while the vertical axis indicates the performance difference between PA-LOOA and PA-ALOOA for individual data points in each group at $2R=400$.}
    \label{app:comparison_looa}
\end{figure}

As shown in Fig.~\ref{app:comparison_looa}, the results under this setting are consistent with those reported in the main paper. PA-ALOOA maintains comparable auditing performance while significantly reducing computational cost. Moreover, it remains effective and reliable for measuring group-level privacy risks, further supporting its applicability in real-world deployments.

\section{Comparison with PA-ACAs}
\label{app:cop_pa-acas}
\begin{figure}[H]
    \centering
        \begin{subfigure}[b]{0.3\textwidth}
        \centering
        \includegraphics[width=\linewidth]{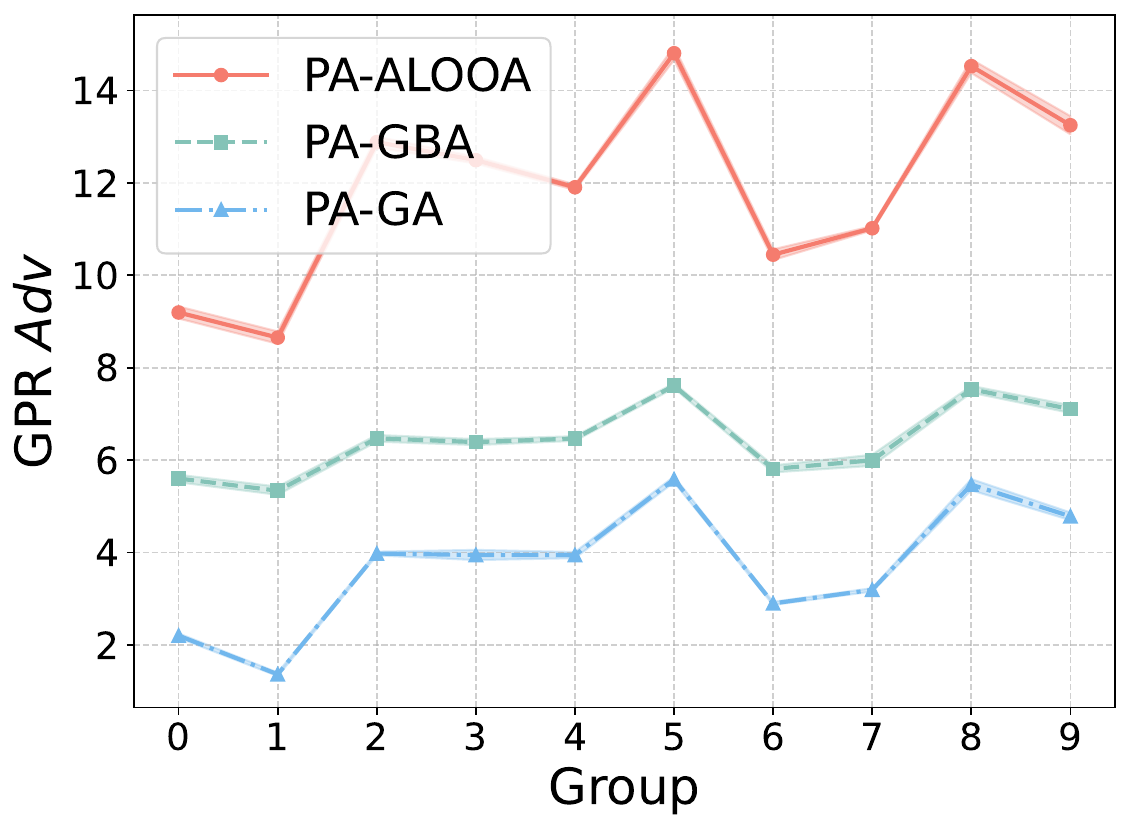} 
        \caption{LR.}
    \end{subfigure}%
    \hfill
    \begin{subfigure}[b]{0.3\textwidth}
        \centering
        \includegraphics[width=\linewidth]{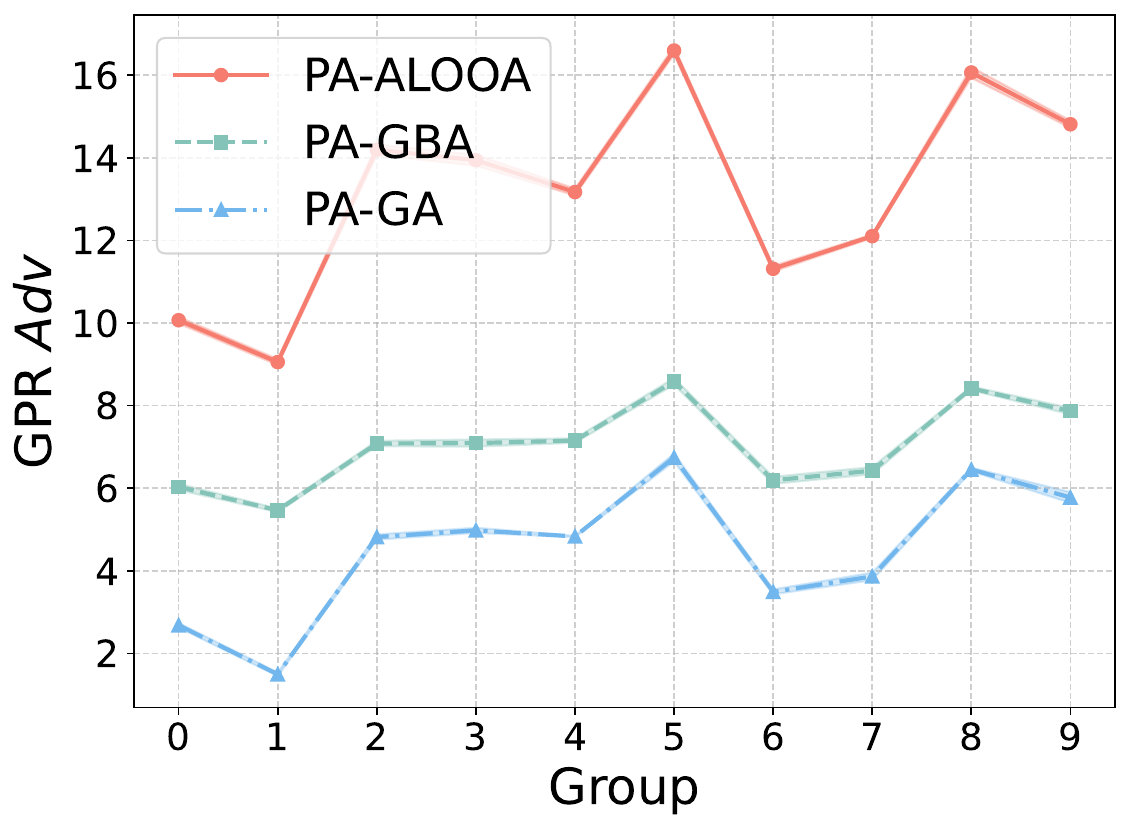} 
        \caption{MLP.}
    \end{subfigure}%
    \hfill
    \begin{subfigure}[b]{0.3\textwidth}
        \centering
        \includegraphics[width=\linewidth]{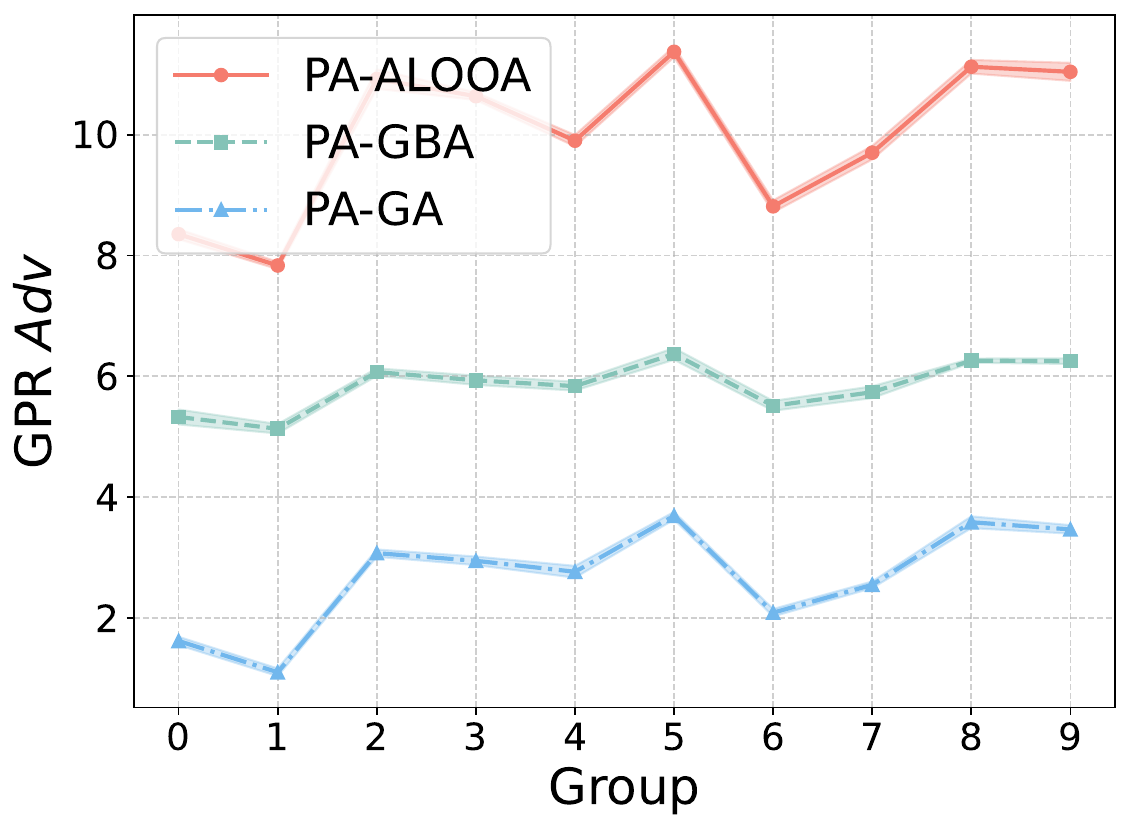} 
        \caption{CNN.}
    \end{subfigure}
    \caption{The comparison of GPR value across three model types—Logistic Regression (LR), Multilayer Perceptron (MLP), and CNN—trained on the MNIST dataset using DP-SGD algorithm with $\epsilon=10$. The x-axis represents the groups, and the y-axis shows the corresponding GPR value at $2R=400$.}
    \label{app3.3:mnist}
\end{figure}

\begin{figure}[H]
    \centering
        \begin{subfigure}[b]{0.2\textwidth}
        \centering
        \includegraphics[width=\linewidth]{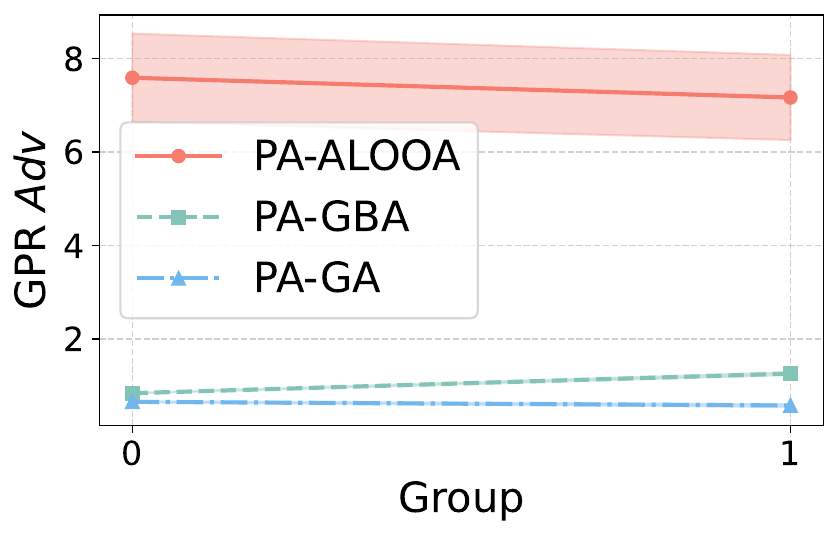} 
        \caption{Adult.}
    \end{subfigure}%
    \hfill
    \begin{subfigure}[b]{0.2\textwidth}
        \centering
        \includegraphics[width=\linewidth]{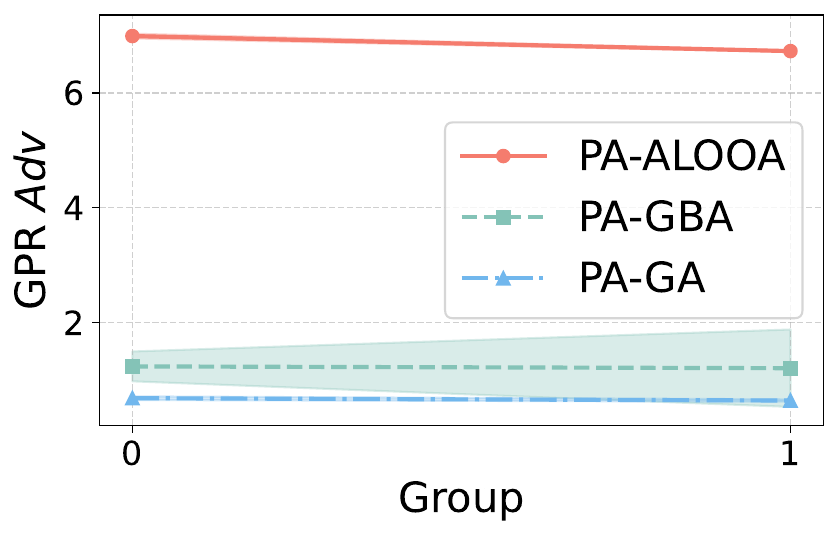} 
        \caption{Bank.}
    \end{subfigure}%
    \hfill
    \begin{subfigure}[b]{0.2\textwidth}
        \centering
        \includegraphics[width=\linewidth]{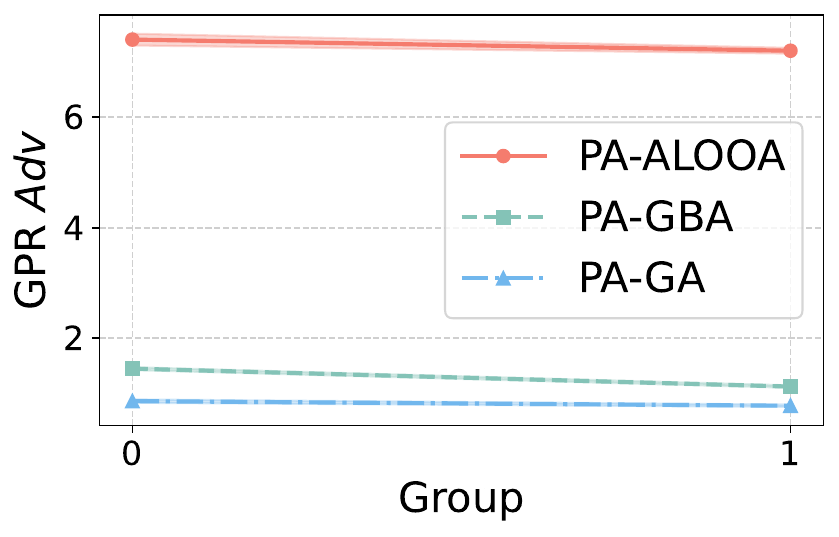} 
        \caption{Credit.}
    \end{subfigure}%
    \centering
        \begin{subfigure}[b]{0.2\textwidth}
        \centering
        \includegraphics[width=\linewidth]{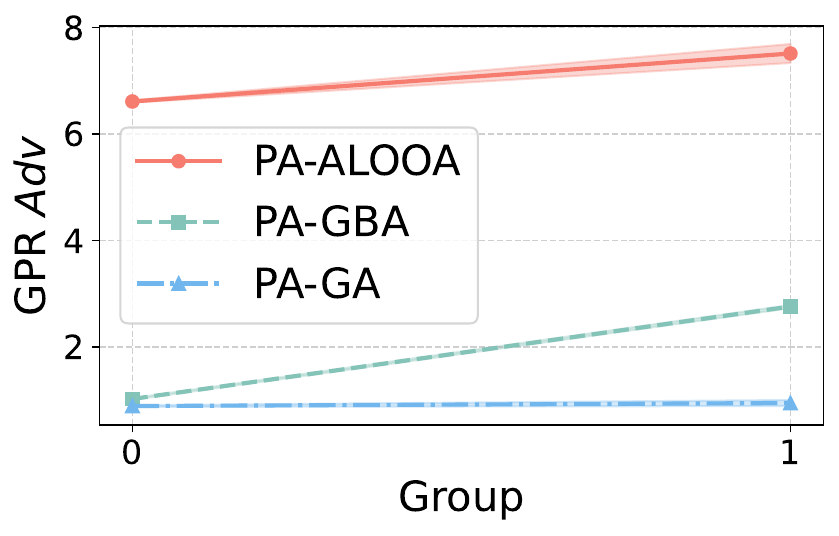} 
        \caption{Law.}
    \end{subfigure}%
    \hfill
    \begin{subfigure}[b]{0.2\textwidth}
        \centering
        \includegraphics[width=\linewidth]{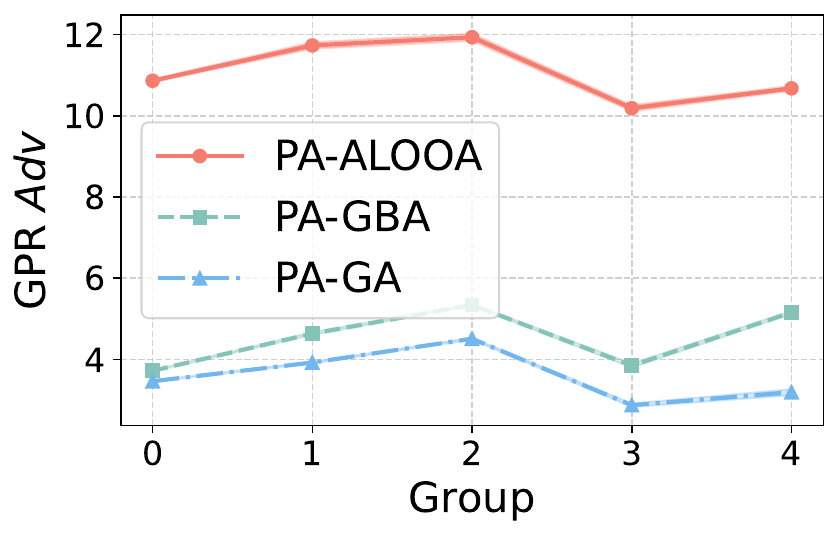} 
        \caption{UTKFace.}
    \end{subfigure}%
    \caption{The comparison of GPR value across three model types—LR, MLP, and CNN—trained on the fairness-related datasets using SGD algorithm. The x-axis represents the groups, and the y-axis shows the corresponding GPR value at $2R=400$.}
    \label{app3.3:fair_dataset_regular}
\end{figure}

\begin{figure}[H]
    \centering
        \begin{subfigure}[b]{0.2\textwidth}
        \centering
        \includegraphics[width=\linewidth]{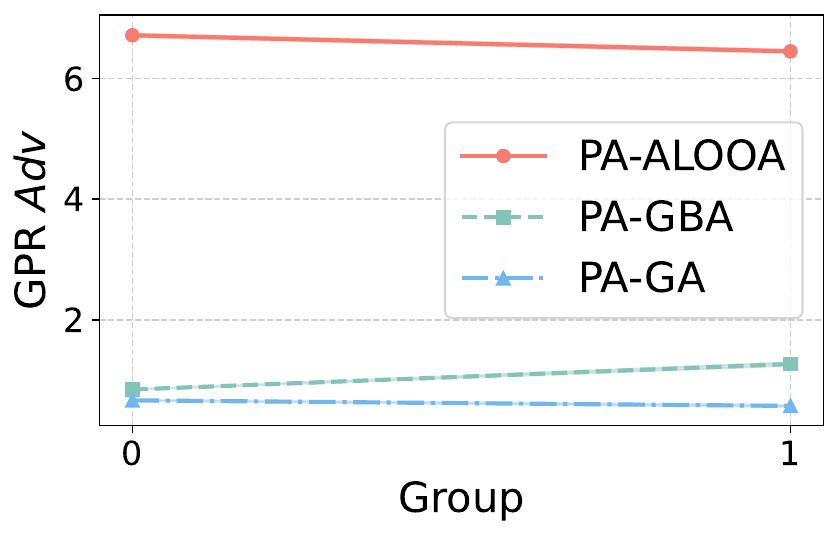} 
        \caption{Adult.}
    \end{subfigure}%
    \hfill
    \begin{subfigure}[b]{0.2\textwidth}
        \centering
        \includegraphics[width=\linewidth]{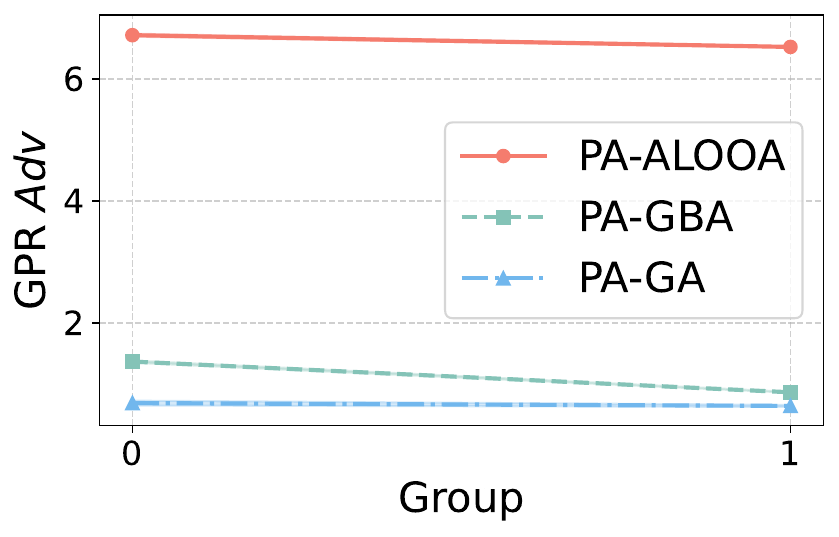} 
        \caption{Bank.}
    \end{subfigure}%
    \hfill
    \begin{subfigure}[b]{0.2\textwidth}
        \centering
        \includegraphics[width=\linewidth]{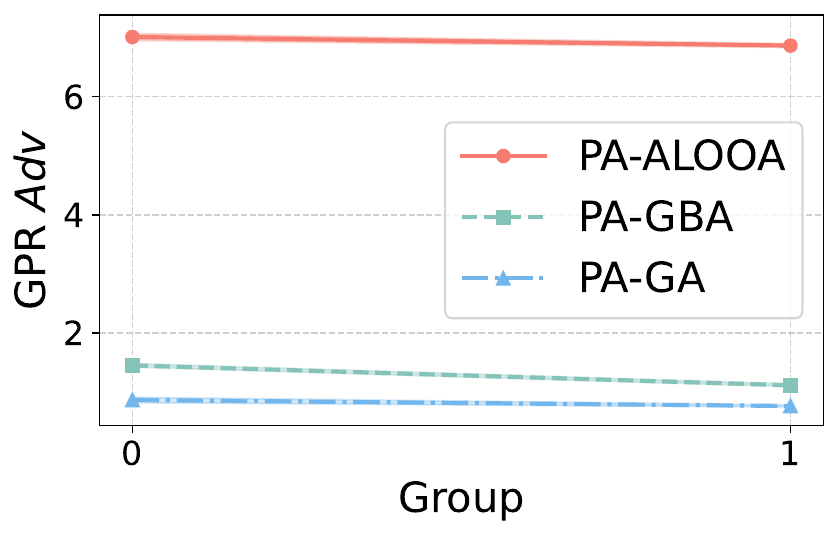} 
        \caption{Credit.}
    \end{subfigure}%
    \centering
        \begin{subfigure}[b]{0.2\textwidth}
        \centering
        \includegraphics[width=\linewidth]{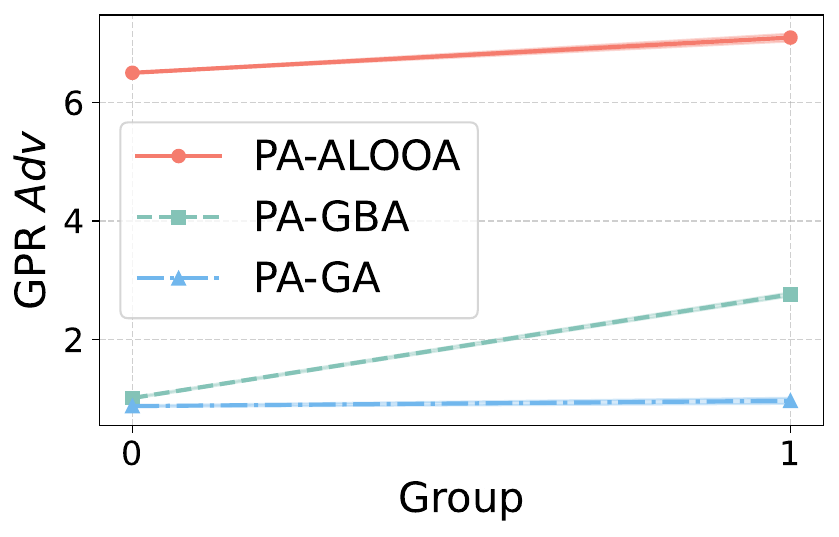} 
        \caption{Law.}
    \end{subfigure}%
    \hfill
    \begin{subfigure}[b]{0.2\textwidth}
        \centering
        \includegraphics[width=\linewidth]{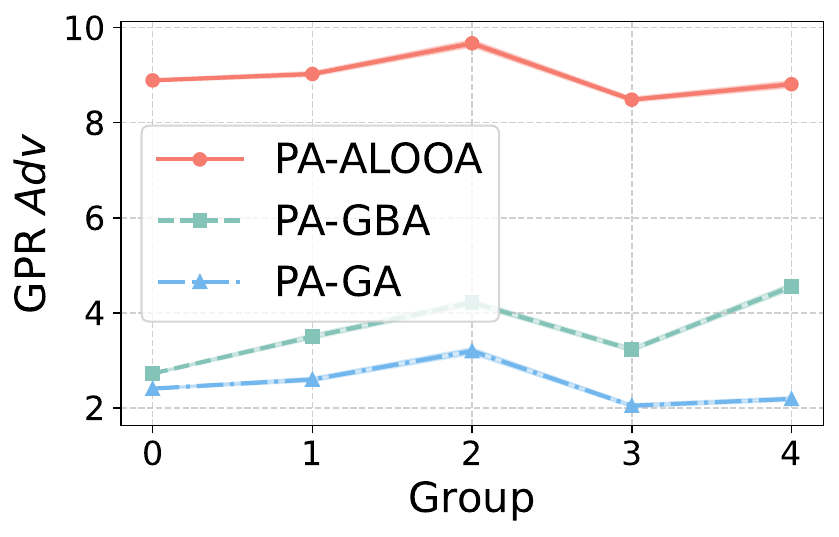} 
        \caption{UTKFace.}
    \end{subfigure}%
    \caption{The comparison of GPR value across three model types—LR, MLP, and CNN—trained on the fairness-related datasets using DP-SGD algorithm with $\epsilon=10$. The x-axis represents the groups, and the y-axis shows the corresponding GPR value at $2R=400$.}
    \label{app3.3:fair_dataset_dpsgd}
\end{figure}

We provide additional results on six datasets: MNIST, Adult, Bank, Credit, Law, and UTKFace. We use both standard SGD and DP-SGD training algorithms. The models employed include LR, MLP, and CNN. Detailed experimental settings are provided in App.~\ref{app:setup}.
The complete results are shown in Figs. \ref{app3.3:mnist}, \ref{app3.3:fair_dataset_regular}, and \ref{app3.3:fair_dataset_dpsgd}, respectively. As clearly shown in the figures, the conclusions remain consistent with those discussed in the main text.

\section{Experimental Details and Results}
\label{app:add_set&res}
\subsection{Details of Experimental Setup}
\label{app:setup}
\paragraph{Datasets.} 
\label{app:datasets_setup}
The MNIST dataset~\cite{lecun1998mnist} comprises 60,000 training and 10,000 testing samples, with each sample being a 28×28 grayscale image of a handwritten digit from 0 to 9, spanning ten classes. Due to computational limitations and to improve training efficiency, we randomly select 1,000 samples per class from the original dataset, resulting in a balanced dataset of 10,000 samples used in our experiments. In our study, we treat the classification label (i.e., the digit class) as a proxy for the demographic group to analyze group-specific privacy risks.

For the tabular fairness-related datasets, Adult, Bank, Credit, and Law~\cite{le2022survey}%
\footnote{Dataset can be download from \url{https://github.com/tailequy/fairness_dataset/tree/main/experiments/data}}, the detailed information is shown in Tab.~\ref{tab:tabular}. For the image-based fairness-related dataset, UTKFace~\cite{zhang2017age}, we conduct evaluation after data cleaning and preprocessing.%
\footnote{Dataset can be downloaded from \url{https://www.kaggle.com/datasets/nipunarora8/age-gender-and-ethnicity-face-data-csv/data?select=age_gender.csv}}
The final dataset consists of 27,305 grayscale facial images of size 1×48×48. In our experiments, we treat ethnicity as the protected attribute and gender as the prediction label. The ethnicity attribute includes five classes, with the number of samples per class being 10,078; 4,526; 3,434; 3,975; and 1,692, respectively.

\begin{table}[h!]
  \begin{center}
    \caption{The information of experimental datasets. Here, $0$ represents the advantaged group, and $1$ represents the disadvantaged group.}
    \label{tab:tabular}
    \begin{tabular}{cccc} 
        \toprule
        \textbf{Dataset} & \textbf{$\#$Instances(cleaned)} & \textbf{Class ratio$(0 : 1)$} & \textbf{Sensitive attribute} \\
        \midrule
        Adult & 45,222 & 2.09 : 1 & Gender \\
        Bank & 40,004 & 2.13 : 1 & Marital \\
        Credit & 30,000 & 1.52 : 1 & Sex \\
        Law & 20,798 & 5.29 : 1 & Race \\
        \bottomrule
    \end{tabular}
  \end{center}
\end{table}

\paragraph{Training algorithms.} 
\label{app:algorithms_setup}
Our study compares three training algorithms: standard SGD, DP-SGD, and our proposed DP-SGD-S. DPML algorithms (i.e., DP-SGD and DP-SGD-S) are implemented using the Opacus library~\cite{opacus}. 
For the SGD algorithm, all datasets are trained using the SGD optimizer with a learning rate of 0.1.
For DPML algorithms, the privacy hyperparameters $(\epsilon, \delta)$ are configured with $(10, 1\mathrm{e}{-5})$ and $(1, 1\mathrm{e}{-5})$. For DP-SGD-S, the default scale bound $\tau$ is set to 2. When $\epsilon = 10$, all datasets use the SGD optimizer with a learning rate of 0.1, and the clipping bound is set to 10. When $\epsilon = 1$, the tabular datasets use the same optimizer and learning rate as above, while the image datasets are trained using the Adam optimizer with a learning rate of 0.005. In this case, the gradient clipping bound is set to 5 for all datasets.
Across all experiments, we use a batch size of 256 and train for 20 epochs.

\paragraph{Models.}
\label{app:models_setup}
For the MNIST dataset, we employ three types of models for training: Logistic Regression (LR), Multilayer Perceptron (MLP), and Convolutional Neural Network (CNN).
For the tabular datasets, we use the LR model.
For the UTKFace dataset, we adopt the CNN model.

The LR model consists of a single fully connected layer that directly maps the input features to the output classes, without any hidden layers or activation functions.
The MLP model includes one hidden layer with 256 neurons and uses the tanh activation function. The input is first flattened and passed through the hidden layer, followed by an output layer.
The CNN model comprises two convolutional layers followed by two fully connected layers. The first convolutional layer uses 16 filters of size 5×5, followed by a 2×2 max pooling layer. The second convolutional layer has 32 filters of size 4×4, also followed by max pooling. After flattening the resulting feature maps, the output is passed through a fully connected layer with 32 neurons and then through the final classification layer. The tanh activation function is applied throughout the network.

\paragraph{Evaluation metrics.}
\label{app:metrics}
We use the PA-ALOOA method to obtain the results of GPR and GPRP metrics. 
We set $2R=400$ and $n=m$, which we believe is a reasonable configuration, as analyzed in Sec.~\ref{sec:3}. We use accuracy to measure model prediction performance.
Specifically, the accuracy of the training algorithms is computed by splitting the datasets into 80\% for training and 20\% for testing. 

\paragraph{Experimental Testbed.}
All our experiments are conducted on a cluster equipped with 10 NVIDIA A100 GPUs, 128 CPU cores, and 540 GB of RAM.
One privacy auditing procedure with $2R=400$ took approximately 5 hours to complete on the MNIST dataset using a CNN model under DP-SGD-S.

\subsection{Supplementary Experimental Results}
\label{app:add_results}

\subsubsection{Comparison between GPR and GRC}
\label{app:4.1}

\begin{figure}[H]
    \centering
    \begin{subfigure}[b]{0.32\textwidth} 
        \centering
        \includegraphics[width=\linewidth]{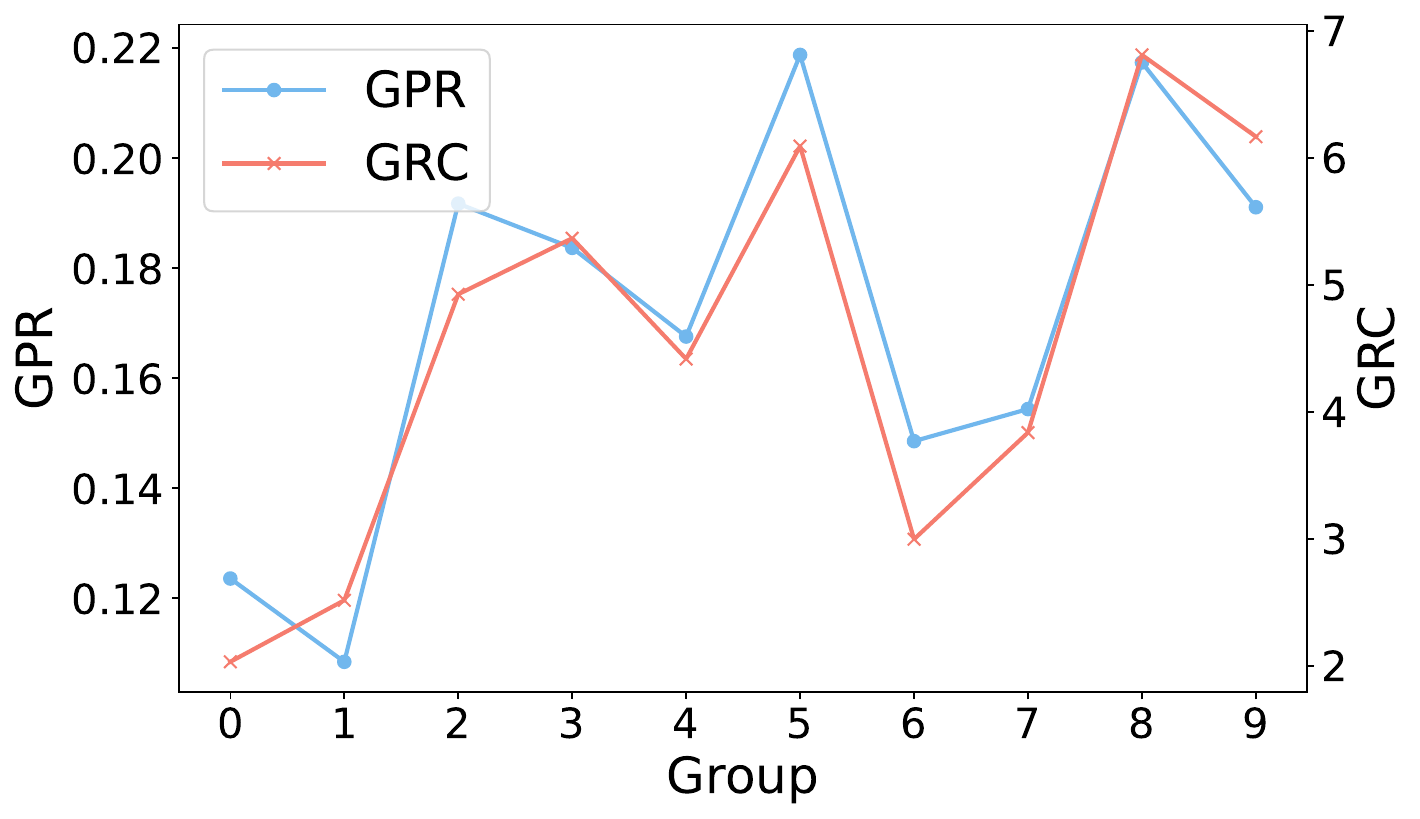} 
        \caption{LR.}
        \label{app:LR_grc}
    \end{subfigure}%
    \hfill
    \begin{subfigure}[b]{0.32\textwidth} 
        \centering
        \includegraphics[width=\linewidth]{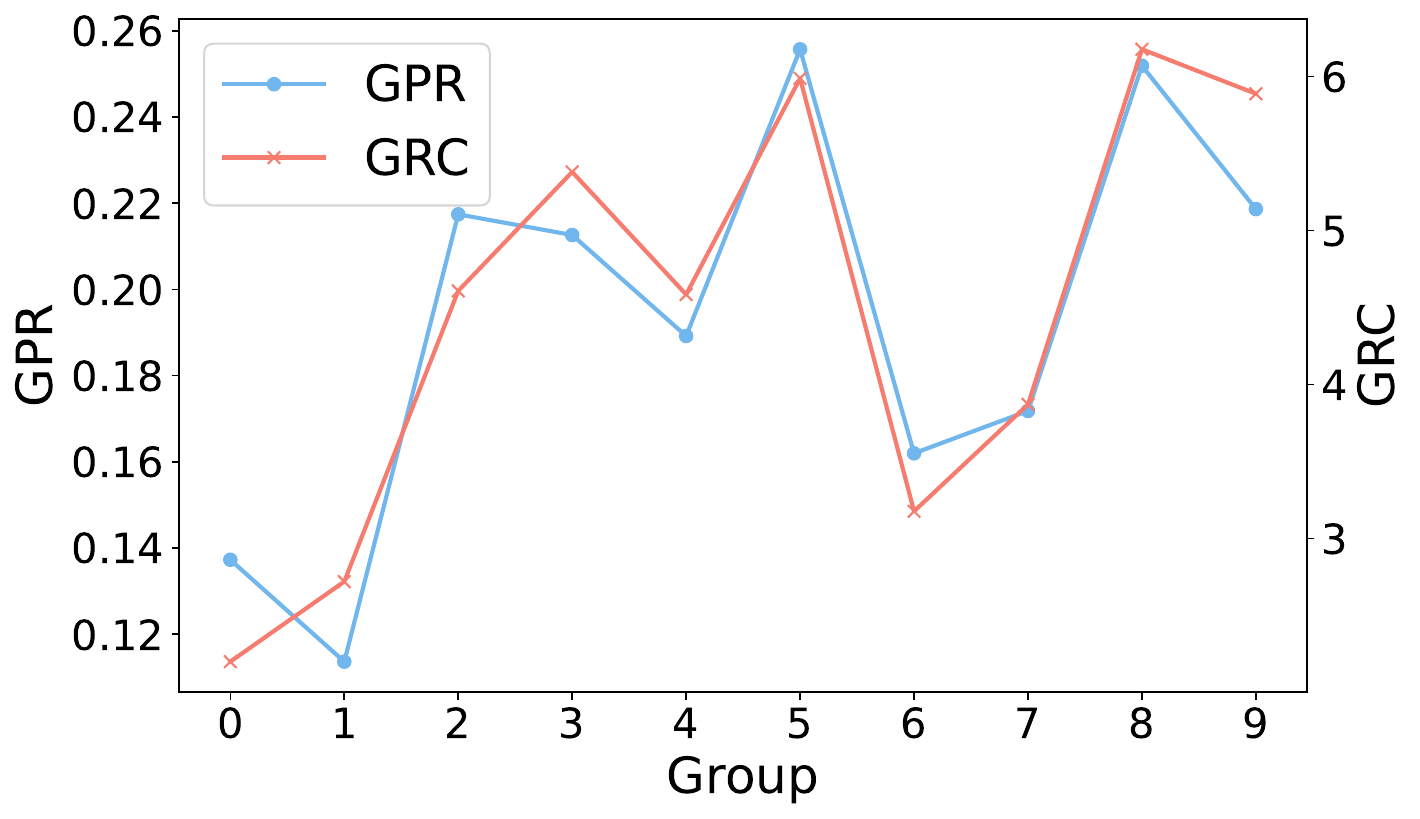} 
        \caption{MLP.}
        \label{app:MLP_grc}
    \end{subfigure}%
    \hfill
    \begin{subfigure}[b]{0.32\textwidth} 
        \centering
        \includegraphics[width=\linewidth]{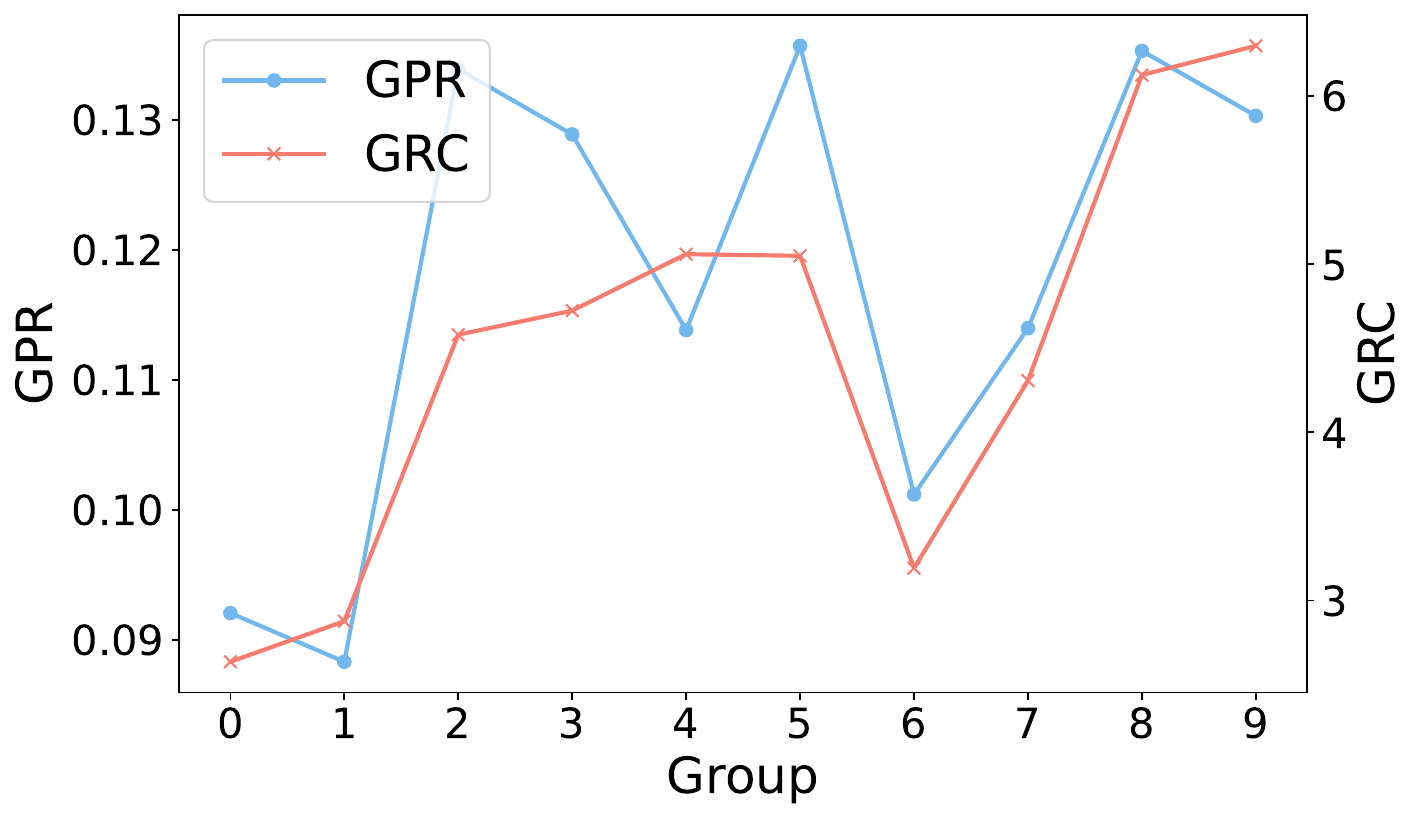} 
        \caption{CNN.}
        \label{app:CNN_grc}
    \end{subfigure}
    \caption{The values of GPR and GRC across three models—LR, MLP, and CNN—trained on the MNIST dataset using SGD training algorithm. Each subfigure shows the GPR and GRC for different groups, with the left y-axis indicating GPR, the right y-axis indicating GRC, and the x-axis representing group. All values are reported at $2R = 400$.}
    \label{app:adv_and_grc}
\end{figure}

As shown in Fig.~\ref{app:adv_and_grc}, there is a significant correlation between GRC and GPR across different models. Specifically, groups contributing more during training exhibit higher privacy leakage risks, with this phenomenon being more pronounced in simpler model architectures.

\subsubsection{Comparison between three training algorithms}
\label{app:5.2}

The complete results are presented in Tabs.~\ref{app_tab:metric_comparison_eps10}, \ref{app_tab:metric_comparison_eps1}, and \ref{tab:MNIST}. These tables provide a comprehensive comparison of algorithm performance across multiple datasets and privacy guarantees. As shown, the observed trends and conclusions are consistent with those discussed in the main text. 
However, there is one notable exception. Specifically, on the UTKFace dataset with $\epsilon=1$, our algorithm DP-SGD-S does not successfully reduce the disparity in group privacy risks. Despite this isolated case, DP-SGD-S consistently improves privacy fairness across all other experimental settings. These results underscore the robustness of DP-SGD-S in promoting group-level privacy fairness under most conditions.

Additionally, the Tab.~\ref{tab:MNIST} clearly shows that simpler model architectures, such as LR and MLP, exhibit relatively large $\Delta$ values. In contrast, more complex architecture CNN leads to a significant reduction in $\Delta$. This suggests that, when deploying models on public platforms, adopting more complex architectures may help reduce the disparity of privacy protection across groups and should be considered as a practical design choice.

\begin{table}[H]
    \centering
    \setlength{\tabcolsep}{0.9mm} 
    \caption{The performance of different algorithms across six datasets under $\epsilon=10$.}
    \label{app_tab:metric_comparison_eps10}
    \begin{tabular}{llcccccc}
        \toprule
        \textbf{Metric} & \textbf{Method} & \textbf{Adult} & \textbf{Bank} & \textbf{Credit} & \textbf{Law} & \textbf{UTKFace} \\
        \midrule
        \multirow{3}{*}{Accuracy~$(\uparrow)$} 
          & SGD & 85.00 $\pm$ 0.07 & 89.93 $\pm$ 0.06
          & 81.80 $\pm$ 0.04 & 89.75 $\pm$ 0.12 & 85.93 $\pm$ 3.79\\
          & DP-SGD & 84.92 $\pm$ 0.04 & 89.94 $\pm$ 0.04
          & 81.83 $\pm$ 0.07 & 89.74 $\pm$ 0.08 & 86.84 $\pm$ 0.48 \\
          & DP-SGD-S & 84.86 $\pm$ 0.11 & 89.91 $\pm$ 0.09
          & 81.83 $\pm$ 0.16 & 89.56 $\pm$ 0.18 & 84.67 $\pm$ 0.21\\
        \midrule
        \multirow{3}{*}{GPRP $\Delta~(\downarrow)$} 
          & SGD & 0.42 $\pm$ 0.04 & 0.26 $\pm$ 0.05
          & 0.20 $\pm$ 0.07 & 0.90 $\pm$ 0.16 & 1.75 $\pm$ 0.11 \\
          & DP-SGD & 0.27 $\pm$ 0.04 & 0.19 $\pm$ 0.03
          & 0.14 $\pm$ 0.05 & 0.59 $\pm$ 0.06 & 1.19 $\pm$ 0.07\\
          & DP-SGD-S & 0.17 $\pm$ 0.02 & 0.13 $\pm$ 0.03
          & 0.08 $\pm$ 0.04 & 0.41 $\pm$ 0.07 & 0.59 $\pm$ 0.07\\
        \bottomrule
    \end{tabular}
\end{table}

\begin{table}[H]
    \centering
    \setlength{\tabcolsep}{0.9mm} 
    \caption{The performance of different algorithms across six datasets under $\epsilon=1$.}
    \label{app_tab:metric_comparison_eps1}
    \begin{tabular}{llccccc}
        \toprule
        \textbf{Metric} & \textbf{Method} & \textbf{Adult} & \textbf{Bank} & \textbf{Credit} & \textbf{Law} & \textbf{UTKFace} \\
        \midrule
        \multirow{2}{*}{Accuracy~$(\uparrow)$} 
          & DP-SGD & 84.84 $\pm$ 0.07 & 89.97 $\pm$ 0.04
          & 81.73 $\pm$ 0.09 & 89.60 $\pm$ 0.15 & 81.60 $\pm$ 0.13 \\
          & DP-SGD-S & 84.64 $\pm$ 0.18 & 89.81 $\pm$ 0.20
          & 81.62 $\pm$ 0.17 & 89.59 $\pm$ 0.12 & 81.43 $\pm$ 0.91\\
        \midrule
        \multirow{2}{*}{GPRP $\Delta~(\downarrow)$} 
          & DP-SGD & 0.20 $\pm$ 0.03 & 0.14 $\pm$ 0.03
          & 0.08 $\pm$ 0.02 & 0.39 $\pm$ 0.04 & 0.24 $\pm$ 0.02\\
          & DP-SGD-S & 0.11 $\pm$ 0.02 & 0.09 $\pm$ 0.03
          & 0.03 $\pm$ 0.02 & 0.21 $\pm$ 0.06 & 0.27 $\pm$ 0.10\\
        \bottomrule
    \end{tabular}
\end{table}

\begin{table}[H]
    \centering
    \setlength{\tabcolsep}{1.2mm} 
    \caption{The performance of three algorithms across various model types under different theoretical privacy budgets, trained on MNIST dataset. The results include model prediction performance, GPRP value, and empirical privacy budget estimates.}
    \label{tab:MNIST}
    \begin{tabular}{cclcc}
        \toprule
        \textbf{Model} &${\epsilon}$ & \textbf{Method} &\textbf{Accuracy} $(\uparrow)$ & GPRP $\Delta (\downarrow)$ \\
        \midrule
        \multirow{7}{*}{LR}
        & / &SGD & 90.25 $\pm$ 0.23 & 11.13 $\pm$ 0.15\\
        \cmidrule{2-5}
        &\multirow{2}{*}{10}
        & DP-SGD & 89.07 $\pm$ 0.30 & 6.16 $\pm$ 0.19\\
        & &DP-SGD-S & 87.37 $\pm$ 0.41 & 4.87 $\pm$ 0.12\\
        \cmidrule{2-5}
        &\multirow{2}{*}{1}
        & DP-SGD & 85.42 $\pm$ 0.37 & 2.30 $\pm$ 0.16 \\
        & &DP-SGD-S & 84.09 $\pm$ 0.44 & 2.12 $\pm$ 0.14 \\
        \midrule
        \multirow{7}{*}{MLP}
        & / &SGD & 92.86 $\pm$ 0.28 & 14.17 $\pm$ 0.09 \\
        \cmidrule{2-5}
        &\multirow{2}{*}{10}
        & DP-SGD & 90.31 $\pm$ 0.41 & 7.54 $\pm$ 0.12 \\
        & &DP-SGD-S & 87.56 $\pm$ 0.41 & 5.89 $\pm$ 0.06 \\
        \cmidrule{2-5}
        &\multirow{2}{*}{1}
        & DP-SGD & 84.94 $\pm$ 0.94 & 2.73 $\pm$ 0.13 \\
        & &DP-SGD-S & 83.95 $\pm$ 0.80 & 2.30 $\pm$ 0.12\\
        \midrule
        \multirow{7}{*}{CNN}
        & / &SGD & 95.89 $\pm$ 0.29 & 4.92 $\pm$ 0.18\\
        \cmidrule{2-5}
        &\multirow{2}{*}{10}
        & DP-SGD & 94.46 $\pm$ 0.13 & 3.54 $\pm$ 0.13\\
        & &DP-SGD-S & 92.63 $\pm$ 0.58 & 2.91 $\pm$ 0.14 \\
        \cmidrule{2-5}
        &\multirow{2}{*}{1}
        & DP-SGD & 89.06 $\pm$ 0.72 & 1.83 $\pm$ 0.09 \\
        & &DP-SGD-S & 88.58 $\pm$ 0.44 & 1.55 $\pm$ 0.14  \\
        \bottomrule
    \end{tabular}
\end{table}

\subsubsection{The results of each group privacy risk}
\label{app:5.3}
\begin{figure}[H]
    \centering
        \begin{subfigure}[b]{0.3\textwidth}
        \centering
        \includegraphics[width=\linewidth]{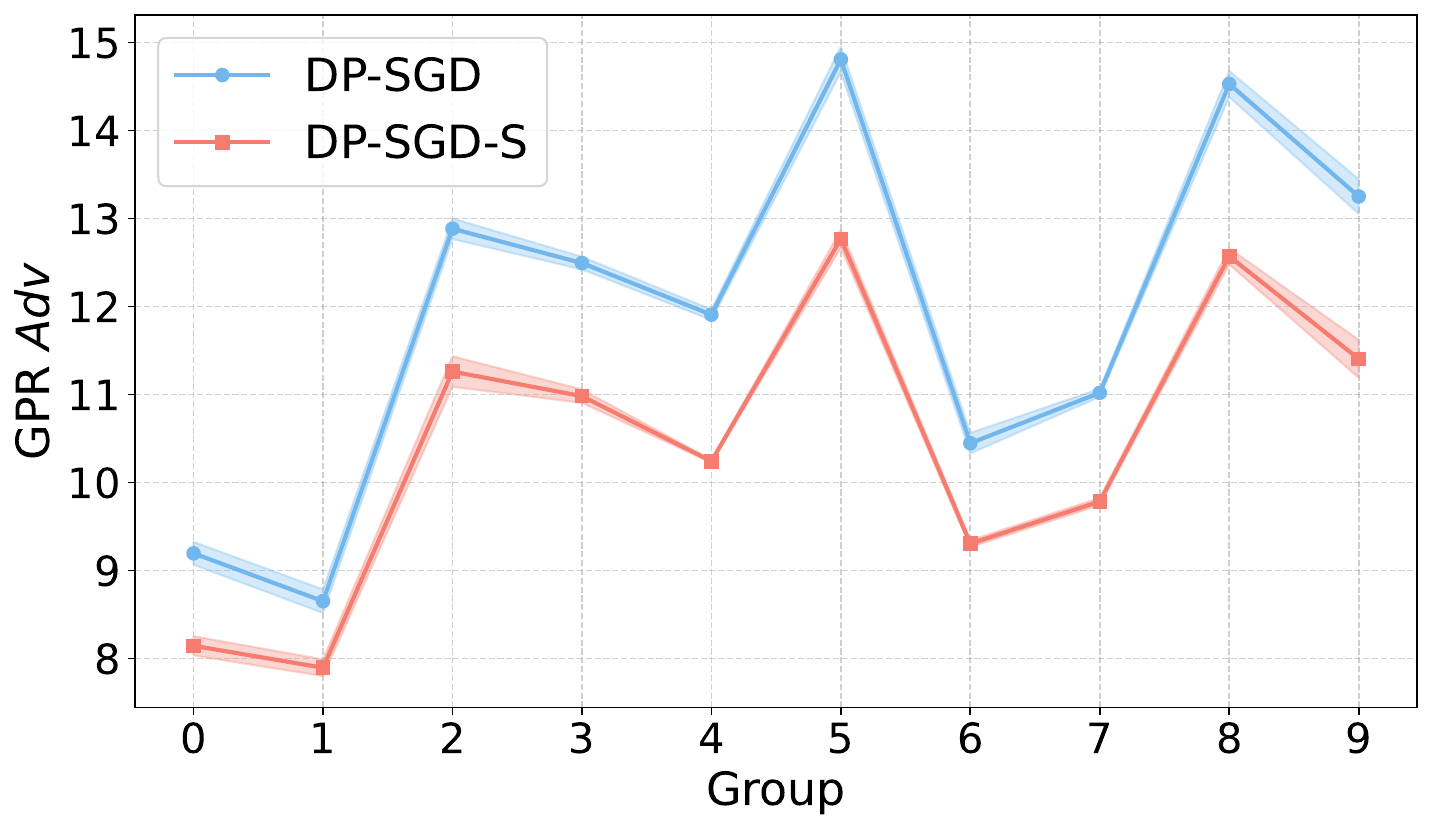} 
        \caption{LR with $\epsilon=10$.}
        \label{app5.3:mnist_subfig1}
    \end{subfigure}%
    \hfill
    \begin{subfigure}[b]{0.3\textwidth}
        \centering
        \includegraphics[width=\linewidth]{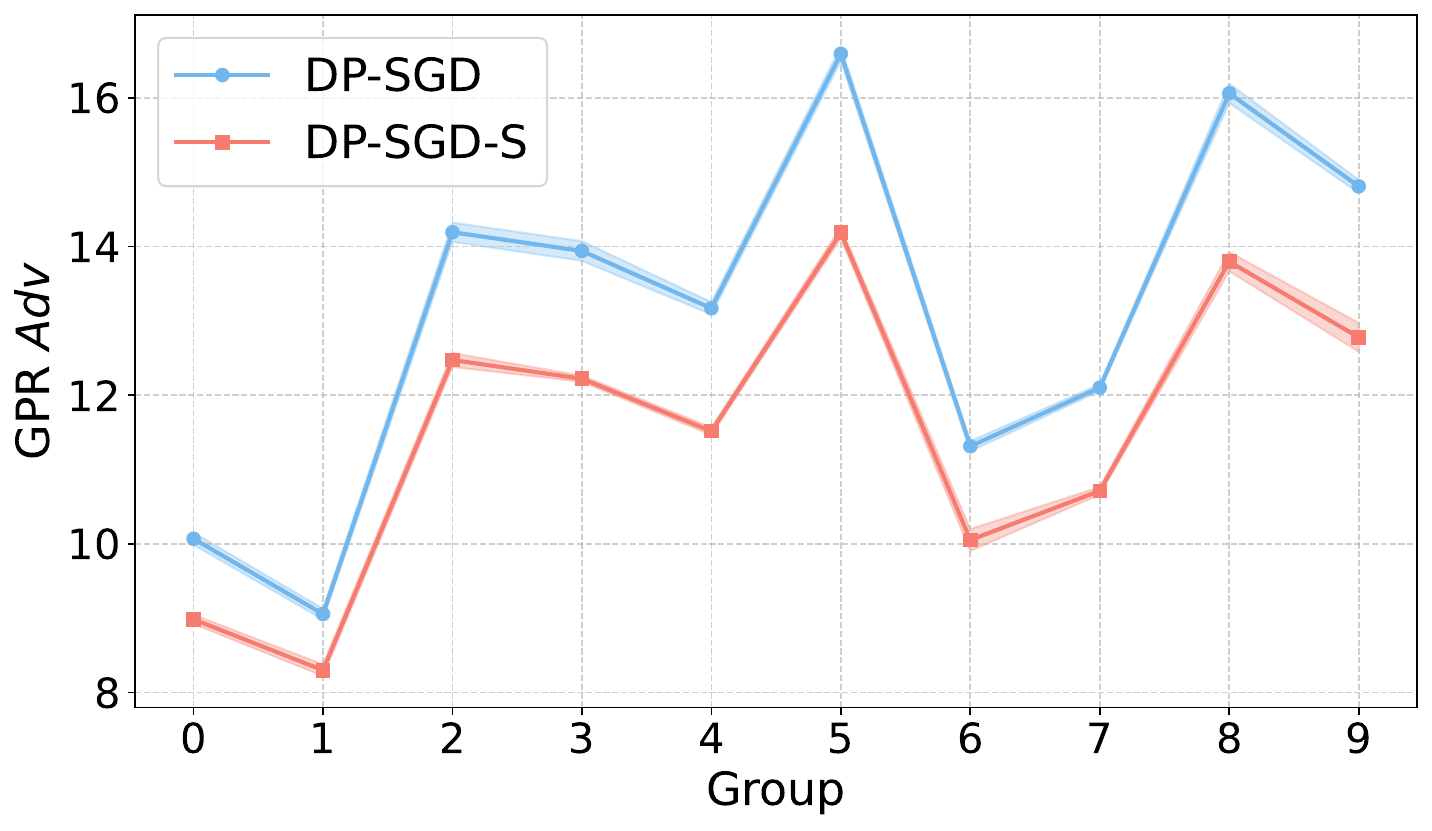} 
        \caption{MLP with $\epsilon=10$.}
        \label{app5.3:mnist_subfig2}
    \end{subfigure}%
    \hfill
    \begin{subfigure}[b]{0.3\textwidth}
        \centering
        \includegraphics[width=\linewidth]{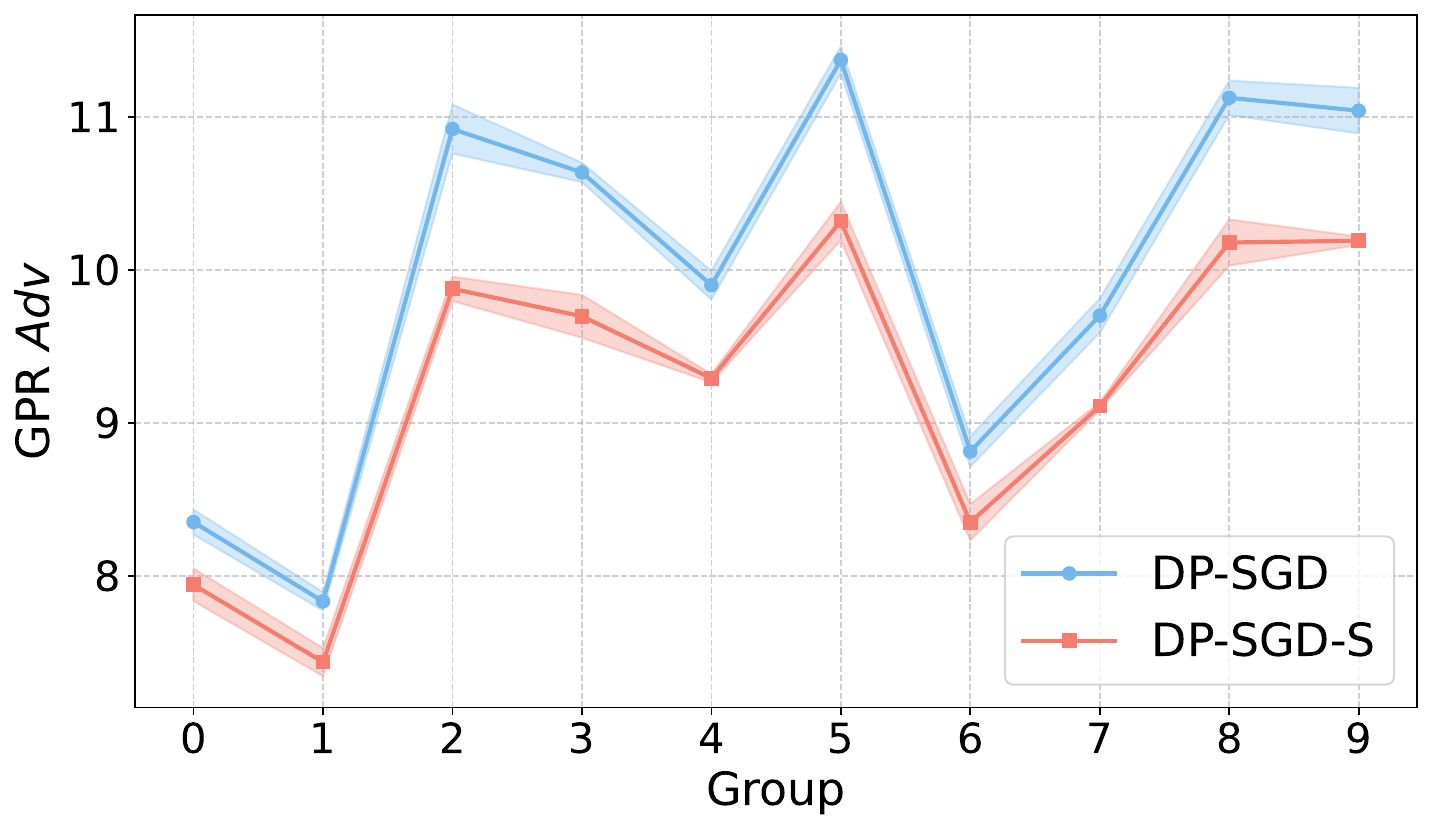} 
        \caption{CNN with $\epsilon=10$.}
        \label{app5.3:mnist_subfig3}
    \end{subfigure}
    \centering
    \begin{subfigure}[b]{0.3\textwidth}
        \centering
        \includegraphics[width=\linewidth]{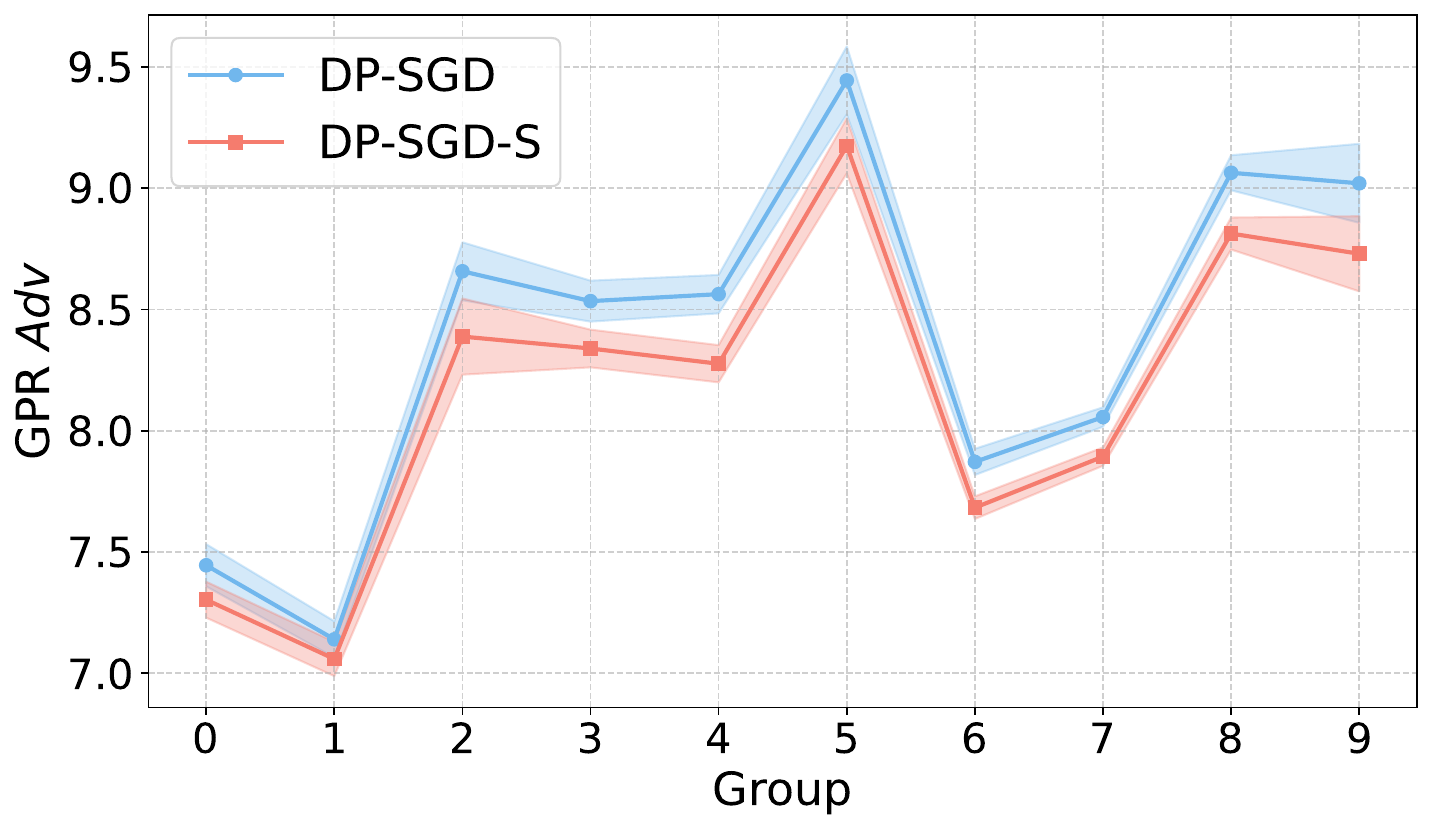} 
        \caption{LR with $\epsilon=1$.}
        \label{app5.3:mnist_subfig4}
    \end{subfigure}%
    \hfill
    \begin{subfigure}[b]{0.3\textwidth}
        \centering
        \includegraphics[width=\linewidth]{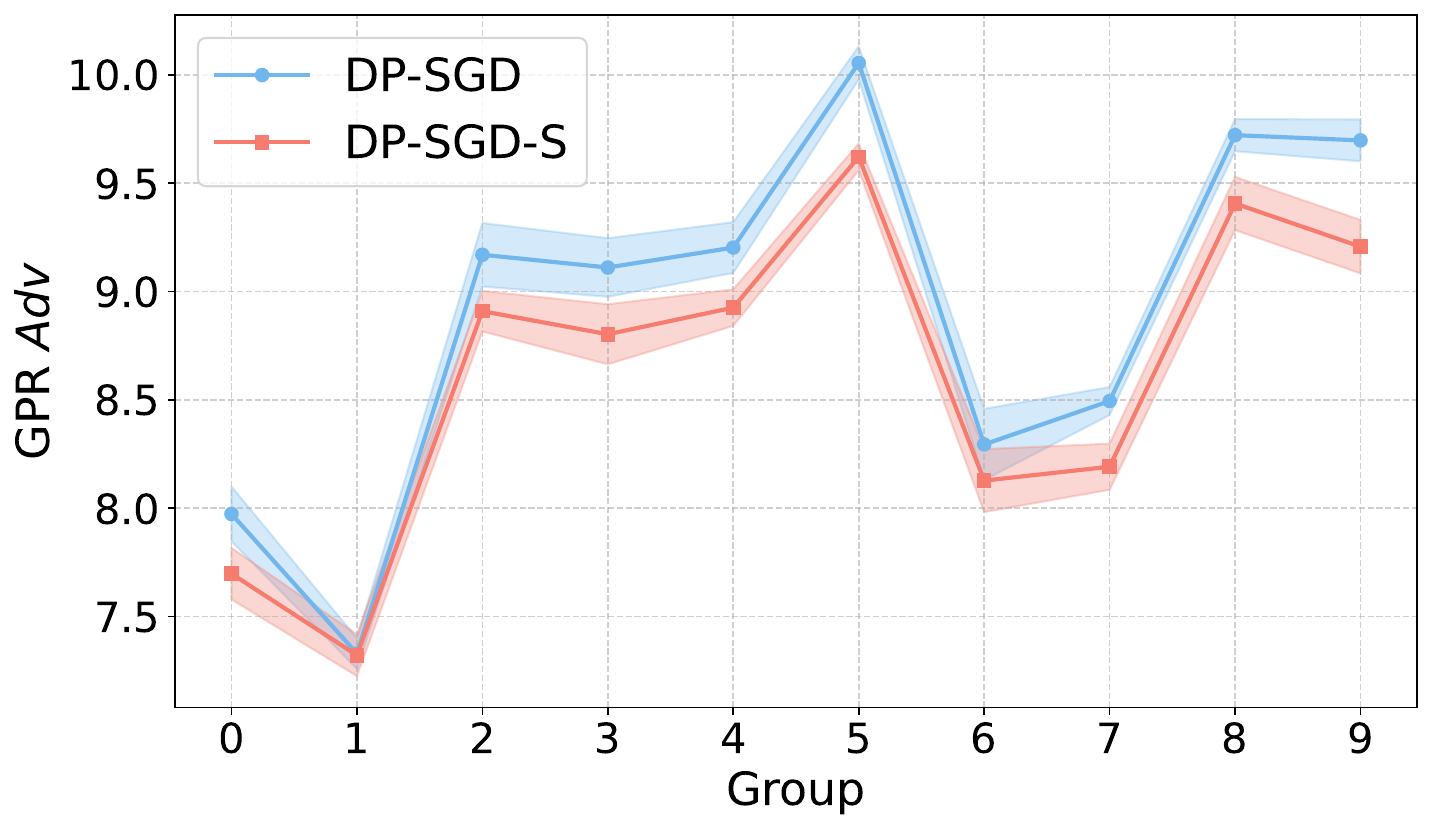} 
        \caption{MLP with $\epsilon=1$.}
        \label{app5.3:mnist_subfig5}
    \end{subfigure}%
    \hfill
    \begin{subfigure}[b]{0.3\textwidth}
        \centering
        \includegraphics[width=\linewidth]{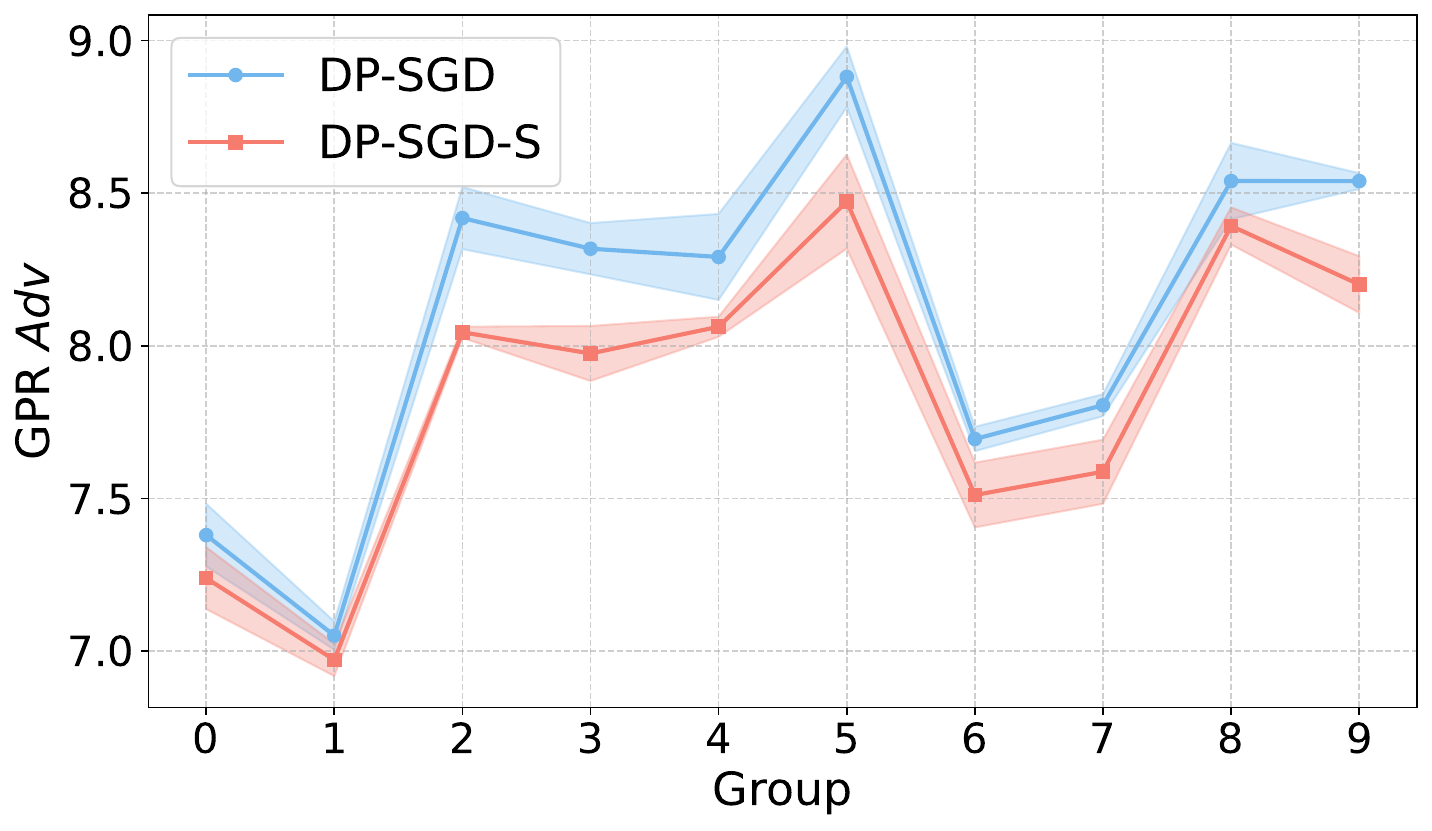} 
        \caption{CNN with $\epsilon=1$.}
        \label{app5.3:mnist_subfig6}
    \end{subfigure}
    \caption{The GPR $Adv$ of each group across three models and different privacy budgets, trained on MNIST dataset. In each subfigure, the vertical axis represents GPR value at $2R=400$.}
    \label{app5.3:mnist}
\end{figure}

\begin{figure}[H]
    \centering
        \begin{subfigure}[b]{0.3\textwidth}
        \centering
        \includegraphics[width=\linewidth]{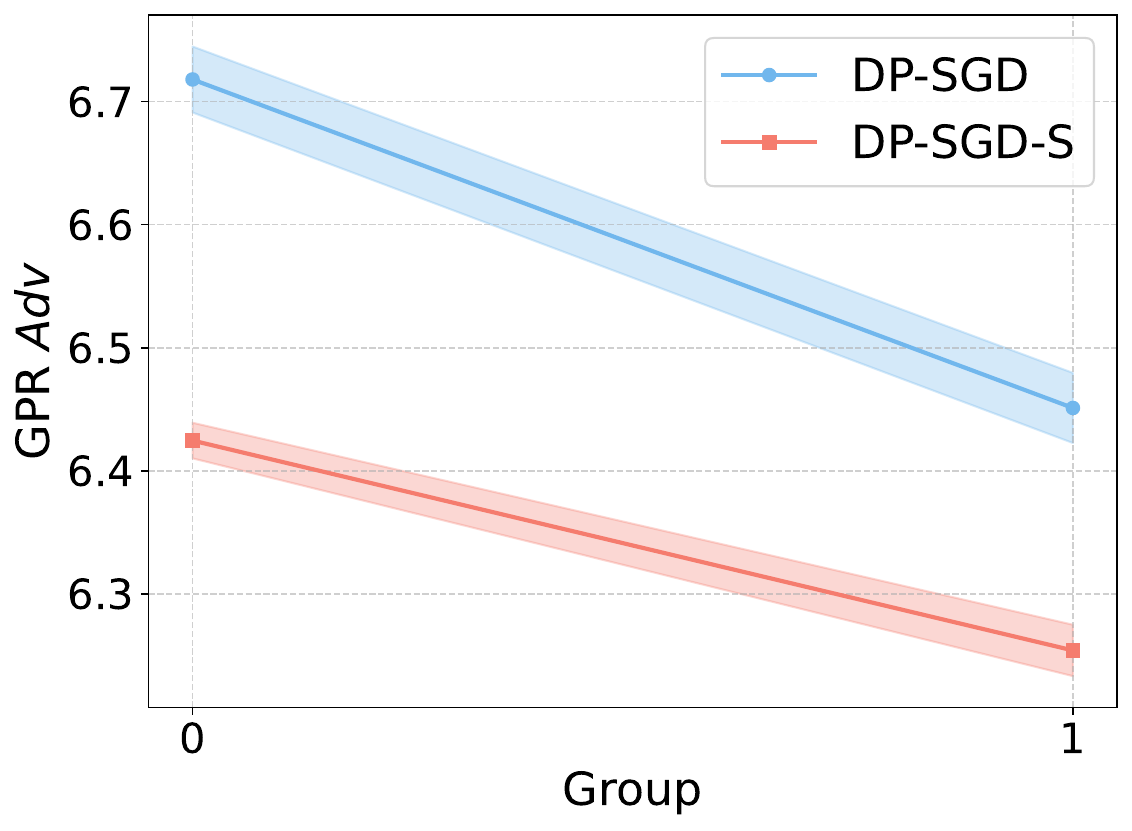} 
        \caption{Adult, $\epsilon=10$.}
        \label{app5.3:adult_eps10}
    \end{subfigure}%
    \hfill
    \begin{subfigure}[b]{0.3\textwidth}
        \centering
        \includegraphics[width=\linewidth]{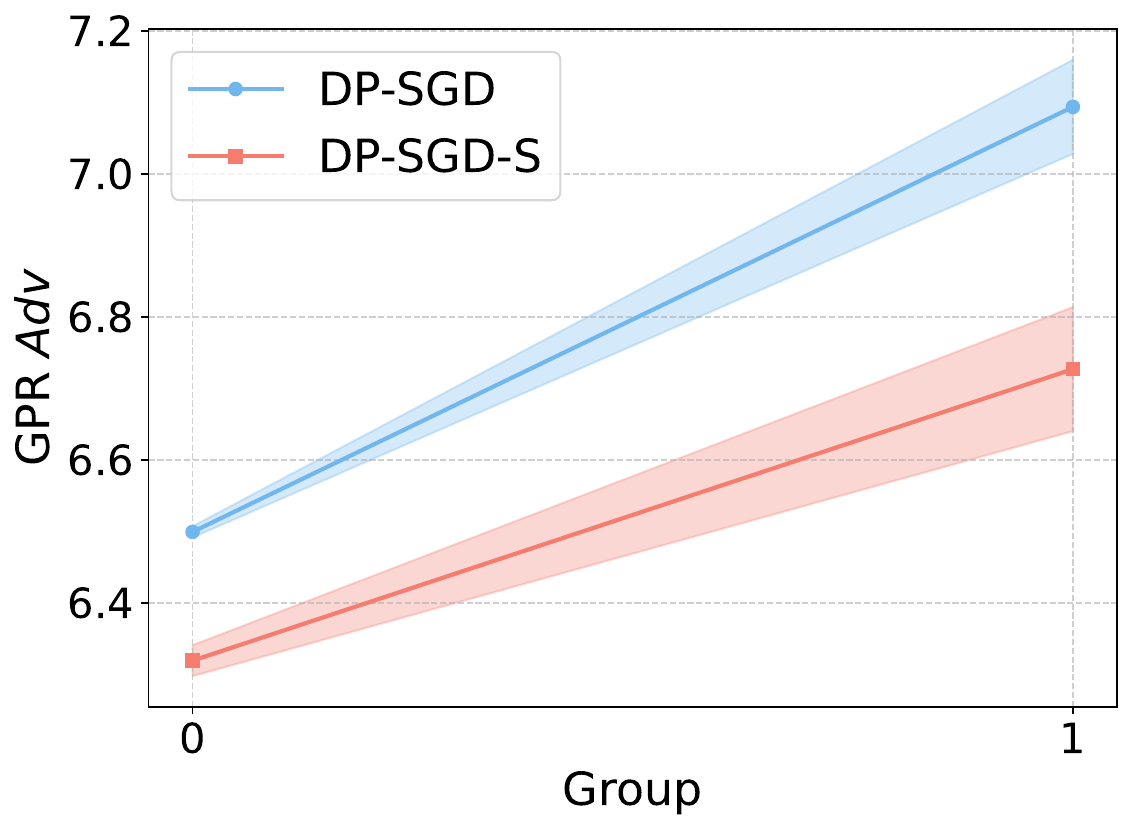} 
        \caption{Law, $\epsilon=10$.}
        \label{app5.3:law_eps10}
    \end{subfigure}%
    \hfill
    \begin{subfigure}[b]{0.3\textwidth}
        \centering
        \includegraphics[width=\linewidth]{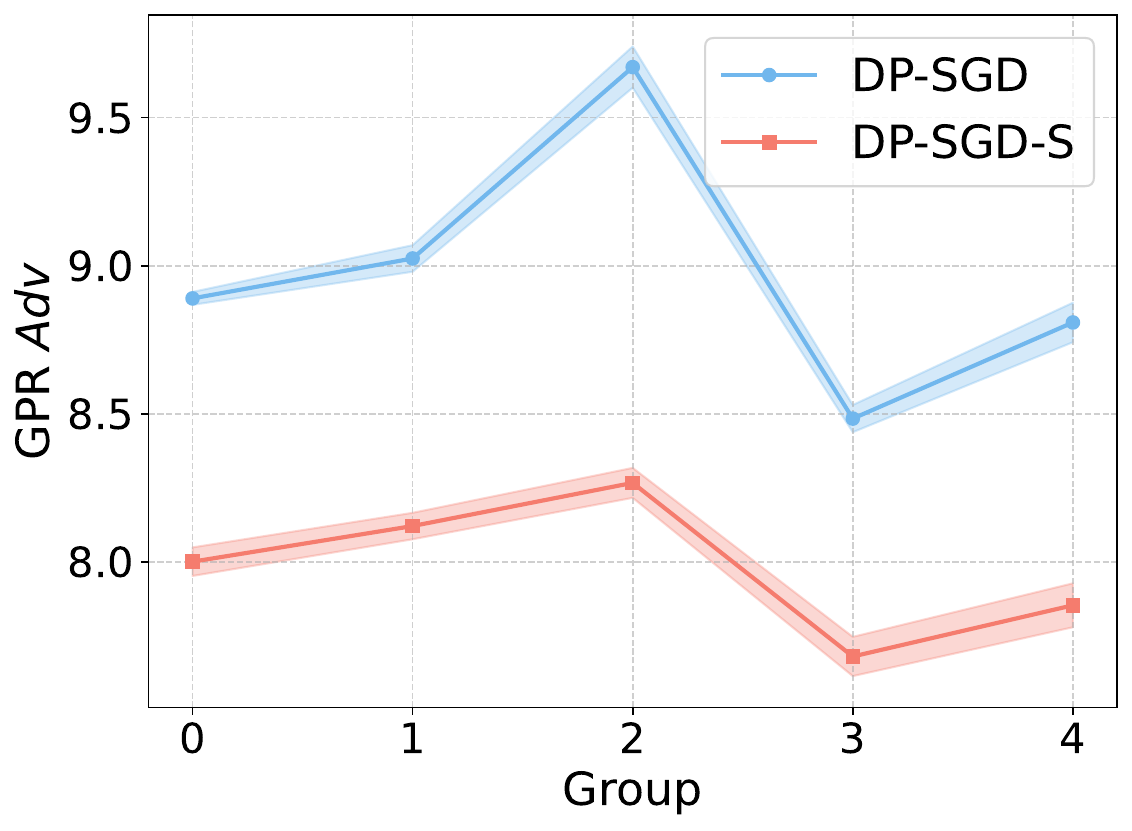} 
        \caption{UTKFace, $\epsilon=10$.}
        \label{app5.3:raceface_eps10}
    \end{subfigure}
    \centering
        \begin{subfigure}[b]{0.3\textwidth}
        \centering
        \includegraphics[width=\linewidth]{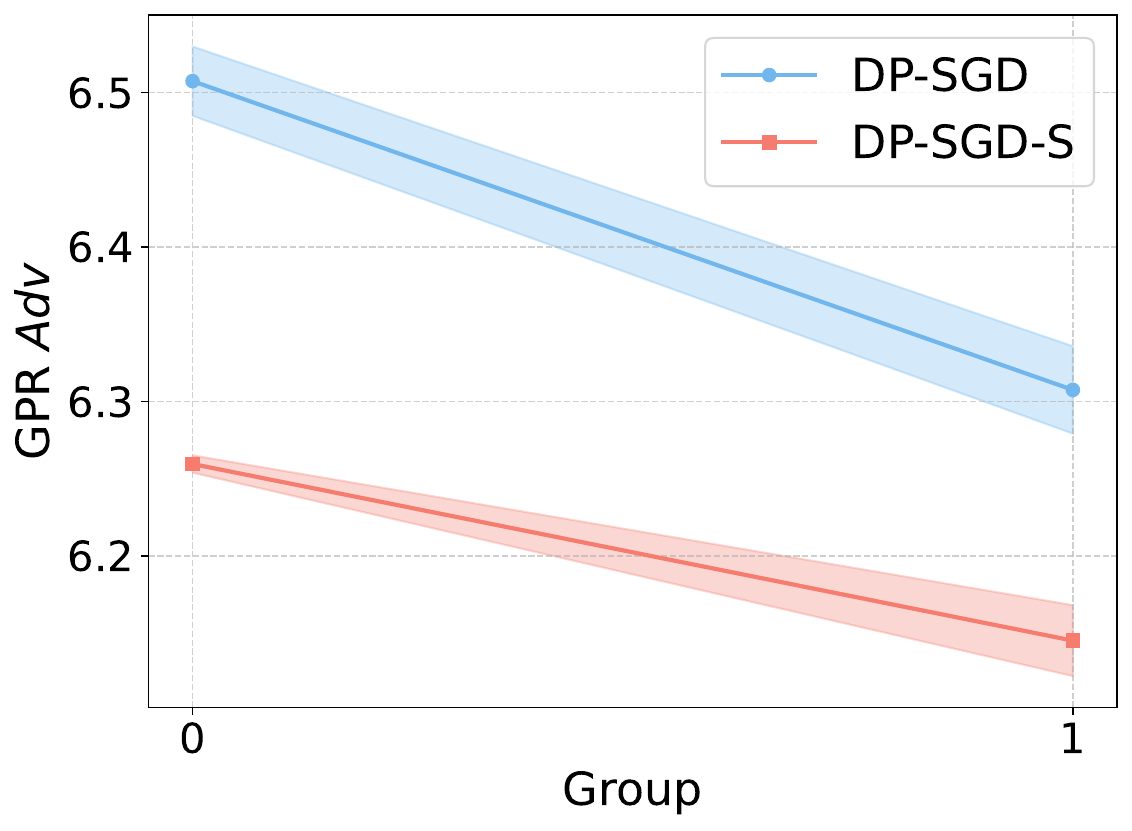} 
        \caption{Adult, $\epsilon=1$.}
        \label{app5.3:adult_eps1}
    \end{subfigure}%
    \hfill
    \begin{subfigure}[b]{0.3\textwidth}
        \centering
        \includegraphics[width=\linewidth]{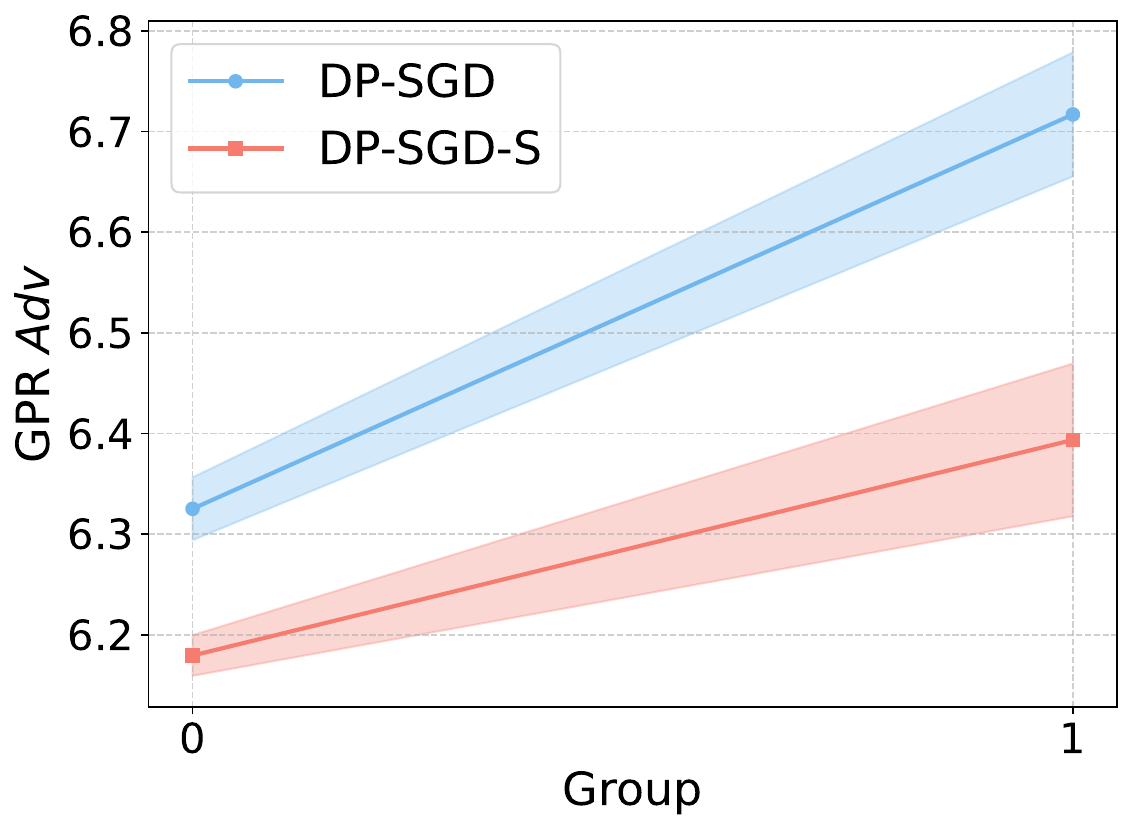} 
        \caption{Law, $\epsilon=1$.}
        \label{app5.3:law_eps1}
    \end{subfigure}%
    \hfill
    \begin{subfigure}[b]{0.3\textwidth}
        \centering
        \includegraphics[width=\linewidth]{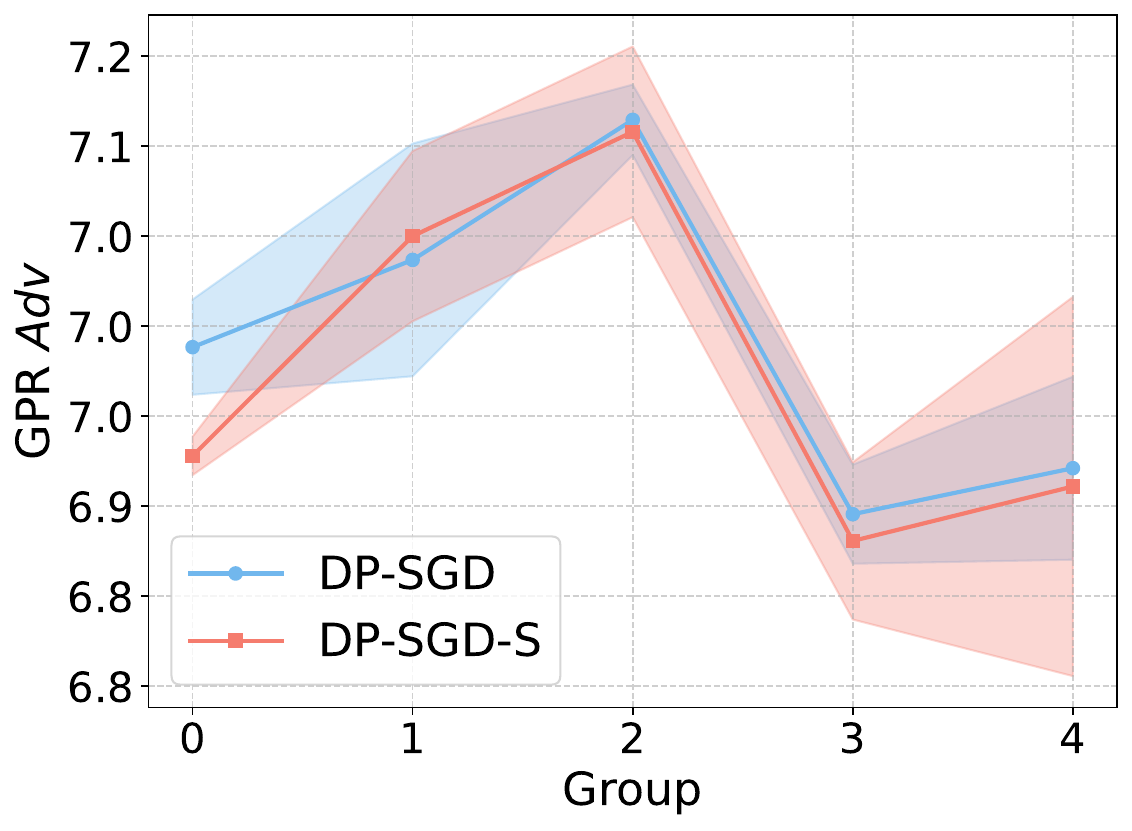} 
        \caption{UTKFace, $\epsilon=1$.}
        \label{app5.3:raceface_eps1}
    \end{subfigure}
    \caption{The GPR $Adv$ of each group across different datasets and privacy budgets. In each subfigure, the vertical axis represents GPR value at $2R=400$.}
    \label{app5.3:fair_dataset1}
\end{figure}

\begin{figure}[H]
    \centering
    \begin{subfigure}[b]{0.5\textwidth}
        \centering
        \includegraphics[width=0.8\linewidth]{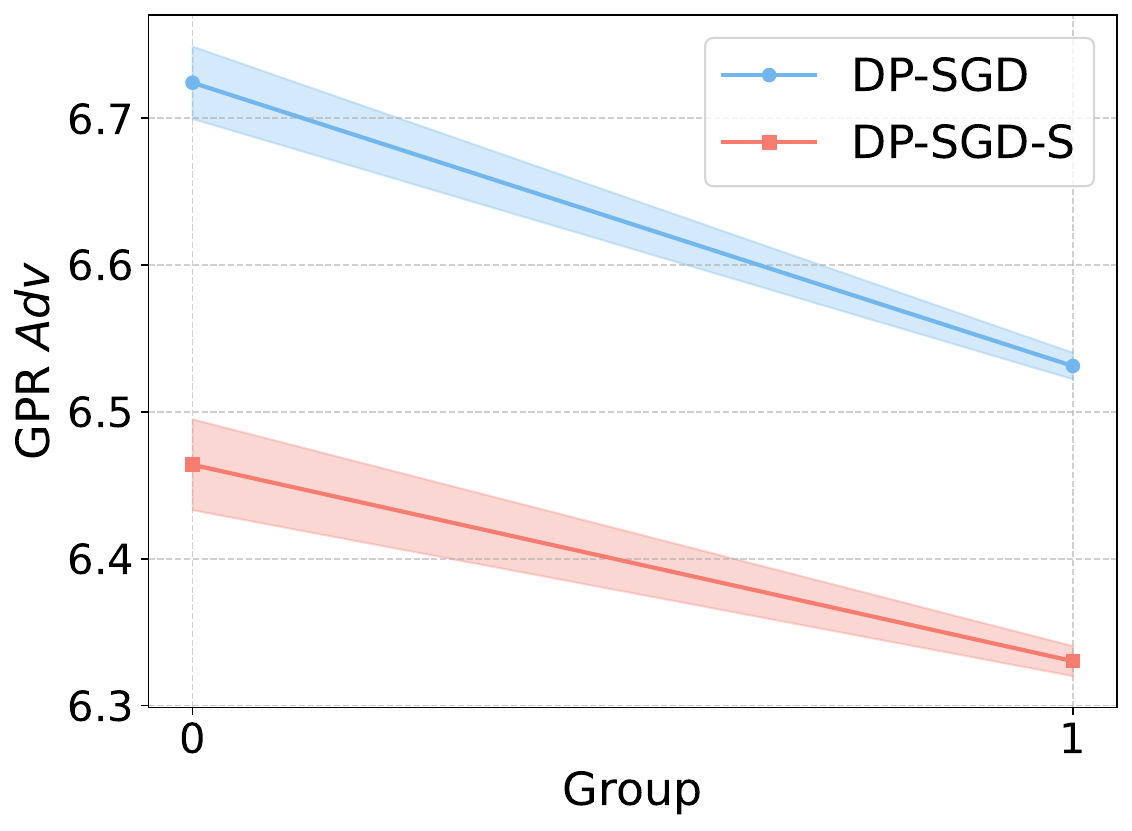} 
        \caption{Bank, $\epsilon=10$.}
        \label{app5.3:bank_eps10}
    \end{subfigure}%
    \hfill
    \begin{subfigure}[b]{0.5\textwidth}
        \centering
        \includegraphics[width=0.8\linewidth]{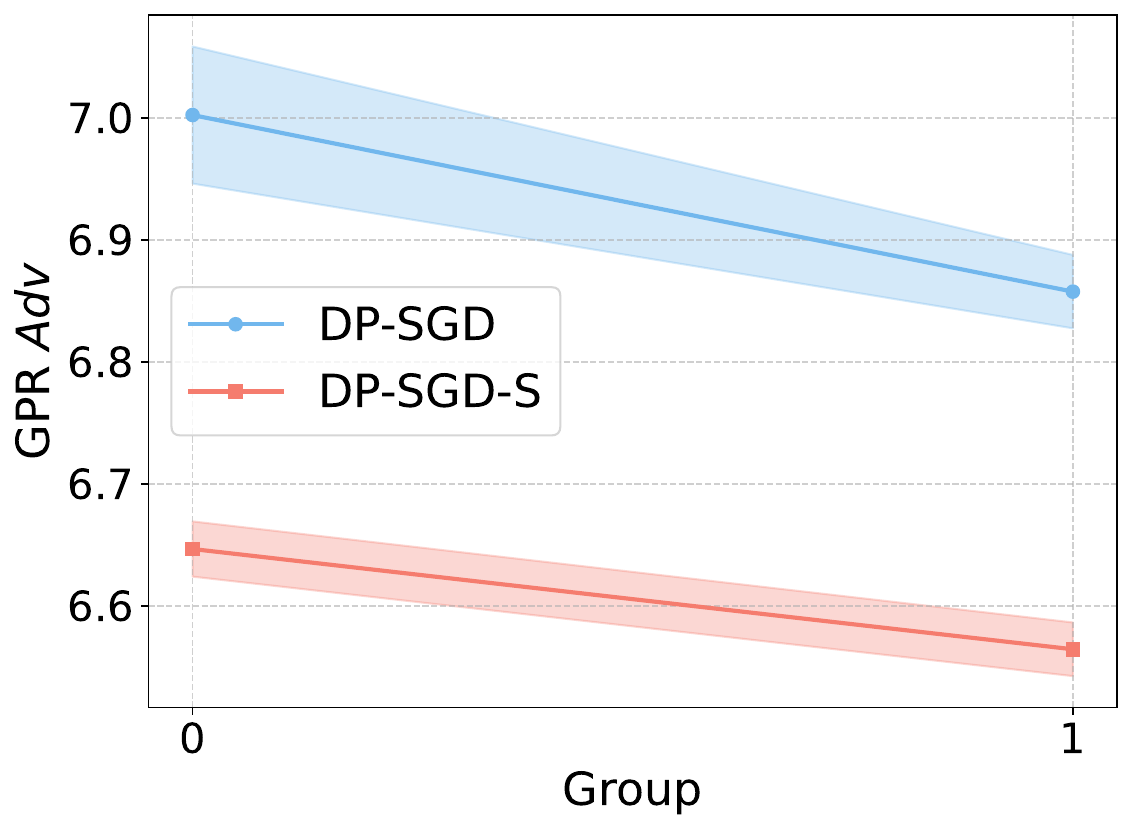} 
        \caption{Credit, $\epsilon=10$.}
        \label{app5.3:credit_eps10}
    \end{subfigure}
    \begin{subfigure}[b]{0.5\textwidth}
        \centering
        \includegraphics[width=0.8\linewidth]{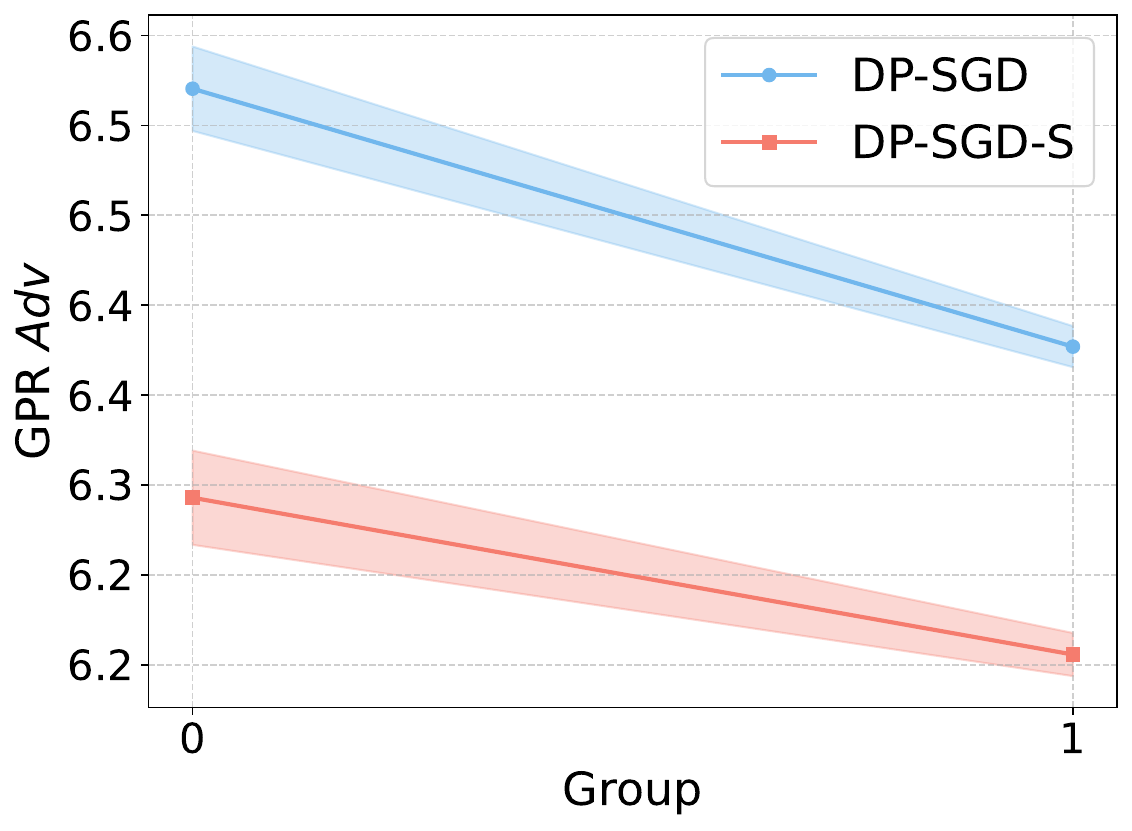} 
        \caption{Bank, $\epsilon=1$.}
        \label{app5.3:bank_eps1}
    \end{subfigure}%
    \hfill
    \begin{subfigure}[b]{0.5\textwidth}
        \centering
        \includegraphics[width=0.8\linewidth]{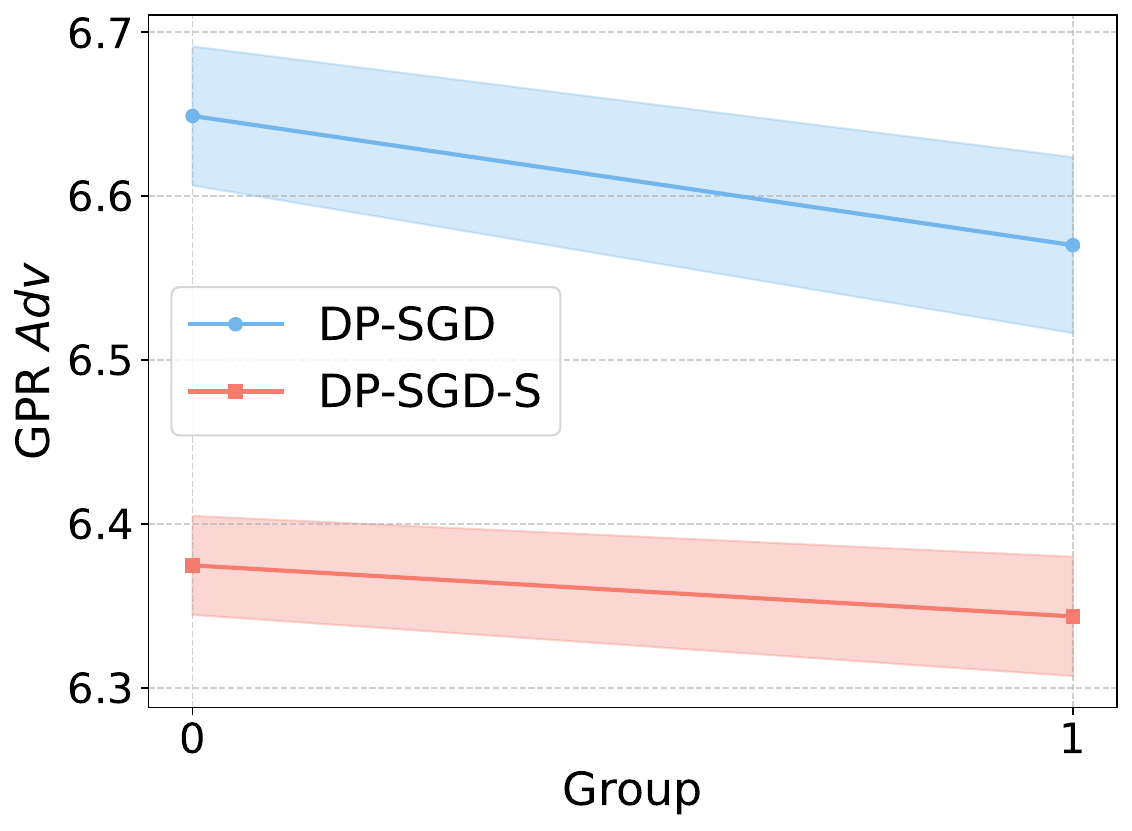} 
        \caption{Credit, $\epsilon=1$.}
        \label{app5.3:credit_eps1}
    \end{subfigure}

    \caption{The GPR $Adv$ of each group across different datasets and privacy budgets. In each subfigure, the vertical axis represents GPR value at $2R=400$.}
    \label{app5.3:fair_dataset2}
\end{figure}

We provide a comprehensive analysis of group-level privacy risks across all experimental settings, as illustrated in Figs.~\ref{app5.3:mnist}, \ref{app5.3:fair_dataset1}, and \ref{app5.3:fair_dataset2}. As observed from these figures, there are 16 subplots in total, each corresponding to a different experimental configuration. Among them, only one case shows that DP-SGD-S increases the privacy leakage risk for a specific group, which occurs on the UTKFace dataset with $\epsilon = 1$. This result demonstrates that improving privacy fairness does not compromise the protection of already well-protected groups. As a result, our method avoids the undesirable “leveling down” effect and ensures more responsible and practical deployment in real-world scenarios.

\subsubsection{The results of outcome fairness}
\label{app:outcome_fairness}
Traditional definitions of outcome fairness are primarily designed for binary classification tasks where the sensitive attribute has only two groups (e.g., male vs. female)~\cite{pessach2022review}. Under this setting, fairness metrics are typically computed using the absolute difference between the values of the two groups. However, in the case of the UTKFace dataset, the sensitive attribute is multi-class, containing more than two categories.
To accommodate this, we adopt a natural extension of the conventional definitions: instead of calculating the absolute difference between two groups, we compute the range across all subgroups, i.e., the difference between the maximum and minimum values of the fairness metric across the group set. This allows us to quantify disparities in outcomes among multiple demographic groups.

The exact formulas used to compute outcome fairness measurements are as follows.  
Here, $s$ represents the sensitive attribute, and $K$ denotes the set of all subgroups:
\begin{itemize}
    \item {Accuracy Parity (AP)}: the maximum disparity in prediction accuracy among different groups:
    \begin{equation*}
        \text{AP} = \max_{k \in K} \left( P(\hat{y} = y \mid s = s_k) \right) 
        - \min_{k \in K} \left( P(\hat{y} = y \mid s = s_k) \right)
    \end{equation*}
    \item {Demographic Parity (DmP)}: the maximum disparity in the rate of positive predictions across groups:
    \begin{equation*}
        \text{DmP} = \max_{k \in K} \left( P(\hat{y} = 1 \mid s = s_k) \right)
        - \min_{k \in K} \left( P(\hat{y} = 1 \mid s = s_k) \right)
    \end{equation*}
    \item {Equal Opportunity (EOp)}: the maximum gap in true positive rates among groups:
    \begin{equation*}
        \text{EOp} = \max_{k \in K} \left( P(\hat{y} = 1 \mid s = s_k, y = 1) \right) 
        - \min_{k \in K} \left( P(\hat{y} = 1 \mid s = s_k, y = 1) \right)
    \end{equation*}
    \item {Equalized Odds (EOd)}: the maximum divergence in both true positive and false positive rates among groups:
    \begin{equation*}
    \begin{aligned}
        \text{EOd} &= \max_{k \in K} \left( 
        P(\hat{y} = 1 \mid s = s_k, y = 1) 
        + P(\hat{y} = 1 \mid s = s_k, y = 0) \right) \\
        &\quad - \min_{k \in K} \left( 
        P(\hat{y} = 1 \mid s = s_k, y = 1) 
        + P(\hat{y} = 1 \mid s = s_k, y = 0) \right)
    \end{aligned}
    \end{equation*}
\end{itemize}

The experimental results for outcome fairness metrics of three algorithms, evaluated under varying theoretical privacy budgets on fairness-related datasets, are presented in Tab.~\ref{tab:fairness_metric}. From the table, we can observe that DP-SGD-S and DP-SGD perform similarly, with no clear superiority of one over the other across these metrics.

\begin{table}[H]
    \centering
    \setlength{\tabcolsep}{0.8mm} 
    \caption{The outcome fairness measurements of three algorithms under different theoretical privacy budgets, including AP, DmP, EOp, and EOd.}
    \label{tab:fairness_metric}
        \begin{tabular}{cclcccc}
        \toprule
        \textbf{Dataset} & $\epsilon$ & \textbf{Method} & \textbf{AP} & \textbf{DmP} & \textbf{EOp} & \textbf{EOd} \\
        \midrule
        \multirow{5}{*}{Adult}
        & / & SGD & 0.1148 $\pm$ 0.0012 & 0.1845 $\pm$ 0.0035 & 0.1103 $\pm$ 0.0058 & 0.1831 $\pm$ 0.0049 \\
        \cmidrule{2-7}
        & \multirow{2}{*}{10}
        & DP-SGD & 0.1175 $\pm$ 0.0015 & 0.1886 $\pm$ 0.0045 & 0.1078 $\pm$ 0.0043 & 0.1853 $\pm$ 0.0045 \\
        & & DP-SGD-S & 0.1160 $\pm$ 0.0022 & 0.1877 $\pm$ 0.0036 & 0.1045 $\pm$ 0.0138 & 0.1811 $\pm$ 0.0159 \\
        \cmidrule{2-7}
        & \multirow{2}{*}{1}
        & DP-SGD & 0.1171 $\pm$ 0.0013 & 0.1889 $\pm$ 0.0067 & 0.1053 $\pm$ 0.0259 & 0.1834 $\pm$ 0.0294 \\
        & & DP-SGD-S & 0.1193 $\pm$ 0.0021 & 0.1925 $\pm$ 0.0006 & 0.1294 $\pm$ 0.0097 & 0.2107 $\pm$ 0.0093 \\
        \midrule
        \multirow{5}{*}{Bank}
        & / & SGD & 0.0283 $\pm$ 0.0012 & 0.0401 $\pm$ 0.0044 & 0.0881 $\pm$ 0.0090 & 0.1034 $\pm$ 0.0112 \\
        \cmidrule{2-7}
        & \multirow{2}{*}{10}
        & DP-SGD & 0.0283 $\pm$ 0.0013 & 0.0406 $\pm$ 0.0040 & 0.0927 $\pm$ 0.0175 & 0.1083 $\pm$ 0.0194 \\
        & & DP-SGD-S & 0.0272 $\pm$ 0.0022 & 0.0435 $\pm$ 0.0065 & 0.1179 $\pm$ 0.0219 & 0.1344 $\pm$ 0.0253 \\
        \cmidrule{2-7}
        & \multirow{2}{*}{1}
        & DP-SGD & 0.0279 $\pm$ 0.0008 & 0.0402 $\pm$ 0.0024 &  0.0900 $\pm$ 0.0106 & 0.1052 $\pm$ 0.0119 \\
        & & DP-SGD-S & 0.0300 $\pm$ 0.0032 & 0.0397 $\pm$ 0.0044 & 0.0902 $\pm$ 0.0161 & 0.1062 $\pm$ 0.0169 \\
        \midrule
        \multirow{5}{*}{Credit}
        & / & SGD & 0.0181 $\pm$ 0.0009 & 0.0365 $\pm$ 0.0018 & 0.0576 $\pm$ 0.0050 & 0.0772 $\pm$ 0.0057 \\
        \cmidrule{2-7}
        & \multirow{2}{*}{10}
        & DP-SGD & 0.0185 $\pm$ 0.0011 & 0.0367 $\pm$ 0.0029 & 0.0569 $\pm$ 0.0048 & 0.0768 $\pm$ 0.0070 \\
        & & DP-SGD-S & 0.0214 $\pm$ 0.0018 & 0.0339 $\pm$ 0.0044 & 0.0454 $\pm$ 0.0096 & 0.0654 $\pm$ 0.0120 \\
        \cmidrule{2-7}
        & \multirow{2}{*}{1}
        & DP-SGD & 0.0213 $\pm$ 0.0012 & 0.0362 $\pm$ 0.0062 & 0.0502 $\pm$ 0.0134 & 0.0717 $\pm$ 0.0172 \\
        & & DP-SGD-S & 0.0202 $\pm$ 0.0030 & 0.0375 $\pm$ 0.0103 & 0.0570 $\pm$ 0.0226 & 0.0785 $\pm$ 0.0290 \\
        \midrule
        \multirow{5}{*}{Law}
        & / & SGD & 0.1499 $\pm$ 0.0053 & 0.2136 $\pm$ 0.0165 & 0.1181 $\pm$ 0.0151 & 0.5262 $\pm$ 0.0265 \\
        \cmidrule{2-7}
        & \multirow{2}{*}{10}
        & DP-SGD & 0.1537 $\pm$ 0.0043 & 0.2161 $\pm$ 0.0145 & 0.1223 $\pm$ 0.0117 & 0.5306 $\pm$ 0.0329 \\
        & & DP-SGD-S & 0.1564 $\pm$ 0.0043 & 0.2248 $\pm$ 0.0153 & 0.1308 $\pm$ 0.0128 & 0.5407 $\pm$ 0.0333 \\
        \cmidrule{2-7}
        & \multirow{2}{*}{1}
        & DP-SGD & 0.1554 $\pm$ 0.0054 & 0.2140 $\pm$ 0.0177 & 0.1223 $\pm$ 0.0117 & 0.5231 $\pm$ 0.0413 \\
        & & DP-SGD-S & 0.1582 $\pm$ 0.0049 & 0.2252 $\pm$ 0.0202 & 0.1315 $\pm$ 0.0160 & 0.5581 $\pm$ 0.0469 \\
        \midrule
        \multirow{5}{*}{UTKFace}
        & / & SGD & 0.0773 $\pm$ 0.0140 & 0.1548 $\pm$ 0.0301 & 0.0829 $\pm$ 0.0562 & 0.2008 $\pm$ 0.1091 \\
        \cmidrule{2-7}
        & \multirow{2}{*}{10}
        & DP-SGD & 0.0595 $\pm$ 0.0107 & 0.1597 $\pm$ 0.0156 & 0.0658 $\pm$ 0.0187 & 0.1528 $\pm$ 0.0378 \\
        & & DP-SGD-S & 0.0741 $\pm$ 0.0101 & 0.1565 $\pm$ 0.0129 & 0.0570 $\pm$ 0.0249 & 0.1676 $\pm$ 0.0350 \\
        \cmidrule{2-7}
        & \multirow{2}{*}{1}
        & DP-SGD &  0.0835 $\pm$ 0.0124 & 0.1973 $\pm$ 0.0147 & 0.1501 $\pm$ 0.0156 & 0.3408 $\pm$ 0.0400 \\
        & & DP-SGD-S & 0.0825 $\pm$ 0.0084 & 0.1868 $\pm$ 0.0223 & 0.1367 $\pm$ 0.0205 & 0.3280 $\pm$ 0.0274 \\
        \bottomrule
    \end{tabular}
\end{table}

\section{Broader Impact}
\label{app:broader}
This paper primarily aims to advance the fairness of privacy protection in machine learning. While differential privacy techniques provide formal privacy guarantees, they often overlook disparities in privacy protection across different groups. Our work addresses this gap by proposing methods that enhance the fairness of privacy protection at the group level. 

By improving the equity of privacy protection, our research contributes to building more trustworthy and ethical machine learning systems, particularly in sensitive application domains where privacy concerns are paramount. This focus on privacy fairness helps mitigate potential harms caused by uneven privacy leakage, thereby supporting more responsible and inclusive AI deployment. Moreover, by rigorously evaluating group privacy risks and proposing a privacy fairness metric, this research fosters a deeper understanding of the inherent trade-offs, guiding practitioners and policymakers toward the responsible and equitable deployment of machine learning technologies.

We acknowledge that no method is without limitations, and further research is necessary to simultaneously address other aspects of fairness, such as outcome fairness. Nonetheless, we believe this research helps pave the way for the ethical and equitable deployment of AI systems.

\end{document}